\documentclass[mnsc, nonblindrev]{informs3}

\OneAndAHalfSpacedXI %

\usepackage{natbib, multirow, amsmath, subcaption} %
\usepackage{caption}
\usepackage{booktabs}
\usepackage{accents}

\bibpunct[, ]{(}{)}{,}{a}{}{,}%
\def\bibfont{\small}%
\def\BIBand{and}%
\usepackage{bm}

\usepackage[dvipsnames]{xcolor}

\usepackage{threeparttable}
\usepackage[mathscr]{euscript}
\usepackage{enumerate}
\usepackage{hyperref}
\hypersetup{hidelinks}

\usepackage{makecell}
\newcommand{\mathbbm}[1]{\text{\usefont{U}{bbm}{m}{n}#1}} %

\newcommand{\bI}{\mathbbm{1}}
\newcommand{\cF}{\mathcal{F}}
\newcommand{\cX}{\mathcal{X}}
\newcommand{\cY}{\mathcal{Y}}
\newcommand{\cN}{\mathcal{N}}
\newcommand{\sN}{\mathscr{N}}
\newcommand{\sM}{\mathscr{M}}
\newcommand{\Var}{\mathrm{Var}}
\newcommand{\bE}{\mathrm{E}}
\newcommand{\rv}{\mathrm{v}}
\newcommand{\rR}{\mathrm{R}}
\newcommand{\cA}{\mathcal{A}}
\newcommand{\sH}{\mathscr{H}}
\newcommand{\tr}{\mathrm{tr}}
\newcommand{\indep}{\perp \!\!\! \perp}

\newcommand\blfootnote[1]{%
  \begingroup
  \renewcommand\thefootnote{}\footnote{#1}%
  \addtocounter{footnote}{-1}%
  \endgroup
}

\TheoremsNumberedThrough     %
\ECRepeatTheorems

\EquationsNumberedThrough    %

\usepackage{color}              %

\MANUSCRIPTNO{}
\renewcommand{\theARTICLETOP}{}

\RRHSecondLine{}
\LRHSecondLine{}

\begin{document}

\RUNTITLE{Fine-Tune, Then Rectify}

\RUNAUTHOR{Ye, Zhao, and Wang}

\TITLE{Fine-Tune, Then Rectify}

\ARTICLEAUTHORS{
\AUTHOR{Zikun Ye\textsuperscript{\textdagger}}
\AFF{Foster School of Business, University of Washington, Seattle, WA, \EMAIL{zikunye@uw.edu}}
\AUTHOR{Jinglong Zhao\textsuperscript{\textdagger}}
\AFF{Questrom School of Business, Boston University, Boston, MA, \EMAIL{jinglong@bu.edu}}
\AUTHOR{Lei Wang}
\AFF{Foster School of Business, University of Washington, Seattle, WA, \EMAIL{lei0603@uw.edu}}
}

\ABSTRACT{Driven by recent advances in artificial intelligence, a growing literature has demonstrated the potential of using large language models (LLMs) as scalable surrogates to generate human-like responses. Two common approaches to improve the performance of LLMs include: \textit{fine-tuning}, which aligns the LLM more closely with human responses, and \textit{rectification}, which corrects biases in LLM outputs. In this paper, we develop a two-stage framework that combines fine-tuning and rectification, and optimally allocates limited labeled samples across the two stages. A key insight is that the conventional fine-tuning objective of minimizing mean squared prediction error is generally not aligned with the downstream rectification stage. For mean estimation, we propose to minimize the variance of the prediction errors; for general $M$-estimation, we propose to minimize a scalarized variance metric as the fine-tuning objective. Building on this insight, we leverage the scaling law of fine-tuning to optimally allocate the limited labeled human data between the fine-tuning and rectification stages. Our empirical analysis validates the fine-tuning scaling law and confirms that our proposed optimal allocation rule reliably identifies the optimal sample allocation. We demonstrate substantial efficiency gains in estimation and inference performance relative to fine-tuning or rectification alone, or to employing the conventional mean squared error objective within the fine-tuning then rectification framework. Such efficiency gains translate to significant cost savings for making reliable decisions.
}

\setcounter{page}{1}
\maketitle

\blfootnote{\textdagger\,These authors contributed equally.}

\vspace{-30pt}

\section{Introduction}
\label{sec:introduction}

Businesses routinely conduct surveys to inform managerial decisions, such as conjoint analysis \citep{green1990conjoint}, new product development \citep{wind1997issues}, and pricing \citep{wertenbroch2002measuring}. Traditional survey methods are often recognized to be costly, time-consuming, and difficult to scale.

The rapid advancement of artificial intelligence (AI), particularly in Large Language Models (LLMs), has offered a potential solution to this bottleneck: the use of LLMs as human surrogates. Rather than solely relying on recruiting thousands of human participants, businesses can use LLMs to simulate human-like responses, such as evaluating products and expressing preferences, with minimal latency and near-zero marginal cost. This solution has proven successful well beyond survey studies. It enables a wide range of business applications, including estimating demand curves and price elasticities \citep{gui2023challenge}, modeling discrete choices for assortment optimization \citep{wang2024market}, and content optimization \citep{ye2025lola}.

There is growing empirical evidence supporting the use of LLMs as human surrogates. Trained on vast corpora of human-generated text, LLMs have demonstrated the capacity to encode complex human behavioral patterns and social norms. Recent studies confirm that these models can closely approximate human responses in diverse market research and social science applications \citep{brand2023using, li2024frontiers, toubia2025database, peng2025mega}. As a result, LLM-as-surrogate has become an increasingly attractive tool for accelerating insight generation across different domains, allowing managers to conduct ``virtual'' market tests and experiments with unprecedented speed and scale. There are a growing number of commercial platforms, such as \href{https://www.syntheticusers.com/}{Synthetic Users}, \href{https://www.delve.ai/}{Delve AI}, and \href{https://www.quantilope.com/}{Quantilope}, which offer tools that leverage human-like AI participants to automate market research.

To serve as effective surrogates, LLMs must be properly aligned with the target human population. However, off-the-shelf LLMs are often not suited for this task, as their pretraining optimizes for next-token prediction rather than human behavioral alignment; furthermore, their reliance on public corpora means they lack access to proprietary datasets, often leading to knowledge gaps in specialized domains. Consequently, empirical studies document consistent and systematic discrepancies between LLM-generated outputs and survey responses \citep{motoki2024more,li2025llm}. Furthermore, LLM predictions are also highly sensitive to prompt designs, where minor syntactic changes can yield widely divergent outcomes \citep{brucks2023prompt}. The accuracy of LLM responses could be improved through inference-time prompting techniques that do not update model parameters, such as chain-of-thought prompting to elicit step-by-step reasoning \citep{wei2022cot}, in-context learning with several demonstrations to clarify the task \citep{brown2020gpt3}, and persona prompting to simulate specific subpopulations \citep{peng2025mega}. However, their ability to consistently and accurately reflect human respondents remains questionable in the literature \citep{pezeshkpour2023order, ye2025lola, krsteski2025valid}. These limitations of inference-time prompting techniques highlight the need for methodological advances that leverage LLMs as human surrogates more effectively. Broadly speaking, existing approaches can be characterized into two methods: \textit{fine-tuning}, and \textit{rectification}.

Fine-tuning aligns an LLM by explicitly updating its underlying parameters during a post-training phase using labeled samples. This method has proven successful in bridging the gap between model outputs and human responses across various business applications \citep{ye2025lola,ofek2025balancing}. Its adoption has been further accelerated by parameter-efficient fine-tuning (PEFT) methods, such as LoRA \citep{hu2021lora}, which trains only small low-rank adapters and thus is computationally more efficient. Recent studies confirm that PEFT achieves performance comparable to full fine-tuning at a fraction of the computational cost \citep{han2024peft}. However, fine-tuning relies heavily on the availability of high-quality, task-specific labeled samples. Empirical evidence indicates that fine-tuning performance generally follows a power-law scaling relationship \citep{kaplan2020scaling, hernandez2021transfer, hoffmann2022chinchilla, zhang2024scalingft}. This implies diminishing marginal returns: the value of each additional labeled sample decreases as the total size of labeled samples grows. Moreover, fine-tuning alone does not provide statistical guarantees on the magnitude of errors that still remain in the model, making it difficult to quantify the reliability of the aligned model. From a business perspective, the absence of guarantees makes it difficult to assess whether errors are small enough to support an important decision.

In contrast to fine-tuning, the rectification method does not update the underlying parameters of the pretrained LLM. Instead, the rectification method treats the LLM as a black box, uses labeled samples to quantify the errors of the LLM, and rectifies the outputs from the LLM using the quantified errors post hoc. One of the most prominent examples of the rectification method is prediction-powered inference (PPI), which combines LLM predictions with labeled samples to construct unbiased estimators and valid confidence intervals \citep{angelopoulos2023prediction}. This focus on bias correction and inference reliability is particularly valuable for business decisions, where even small systematic errors can be costly. Similar to fine-tuning, one primary requirement for PPI is also the availability of high-quality labeled samples. Because the variance of the PPI estimator scales inversely with the sample size, this approach also exhibits diminishing marginal returns: the value of each additional labeled sample decreases as the total size of labeled samples grows.

Since both fine-tuning and rectification are subject to diminishing marginal returns, relying exclusively on either method could be inefficient. This motivates us to combine both fine-tuning and rectification methods to take advantage of the rapid initial gains of each method, efficiently allocating our limited labeled data among two methods.

\subsection{A Fine-Tuning then Rectification Framework}

In this paper, we propose a new \textit{fine-tuning then rectification} framework designed to combine both methods.
At a high level, our framework partitions the labeled samples into two disjoint subsets: the first is used to fine-tune the LLM, and the second is used in the subsequent rectification stage.

We analyze how the variance of the final estimator depends on the quality of the fine-tuned LLM.
This analysis motivates us to use a new metric, which we refer to as the ``residual variance,'' as the fine-tuning objective function.
Theoretically, we show that this objective function is more suitable than the traditional mean squared error (MSE) objective function for the downstream rectification stage.
Empirically, we demonstrate the advantage of using the residual variance objective over the MSE objective.
Our usage of the residual variance is conceptually similar to the predict-then-optimize literature \citep{elmachtoub2022smart, ferreira2016analytics}.
This paper can be viewed as proposing a surrogate objective function to replace the traditional MSE objective function.
Our usage of the residual variance is also conceptually similar to the empirical efficiency maximization literature \citep{rubin2008empirical, van2026calibeating}.
This paper can be viewed as a model fine-tuning version of empirical efficiency maximization for PPI.

Building on this insight, we propose a data-driven procedure that leverages empirical scaling laws to determine the optimal sample allocation between fine-tuning and rectification.
The literature has extensively studied the scaling law, showing that LLM performance improves in smooth, predictable ways as model size and sample size increase, often following simple power-law patterns in both pretraining and fine-tuning stages \citep{kaplan2020scaling, hernandez2021transfer, hoffmann2022chinchilla, zhang2024scalingft}.
In our setting, the scaling law describes how the quality of the fine-tuned LLM improves with additional labeled data, which enables us to identify the optimal allocation of limited labeled samples between fine-tuning and rectification that minimizes the variance of the final estimator.
Furthermore, we characterize the properties of the optimal allocation rule and analyze the magnitude of variance reduction under our proposed framework.

In the sample allocation procedure described above, each sample is used either for fine-tuning or for rectification, but not both.
Our framework can be naturally combined with cross-fitting type sample splitting, so that, across different splits, each sample contributes to both fine-tuning and rectification.
This idea has been studied by \citet{zrnic2024cross} in the context of PPI, and has long been recognized in the cross-validation literature as a way to improve upon a single train-test split \citep{bates2024cross, geisser1975predictive, stone1974cross}.
In Online Appendix~\ref{appsec:CrossFitting}, we illustrate how to naturally combine our framework with cross-fitting type sample splitting.
Importantly, cross-fitting does not eliminate the need to choose a new fine-tuning objective; see Online Appendix~\ref{appssec:crossfit}.

We validate our framework through numerical experiments using the Wine Reviews dataset to estimate (i) the average ratings in the population and (ii) the price-rating regression coefficient.
These tasks mirror the day-to-day operations decisions faced by store managers in brick-and-mortar retailing.
When store managers select which stock-keeping units (SKUs) to sell in their stores, they usually need to understand the perceived product quality of each SKU.
We show how to formulate this problem as a mean estimation problem, which we consider in Sections~\ref{sec:setup} and~\ref{sec:method}.
Additionally, when the store managers make pricing decisions, they usually need to understand how perceived quality changes with price.
We show how to formulate this problem as a linear regression problem through the more general view of $M$-estimation, which we consider in Section~\ref{sec:M_estimation}.
In both cases, we show that our framework provides an effective way to understand the perceived quality at scale without incurring huge costs or delays associated with extensive human labeling.

Our empirical analysis shows four main findings.
First, we empirically confirm a scaling law in LLM fine-tuning with our proposed variance-based training objective.
This empirical scaling law predicts the marginal value of one additional labeled sample, which we then use to guide the optimal sample allocation.
Second, we validate our optimal sample allocation.
We show that the theoretical optimum derived from the scaling law accurately guides the optimal sample allocation.
Third, we show the empirical advantage of our proposed framework over a number of existing benchmarks: the sample mean estimator using only the human labeled dataset, standard fine-tuning, standard PPI, a conceptually similar prompting then rectification framework that replaces fine-tuning with prompting, and fine-tuning then rectification but under the standard mean squared error objective.
Our proposed framework outperforms the above benchmarks in most cases.
The only exception is the prompting then rectification framework, which performs better when the sample size is very small.
Beyond mean estimation, when estimating the price-rating regression coefficient, our fine-tuning then rectification framework significantly outperforms the benchmark of linear regression using only the human labeled dataset.

These performance gains translate into substantial sample savings: our method reduces the labeling budget by $48\%$ to $54\%$ compared to the sample mean estimator, to achieve the same level of accuracy.
Finally, we show that the choice of fine-tuning objective is important for the fine-tuning then rectification framework.
Replacing the MSE fine-tuning objective with our proposed variance-based objective brings $22\%$ to $42\%$ variance reduction across different data budget settings.

\subsection{Other Related Literature}
\label{sec:literature}

In addition to the works mentioned above, there are three lines of other related works.

\subsubsection*{LLM Fine-Tuning.}
Fine-tuning has become a standard approach to adapt pretrained language models to downstream tasks by updating model parameters using labeled samples \citep{brown2020gpt3, ouyang2022instruct}. For large models, parameter-efficient fine-tuning methods, such as LoRA \citep{hu2021lora} introduce task-specific low-rank matrix adapters while freezing most parameters, making fine-tuning computationally efficient in practice. A large empirical literature on neural scaling laws shows that model performance tends to follow smooth power-law relationships with model size, data size, and compute scale \citep{kaplan2020scaling, hernandez2021transfer, hoffmann2022chinchilla, zhang2024scalingft}, providing us with a parametric function to use the size of labeled samples to predict the model quality.

Other fine-tuning methods, such as RLHF and DPO, update models using a separate trained reward model or preference data \citep{ouyang2022instruct, rafailov2023direct}. Beyond vanilla fine-tuning with commonly used objective functions such as mean squared errors and cross-entropy losses, recent works have started to modify objectives to target downstream tasks directly. For example, \cite{vafa2025estimating} propose a debiased fine-tuning method for using foundation models to estimate the gender wage gap. In our proposed framework, we focus on the setting where an LLM is used as a surrogate for a PPI estimator and design a fine-tuning objective that directly minimizes the variance governing the efficiency of the final PPI estimator.

To the best of our knowledge, this is the first work to formally analyze the connection between fine-tuning and rectification.
While recent independent work by \citet{krsteski2025valid} empirically observes that combining these methods yields improvements, their approach relies on standard fine-tuning procedures that do not take into account the downstream rectification task.
In contrast, our framework derives a fine-tuning objective explicitly tailored to the rectification stage and provides a rigorous method for optimally allocating labeled samples between the two stages.

\subsubsection*{Prediction-Powered Inference.}
The PPI method combines a small set of labeled samples with a large set of model predictions to construct unbiased estimators and confidence intervals, while treating the prediction model as a potentially biased black box \citep{angelopoulos2023prediction}.
Follow-up works, such as the PPI++ method from \citet{angelopoulos2023ppi}, develop more efficient variants.
Recent research further connects PPI to the classical surrogate outcome literature, showing that PPI-type estimators can be interpreted as using predictions as surrogates and proposing re-calibration procedures that achieve improved efficiency \citep{ji2025predictions}.
Our work builds on this line of research by taking the prediction model to be an LLM that can itself be improved via fine-tuning, and by explicitly designing both the fine-tuning objective and the allocation of labeled data to minimize the asymptotic variance of the PPI estimator.

In an earlier work, \citet{zrnic2024cross} propose a method called the cross-PPI, which allows every label to serve both model training and rectification through cross-fitting.
The cross-PPI approach and our fine-tuning then rectification approach are not mutually exclusive: our approach can itself be combined with cross-fitting.
Online Appendix~\ref{appsec:CrossFitting} describes the details of combining our approach with cross-fitting, and Online Appendix~\ref{appssec:crossfit} reports the results from numerical experiments.
Importantly, cross-fitting does not eliminate the need to choose a new fine-tuning objective.
Online Appendix~\ref{appssec:crossfit} shows that the variance-based objective consistently improves conventional MSE objective when our approach is combined with cross-fitting.

PPI is only one rectification method.
There are other rectification methods.
For example, \citet{wang2024market} propose a transfer learning based method to correct the bias from LLMs. \cite{han2026debiasing} consider a penalized regression estimator that shrinks an estimator generated from human data toward the LLM simulator. In this paper, we focus on the standard PPI method for clarity, though our core insights extend to other rectification approaches.

\subsubsection*{LLMs for Business Applications.}
There is a fast growing literature that evaluates whether LLMs can serve as surrogates for human respondents in high-fidelity business applications, ranging from demand estimation to consumer choice modeling. \citet{brand2023using} examine the use of LLMs to estimate willingness to pay. They document mixed results: LLMs can reproduce aggregate patterns reasonably well, but they generate insufficient heterogeneity across individuals even after fine-tuning. And the fine-tuned models struggle to extrapolate beyond the training distribution. Other work studies whether LLMs can reproduce human preferences or attitudes in specific domains, such as brand perception \citep{li2024frontiers}, computational social science \citep{argyle2023out, ziems2024can}, and content experiments \citep{ye2025lola}, all highlighting their potential to augment, rather than replace, human data. At the same time, several recent papers also document important limitations. \citet{gui2023challenge} raise endogeneity concerns when using LLMs to simulate human behavior from a causal-inference perspective. \citet{wang2024market} show that while LLM-generated responses are useful for data augmentation in choice modeling, they often exhibit systematic biases relative to human samples and require careful calibration.

Our contribution is methodological and complements this literature. Rather than assessing LLMs as stand-alone survey respondents, we study how to combine fine-tuning and rectification when LLM outputs are used as surrogate responses. We further show how to optimally allocate scarce human-labeled data between improving the surrogate model and correcting its residual biases to obtain accurate estimates and make reliable decisions.

\subsection*{Roadmap} The rest of the paper is structured as follows.
In Section~\ref{sec:setup}, we formalize the problem setup and define the combined fine-tuning and rectification estimator. In Section~\ref{sec:method}, we derive the main theoretical results of our framework. We analyze the estimator's variance to motivate a variance-based fine-tuning objective. We then leverage empirical scaling laws to derive the optimal sample allocation rule, and establish the theoretical conditions under which our method outperforms standard baselines. In Section~\ref{sec:M_estimation}, we extend this framework from mean estimation to the more general $M$-estimation problems. In Section~\ref{sec:empirical}, we conduct empirical analysis for mean estimation on the Wine Reviews dataset. Section~\ref{sec:mest-case} studies the price-rating regression coefficient as an $M$-estimation case study. In Section~\ref{sec:conclusion}, we conclude the paper.

\section{Problem Setup}
\label{sec:setup}

We consider a mean estimation problem in a setting where we have access to both a small, labeled dataset and a large, unlabeled dataset. We extend this mean estimation problem to the more general $M$-estimation problem in Section~\ref{sec:M_estimation}. Let there be $n$ samples in the small labeled dataset $\sN = \big\{ (\bm{X}_i, Y_i) \big\}_{i=1}^n$, where each sample $(\bm{X}_i,Y_i)$ is an independent and identically distributed (i.i.d.) replica of a representative random variable $(\bm{X},Y)$ sampled from a joint distribution $(\bm{X},Y) \sim \cF$. Let there be $m$ samples in the large unlabeled dataset $\sM = \big\{ \tilde{\bm{X}}_j \big\}_{j=1}^m$, where each sample $\tilde{\bm{X}}_j$ is an i.i.d. replica of a representative random variable $\bm{X}$ sampled from the marginal distribution $\bm{X} \sim \cF_{\bm{X}}$, where $\cF_{\bm{X}}$ stands for the marginal distribution over $\bm{X}$ according to the joint distribution $\cF$. Let $\cX \times \cY$ be the set that $\cF$ takes values from. We also have access to an LLM predictor $f: \cX \to \cY$ that is trained on data independent of both datasets. This LLM predictor $f$ can take unstructured data, such as textual information, as inputs. See our empirical application in Section~\ref{sec:empirical} for an illustration of the inputs and outputs of an LLM predictor.

Given these two datasets and the predictor, there are two families of methods for leveraging LLMs as human surrogates: fine-tuning, which adapts the model using the small labeled dataset, and rectification, which corrects model predictions without modifying the model itself.

\begin{enumerate}
\item \textbf{Fine-Tuning (FT)}. We can use the small labeled dataset to fine-tune the LLM predictor, and then use the fine-tuned LLM predictor to construct estimators.
\item \textbf{Prediction-Powered Inference (PPI)}. We can use the small labeled dataset to measure the errors of the LLM predictor, and then rectify the errors on the large unlabeled dataset.
\end{enumerate} In this paper, we propose to combine fine-tuning and prediction-powered inference to achieve the merits of both approaches. We study a \emph{single-fit sample-splitting design}, in which each labeled sample enters exactly one stage --- fine-tuning (FT) or prediction-powered inference (PPI) --- and ask for the optimal allocation of the $n$ samples between the two.

For any $0 < s < n$, let $s$ be the number of samples we allocate to FT. Consequently, $n-s$ is the number of samples we allocate to PPI. Let $f^{(s)}: \cX \to \cY$ be the LLM predictor that is fine-tuned on $s$ samples of the small labeled dataset. We propose to use the prediction-powered estimator \citep{angelopoulos2023prediction} based on the fine-tuned predictor $f^{(s)}(\cdot),$ defined as follows,
\begin{align}
\widehat{\mu} = \frac{1}{n-s} \sum_{i=1}^{n-s} \big( Y_i - f^{(s)}(\bm{X}_i) \big) + \frac{1}{m} \sum_{j=1}^m f^{(s)}(\tilde{\bm{X}}_j). \label{eqn:Estimator}
\end{align}

It is easy to see that $\widehat{\mu}$ as defined in \eqref{eqn:Estimator} is unbiased,
\begin{align*}
\bE[\widehat{\mu}] = \bE\big[Y\big] - \bE\big[f^{(s)}(\bm{X})\big] + \bE\big[f^{(s)}(\bm{X})\big] = \bE\big[Y\big],
\end{align*} where the first equality is because $\bm{X}_i$ from the small labeled dataset and $\tilde{\bm{X}}_j$ from the large unlabeled dataset are sampled from the same distribution. As the estimator $\widehat{\mu}$ is unbiased, the mean squared error of the estimator is equal to its variance. A lower variance directly implies more accurate estimates, narrower confidence intervals, and more efficient use of scarce labeled data. In the following, our goal is to combine fine-tuning and PPI to make the most efficient use of limited human labels.

\section{Our Method}
\label{sec:method}

In this section, we begin by analyzing the variance of the PPI estimator and derive its implications for how the LLM should be fine-tuned in Section~\ref{ssec:loss_func}. We then use the scaling law in fine-tuning to characterize the optimal allocation of labeled samples between fine-tuning and PPI in Section~\ref{ssec:sample_allocate}, assuming the exact fine-tuning scaling law is known. We provide a data-driven method to estimate the parameters of the scaling law in Online Appendix~\ref{appsec:ramp-up}. Finally, we analyze the performance of our proposed method in Section~\ref{ssec:perform_analysis} and provide inference results in Section~\ref{ssec:inference}.

\subsection{A New Loss Function for Fine-Tuning}
\label{ssec:loss_func}

We start by examining the variance of $\widehat{\mu}$ as defined in \eqref{eqn:Estimator}, following the same analysis in \citet{angelopoulos2023prediction}.
\begin{align}
& \bE\big[(\widehat{\mu} - \bE[Y])^2\big] \nonumber \\
= & \bE\bigg[ \Big( \frac{1}{n-s} \sum_{i=1}^{n-s} \big( Y_i - f^{(s)}(\bm{X}_i) \big) + \frac{1}{m} \sum_{j=1}^m f^{(s)}(\tilde{\bm{X}}_j) - \bE[Y] \Big)^2 \bigg] \nonumber \\
= & \bE\bigg[ \Big( \frac{1}{n-s} \sum_{i=1}^{n-s} \big( Y_i - f^{(s)}(\bm{X}_i) + \bE\big[f^{(s)}(\bm{X})\big] - \bE[Y] \big) + \frac{1}{m} \sum_{j=1}^m \big( f^{(s)}(\tilde{\bm{X}}_j) - \bE\big[f^{(s)}(\bm{X})\big] \big) \Big)^2 \bigg] \nonumber \\
= & \ \frac{1}{(n-s)^2} \sum_{i=1}^{n-s} \bE\bigg[ \big( Y_i - f^{(s)}(\bm{X}_i) + \bE\big[f^{(s)}(\bm{X})\big] - \bE[Y] \big)^2 \bigg] + \frac{1}{m^2} \sum_{j=1}^m \bE\bigg[ \big( f^{(s)}(\tilde{\bm{X}}_j) - \bE\big[f^{(s)}(\bm{X})\big] \big)^2 \bigg] \nonumber \\
= & \ \frac{1}{n-s} \Var\bigg( Y - f^{(s)}(\bm{X}) \bigg) + \frac{1}{m} \Var\bigg( f^{(s)}(\bm{X}) \bigg) \label{eqn:VarianceDecomposition}
\end{align} where the third equality is because $\big( Y_i - f^{(s)}(\bm{X}_i) \big)$ and $f^{(s)}(\tilde{\bm{X}}_j)$ are independent conditional on the fine-tuning stage.

Because we have a large unlabeled dataset, the second term is small when $m$ is large. So we focus our attention primarily on the first term, $\frac{1}{n-s} \Var\big( Y - f^{(s)}(\bm{X}) \big)$. Recall that in the single-fit splitting design each labeled sample enters exactly one of FT or PPI. We make the following two observations regarding the benefits of allocating more samples to the FT stage and to the PPI stage, which present a trade-off in sample allocation.
\begin{enumerate}
\item The term $\frac{1}{n-s} \Var\big( Y - f^{(s)}(\bm{X}) \big)$ decreases when $n-s$ increases. This observation indicates that the more data we use toward PPI, the smaller this term becomes.
\item The term $\frac{1}{n-s} \Var\big( Y - f^{(s)}(\bm{X}) \big)$ decreases when $\Var\big( Y - f^{(s)}(\bm{X}) \big)$ decreases. This observation indicates that the more data we use toward FT, the higher quality is the fine-tuned predictor $f^{(s)}(\cdot)$, which then makes $\Var\big( Y - f^{(s)}(\bm{X}) \big)$ smaller.
\end{enumerate} It is worth noting that the bias of the fine-tuned predictor $f^{(s)}(\cdot)$ does not play an important role in the expression in \eqref{eqn:VarianceDecomposition}. Even when the fine-tuned predictor $f^{(s)}(\cdot)$ is biased, our estimator $\widehat{\mu}$ as defined in \eqref{eqn:Estimator} can still have a small variance, as long as the variance of $Y - f^{(s)}(\bm{X})$ is small. See Example~\ref{exa:BiasedPredictor} below.

\begin{example}[Fine-tuning a Biased Predictor]
\label{exa:BiasedPredictor} We consider an extreme case where the fine-tuned predictor always under-predicts by a constant, that is, there exists $c > 0$ such that $Y - f^{(s)}(\bm{X}) = c$. In this case, the mean squared error $\bE\big[\big(Y-f^{(s)}(\bm{X})\big)^2\big] = c^2$ can be large. But the variance $\Var\big( Y - f^{(s)}(\bm{X}) \big) = 0$ is always equal to zero. This implies that a predictor with high bias can still be optimal for the prediction-powered estimator, provided that the bias is a \textit{constant} with zero variance.
\hfill \halmos
\end{example}

Example~\ref{exa:BiasedPredictor} motivates the use of variance $\Var(Y - f^{(s)}(\bm{X}))$, rather than the traditional mean squared error (MSE), as the loss function for fine-tuning.
We refer to this variance $\Var(Y - f^{(s)}(\bm{X}))$ as the ``residual variance.''
The residual variance is different from $\Var(f^{(s)}(\bm{X}))$, the variance of the fine-tuned LLM.

\begin{remark}
\label{rmk:objectives}
For mean estimation, the MSE and residual variance objectives are equivalent under some conditions.
When the model class allows for additive constant shifts, such as through an unconstrained intercept, minimizing MSE implicitly also minimizes the residual variance.
In the setting with an unconstrained intercept, the two objectives essentially lead to the same solution.
However, the two objectives can behave differently in more practical settings where data is not large enough or when the model is trained with mini batches, under regularization, or under early stopping.
In Section~\ref{sssec:compare-estimators}, we empirically compare the two objectives and demonstrate the benefits of using the residual variance as the objective.

For general $M$-estimation, however, the objectives are generally not equivalent. The residual that governs downstream precision is weighted by each observation's influence on the estimand (Section~\ref{sec:M_estimation}); when that weight varies across observations, an additive shift of the surrogate changes the weighted residual by a nonconstant amount, so a free intercept no longer aligns MSE with the estimand-aware criterion.
Standard fine-tuning typically minimizes cross-entropy in next-token prediction, which equivalently maximizes the likelihood of predicting the next token.
Such a standard fine-tuning objective is not tied to the downstream rectification stage at all.
Sometimes during fine-tuning, numerical outcomes are also commonly handled by a discrete classification head, which we discuss in Online Appendix~\ref{appsec:ce}.
Section~\ref{sec:M_estimation} derives the general estimand-aware objective, shows that minimizing it weakly dominates any alternative fine-tuning objective, and exhibits a strict gap in an example; Section~\ref{sec:mest-case} provides an empirical boundary case in which the two objectives nearly coincide and a controlled setting in which they do not.
\hfill \halmos
\end{remark}

\subsection{Optimal Sample Allocation}
\label{ssec:sample_allocate}

Fine-tuning is a flexible method that allows us to use variance $\Var\big( Y - f^{(s)}(\bm{X}) \big)$ as the loss function. Empirically, it is widely accepted that fine-tuning can effectively reduce the loss of the LLM predictor following a smooth power-law relationship, i.e., scaling law \citep{kaplan2020scaling,hernandez2021transfer,hoffmann2022chinchilla,zhang2024scalingft},
\begin{align*}
\Var\Big( Y - f^{(s)}(\bm{X}) \Big) = a s^{-\alpha} + b,
\end{align*} where $\alpha, a>0,~ b \geq 0$ are task-specific constants. Note that scaling law formulations typically incorporate model size as an independent variable. In our setting, we treat the size of the underlying foundation model as fixed, because size choices are typically constrained by practical considerations such as inference cost and available base models. Moreover, many open-weight foundation models offer only a small set of discrete size options. OpenAI’s gpt-oss provides only 20B-parameter and 120B-parameter variants~\citep{openai2025gptoss}. Thus, we abstract away from the model size decision and treat it as given.

When the constants are known, we can minimize variance in the expression in \eqref{eqn:VarianceDecomposition} by solving the following univariate optimization problem,
\begin{align}
\min_{s \in (0,n)} \ \frac{\Var\big( Y - f^{(s)}(\bm{X}) \big)}{n-s} \ = \ \min_{s \in (0,n)} \ \frac{a s^{-\alpha} + b}{n-s}. \label{eqn:Minimization}
\end{align} We can express the optimal solution to \eqref{eqn:Minimization} through the following theorem.

\begin{theorem}[Optimal Allocation Rule]
\label{thm:OptimalAllocation} The optimal solution to \eqref{eqn:Minimization}, $s^*$, is given as the unique solution to
\begin{align}
\alpha a n s^{-\alpha-1} - (\alpha+1) a s^{-\alpha} - b = 0. \label{eqn:NumericalSolve}
\end{align}
\end{theorem}

The optimal solution $s^*$ can easily be calculated through numerically solving equation \eqref{eqn:NumericalSolve} via a binary search, using the property that the left-hand side of equation \eqref{eqn:NumericalSolve} is monotone decreasing in $s$. The proof can be found in Online Appendix~\ref{appsec:proof}.

Next, we investigate the properties of the optimal allocation $s^*(a,b,\alpha,n)$ and its dependence on scaling law parameters $a,b,\alpha$ and the total sample size $n$.

\begin{proposition}[Characterization of the Optimal Allocation Rule]
\label{prop:monotonicity_abn} The unique solution $s^*(a,b,\alpha,n)$ to the optimization problem \eqref{eqn:Minimization} has the following properties:
\begin{enumerate}[(i)]
\item $s^*(a,b,\alpha,n)$ is nondecreasing in $a$, and strictly increasing in $a$ whenever $b>0$.
\item $s^*(a,b,\alpha,n)$ is strictly decreasing in $b$.
\item $s^*(a,b,\alpha,n)$ is strictly increasing in $n$.
\item In the special case when $b=0$,
\[
\frac{s^*}{n}=\frac{\alpha}{\alpha+1},
\]
which is independent of $a$ and $n$.
\item The fine-tuning fraction $s^*/n$ is nondecreasing in $a$ (strictly increasing if $b>0$), strictly decreasing in $b$, and nonincreasing in $n$. Specifically, if $b>0$, the fraction vanishes asymptotically
\[
\lim_{n \to \infty} \frac{s^*}{n} = 0.
\]
Moreover, its derivative with respect to $n$ is uniformly bounded if $\alpha<1$:
\begin{align}
0 \;\ge\; \frac{\partial}{\partial n}\Big(\frac{s^*}{n}\Big)
\;\ge\;
-\,\frac{\alpha}{\alpha+1}\,\frac{b}{a}\,n^{\alpha-1}.
\label{eq:dx_dn_bound}
\end{align}
\end{enumerate}
\end{proposition}

The proof of Proposition~\ref{prop:monotonicity_abn} can be found in Online Appendix~\ref{appsec:proof}. The first three properties describe how the optimal number of fine-tuning samples, $s^*$, reacts to changes in the parameters $a$, $b$, and $n$. First, $s^*$ is nondecreasing in $a$. Since $a$ is the scale of the reducible error component, a higher value implies a larger marginal benefit from additional fine-tuning data at any fixed $s$. Second, $s^*$ is strictly decreasing in $b$. The parameter $b$ can be interpreted as the noise floor, which is due to the intrinsic variance of the data or approximation error of the model. If $b$ increases, it is optimal to allocate fewer samples for fine-tuning. Third, it is intuitive that when the total available samples $n$ grow, the samples allocated to fine-tuning naturally grow with it.

The last two properties characterize the behavior of the fine-tuning fraction, $s^*/n$. Property \textit{(iv)} establishes a theoretical baseline: in the case where there is no irreducible variance, i.e., $b=0$, the optimal fraction is a perfect constant, ${\alpha}/(\alpha+1)$, which depends solely on the power-law exponent. This value serves as the upper bound for the allocation ratio. In the more common case where $b > 0$, property \textit{(v)} reveals that the ratio $s^*/n$ deviates from this baseline. It is strictly decreasing in both $b$ and $n$. Theoretically, this fraction vanishes asymptotically as the dataset grows indefinitely, i.e., $\lim_{n \to \infty} s^*/n = 0$. This implies that given an infinite budget of labeled data, the optimal fine-tuning set size $s^*$ would grow indefinitely, yet the fraction $s^*/n$ will asymptotically vanish.

For practical purposes, Equation \eqref{eq:dx_dn_bound} provides an important bound on this behavior. It establishes that the rate of this decay is uniformly bounded by a term proportional to $b/a$. This implies that if $b/a$ is small, the optimal ratio remains close to the theoretical ceiling $\alpha/(\alpha+1)$ if the dataset size $n$ is not too large.

\subsection{Performance Analysis of FT+PPI}
\label{ssec:perform_analysis}

Theorem~\ref{thm:OptimalAllocation} shows that, under the scaling law, the variance minimization problem has a unique interior minimizer $s^* \in (0,n)$. While a closed-form solution for $s^*$ is not available, Proposition~\ref{prop:monotonicity_abn} characterizes the properties of the optimal fine-tuning sample size and ratio.

We now turn to the performance analysis of the proposed method. We study conditions under which the FT+PPI estimator outperforms the naive sample mean in terms of variance, given that both methods are unbiased. In the following analysis, the scaling law parameters $a, b, \alpha$ are known quantities. In practice, these parameters can be estimated from a data-driven ``ramp-up'' procedure. See Online Appendix~\ref{appsec:ramp-up}.

For a given fine-tuning budget $s$, we define the variance-based goodness-of-fit as
\begin{align*}
R^{2}(s) = 1-\frac{\Var\!\left(Y-f^{(s)}(\bm{X})\right)}{\Var(Y)}.
\end{align*} Given $n$ labeled samples $(\bm{X}_i,Y_i)$ and $m$ unlabeled samples $\tilde{\bm{X}}_j$, the FT+PPI estimator $\widehat{\mu}$ utilizes the fine-tuned model $f^{(s)}$. Assuming the number of unlabeled samples $m$ is sufficiently large, the variance of the estimator is dominated by the prediction error on the small labeled dataset (see expression~\eqref{eqn:VarianceDecomposition}). On the other hand, the baseline sample mean estimator, $\widehat{\mu}_{\mathrm{mean}}=\frac{1}{n}\sum_{i=1}^{n} Y_i$, has variance $\Var(\widehat{\mu}_{\mathrm{mean}})=\Var(Y)/n$.

Proposition~\ref{prop:ppi_vs_mean_R2} below provides a necessary and sufficient condition on $R^2(s)$ for the FT+PPI estimator to outperform the sample mean estimator. It shows that the goodness-of-fit $R^2(s)$ obtained in the fine-tuning stage is important for deciding whether the subsequent PPI stage reduces variance relative to the baseline. The proof of Proposition~\ref{prop:ppi_vs_mean_R2} is given in Online Appendix~\ref{appsec:proof}.

\begin{proposition}[Criterion for Variance Reduction]
\label{prop:ppi_vs_mean_R2} Fix $n$ and $s\in(0,n)$, and assume the number of unlabeled samples is large $m \to \infty$. Then, $\Var(\widehat{\mu})<\Var(\widehat{\mu}_{\mathrm{mean}})$ holds if and only if
\begin{align*}
R^{2}(s)>\frac{s}{n}.
\end{align*}
\end{proposition}

Proposition~\ref{prop:ppi_vs_mean_R2} shows that FT+PPI strictly outperforms the sample mean estimator if and only if the $R^2(s)$ from the fine-tuned model exceeds the fraction of labeled data consumed by fine-tuning, $s/n$. Intuitively, Proposition~\ref{prop:ppi_vs_mean_R2} suggests that the variance reduction achieved by the fine-tuned model $f^{(s)}(\bm{X})$ must outweigh the statistical cost of reducing the sample size from $n$ to $n-s$.

Proposition~\ref{prop:ppi_vs_mean_R2} assumes an asymptotic regime where the number of unlabeled samples $m \to \infty$, implying that the second variance term in equation~\eqref{eqn:VarianceDecomposition} vanishes. This assumption is justified by the fact that the costs of acquiring unlabeled data and computing LLM responses are typically negligible compared to the costs of collecting human labels. We can also easily incorporate the variance contribution from finite $m$ into Proposition~\ref{prop:ppi_vs_mean_R2}.

Proposition~\ref{prop:ppi_vs_mean_R2} provides a condition for variance reduction. Yet it does not directly suggest whether this condition is achievable.
Next, we incorporate the scaling law to derive the boundary conditions for improvement. For notational convenience, denote $\sigma^2 = \Var(Y)$. The proof of Proposition~\ref{prop:variance_reduction_conditions} is given in Online Appendix~\ref{appsec:proof}.

\begin{proposition}[Conditions for Variance Reduction under Scaling Laws]
\label{prop:variance_reduction_conditions} Suppose the residual variance follows the scaling law $\Var(Y-f^{(s)}(\bm{X}))=a s^{-\alpha}+b$ with parameters $a, \alpha > 0$, $b \ge 0$, and assume the number of unlabeled samples $m \to \infty$. Define the variance discriminant function $q(s)$ as follows
\begin{align*}
q(s) := \underbrace{\sigma^{2}(n-s)}_{\text{Baseline Contribution}} - \underbrace{n\bigl(a s^{-\alpha}+b\bigr)}_{\text{FT+PPI Contribution}}, \qquad s \in (0, n).
\end{align*} Then the following properties hold:
\begin{enumerate}[(i)]
\item {Equivalence:} There exists a valid allocation $s \in (0, n)$ (and thus an optimal $s^*$) such that FT+PPI strictly improves upon the sample mean estimator if and only if $\sup_{s\in(0,n)} q(s) > 0$.

\item {Feasibility Condition:} Due to the strict concavity of $q(s)$, and because the feasibility condition places the maximizer inside $(0,n)$, the maximum is attained and unique. There exists a valid allocation $s \in (0, n)$ (and thus an optimal $s^*$) such that FT+PPI strictly improves upon the sample mean estimator if and only if:
\begin{align}
\frac{b}{\sigma^{2}}
<
1-\left(1+\frac{1}{\alpha}\right)
\left(\frac{a\alpha n^{-\alpha}}{\sigma^{2}}\right)^{\!1/(\alpha+1)}.
\label{eq:feasibility_condition}
\end{align}
\end{enumerate}
\end{proposition}

Condition~\eqref{eq:feasibility_condition} serves as a verifiable decision rule, showing that the variance reduction of FT+PPI depends on the following two factors. First, the task learnability, represented by the ratio of irreducible error $b/\sigma^2$, plays a limiting role. The parameter $b$ dictates the best achievable error of the fine-tuned model given enough data. If the irreducible error $b$ is too high relative to the inherent data variance $\sigma^2$, the model's predictive power will never suffice to justify the ``cost'' of removing samples from the PPI estimation set to fine-tuning. Second, the total dataset size $n$ determines the threshold for success. In Condition~\eqref{eq:feasibility_condition}, the sample size $n$ appears in the negative term on the right-hand side. As $n$ increases, the term proportional to $n^{-\alpha/(\alpha+1)}$ decreases, causing the entire right-hand side to increase and approach 1. Consequently, a larger $n$ raises the feasibility threshold, making it easier for the condition to hold. This implies that even for tasks with moderate to hard learnability, FT+PPI can become viable if the dataset becomes sufficiently large.

Condition~\eqref{eq:feasibility_condition} also serves as a valuable diagnostic tool. In practice, one does not need to execute the entire two-stage fine-tuning and rectification procedures to determine whether it leads to variance reduction. Instead, by relying on prior knowledge of the task's scaling law or estimating the parameters $a, b$, and $\alpha$ on a subset of the data, we can evaluate whether Condition~\eqref{eq:feasibility_condition} holds. If the inequality is not satisfied, the fitted FT+PPI specification cannot strictly improve on the sample mean; the practitioner should then compare the sample mean with the $s = 0$ rectified zero-shot design and any other available surrogate class.

\subsection{Inference}
\label{ssec:inference}

To conclude the discussion of our method, we state the inference method provided by \citet{angelopoulos2023prediction} to construct a confidence interval for mean estimation.

First, we estimate the sample variance
\begin{align*}
\widehat{\sigma}_{f^{(s)}}^2 = \frac{1}{m-1} \sum_{j=1}^m \bigg( f^{(s)}(\tilde{\bm{X}}_j) - \frac{1}{m}\sum_{j=1}^m f^{(s)}(\tilde{\bm{X}}_j) \bigg)^2
\end{align*} in the large unlabeled dataset. Second, we estimate the sample variance
\begin{align*}
\widehat{\sigma}_{Y-f^{(s)}}^2 = \frac{1}{n-s-1} \sum_{i=1}^{n-s} \bigg( Y_i - f^{(s)}(\bm{X}_i) - \frac{1}{n-s}\sum_{i=1}^{n-s} \big(Y_i - f^{(s)}(\bm{X}_i)\big) \bigg)^2
\end{align*} in the small labeled dataset. Finally, for any $\delta \in (0,1)$, we calculate the following confidence interval
\begin{align*}
\bigg[\widehat{\mu} - z_{1-\frac{\delta}{2}}\sqrt{\frac{\widehat{\sigma}_{Y-f^{(s)}}^2}{n-s} + \frac{\widehat{\sigma}_{f^{(s)}}^2}{m}}, \ \widehat{\mu} + z_{1-\frac{\delta}{2}}\sqrt{\frac{\widehat{\sigma}_{Y-f^{(s)}}^2}{n-s} + \frac{\widehat{\sigma}_{f^{(s)}}^2}{m}}\bigg].
\end{align*} Proposition~1 of \citet{angelopoulos2023prediction} suggests that the confidence interval constructed in this way is asymptotically valid.

\section{Extension to \texorpdfstring{$M$}{M}-Estimation}
\label{sec:M_estimation}

In this section, we provide a more general framework for $M$-estimation. We follow the same problem setup as in Section~\ref{sec:setup}. Instead of population mean estimation, we consider the estimation of a vector of $d$-dimensional parameters $\bm{\theta}^* \in \rR^d$ defined as
\begin{align*}
\bm{\theta}^* = \argmin_{\bm{\theta} \in \rR^d} \bE_{(\bm{X},Y) \sim \cF}\Big[l(\bm{X}, Y; \bm{\theta})\Big],
\end{align*} where $l(\bm{X},Y;\bm{\theta})$ is the loss function that is twice differentiable with respect to $\bm{\theta}$ and such that the expected loss has a unique minimizer. The parameter vector $\bm{\theta}^* \in \rR^d$ is thus the solution to the following first order condition
\begin{align*}
\bE_{(\bm{X},Y) \sim \cF}\Big[\bm{\psi}(\bm{X},Y;\bm{\theta})\Big] = \bm{0},
\end{align*} where $\bm{\psi}(\bm{X},Y;\bm{\theta}) = \nabla_{\bm{\theta}} l(\bm{X},Y;\bm{\theta})$ is the score function. We illustrate the above notations through the following examples.

\begin{example}[Mean Estimation] In the mean estimation problem, we would like to estimate the mean of an outcome $\theta = \bE[Y]$. Mean estimation is a special case of $M$-estimation where the loss function is defined as $l(\bm{X},Y;\theta) = \frac{1}{2}(Y - \theta)^2$. In this case, the score function is given by $\psi(\bm{X},Y;\theta) = \theta - Y$. The first order condition $\bE\big[\psi(\bm{X},Y;\theta)\big] = 0$ implies that $\theta = \bE[Y]$, which is exactly the mean of an outcome.
\hfill \halmos
\end{example}

\begin{example}[Categorical Response Estimation] Estimating the distribution of categorical responses is fundamental in survey-based market research, such as determining the market share of $d$ options (e.g., options A, B, C, D when $d=4$) in a multiple-choice survey. Beyond surveys, such categorical response estimation problem is also useful in revenue management applications such as stock keeping unit (SKU) rationalization, where managers identify underperforming products based on their market shares \citep{lee1993hewlett, lee1997modelling, alfaro2003value}. Here, $Y \in [d]$ represents the selection of one specific option. We aim to estimate the probability vector $\bm{\theta} = \big(\Pr(Y=1), \Pr(Y=2), \dots, \Pr(Y=d)\big)^\top$. This is a special case of $M$-estimation where the loss function is defined as $l(\bm{X},Y;\bm{\theta}) = \sum_{l=1}^d \frac{1}{2}(\bI\{Y=l\} - \theta_l)^2$. In this case, the score function is given by $\bm{\psi}(\bm{X},Y;\bm{\theta}) = \bm{\theta} - \big(\bI\{Y=1\}, \dots, \bI\{Y=d\}\big)^\top$. The first order condition $\bE\big[\bm{\psi}(\bm{X},Y;\bm{\theta})\big] = \bm{0}$ implies that $\bm{\theta} = \big(\Pr(Y=1), \dots, \Pr(Y=d)\big)^\top$, which recovers the categorical response probabilities.
\hfill \halmos
\end{example}

\begin{example}[Linear Regression] Linear regression is widely used in business applications. A classic use case is estimating the price-demand relationship \citep{gallego2019revenue, gui2023challenge, talluri2006theory, tirole1988theory}, where $Y$ represents demand and $\bm{X}$ represents product attributes (e.g., prices). Note that in our LLM framework, $\bm{X}$ typically denotes the raw textual query given to LLMs. Here, with a slight abuse of notation, we let $\bm{X}$ be the feature vectors extracted from the text, e.g., product profile and price. Linear regression is a special case of $M$-estimation where the loss function is defined as $l(\bm{X},Y;\bm{\theta}) = \frac{1}{2}(Y - \bm{X}^\top \bm{\theta})^2$. In this case, the score function is given by $\bm{\psi}(\bm{X},Y;\bm{\theta}) = \bm{X} (\bm{X}^\top \bm{\theta} - Y)$. The first order condition $\bE\big[\bm{\psi}(\bm{X},Y;\bm{\theta})\big] = \bm{0}$ implies that $\bm{\theta} = \bE[\bm{X}\bm{X}^\top]^{-1}\bE[\bm{X} Y]$, which is the population ordinary least squares estimator.
\hfill \halmos
\end{example}

\begin{example}[Choice Model Estimation] Choice model estimation, particularly using the Multinomial Logit (MNL) model, is widely used in revenue management applications such as assortment optimization and pricing, where a manager chooses the SKUs and sets their prices to maximize profit, revenue, or market share \citep{berry1993automobile, gallego2019revenue, talluri2006theory}, as well as market research, e.g., recent work has also explored integrating LLMs with such choice models to analyze consumer choice behavior \citep{brand2023using, wang2024market}. In this problem, we observe samples drawn from the joint distribution of $(\bm{X}_1, \bm{X}_2, \dots, \bm{X}_K, Y)$, where $Y \in \{0, 1, \dots, K\}$ represents the choice among $K$ options (with $0$ denoting the no-purchase option). Similar to the linear regression example, while $\bm{X}$ typically denotes the raw textual input in our framework, here we abuse notation to let $\bm{X}_k \in \mathbb{R}^d$ denote the feature vector for the $k$-th option extracted from the text here. We aim to estimate the coefficients $\bm{\theta}$ in the standard MNL model:
\begin{align*}
\Pr(Y=k) = \frac{\exp(\bm{X}_k^\top \bm{\theta})}{1 + \sum_{\kappa=1}^K \exp(\bm{X}_\kappa^\top \bm{\theta})}, && \forall k \in [K].
\end{align*} MNL estimation is a special case of $M$-estimation where the loss function is defined as
\begin{align*}
l(\bm{X}_1,\bm{X}_2,...,\bm{X}_K,Y;\bm{\theta}) = \log\Big( 1+\sum_{k=1}^K \exp{(\bm{X}_k^\top \bm{\theta})} \Big) - \sum_{k=1}^K \bm{X}_k^\top \bm{\theta} \bI\{Y=k\}.
\end{align*} With some calculations, the score function is given by
\begin{align*}
\bm{\psi}(\bm{X}_1,\bm{X}_2,...,\bm{X}_K,Y;\bm{\theta}) = \sum_{k=1}^K \bigg( \frac{\exp(\bm{X}_k^\top \bm{\theta})}{1 + \sum_{\kappa=1}^K \exp(\bm{X}_\kappa^\top \bm{\theta})} - \bI\{Y=k\} \bigg) \bm{X}_k
\end{align*} The first order condition $\bE\big[\bm{\psi}(\bm{X}_1,\bm{X}_2,...,\bm{X}_K,Y;\bm{\theta})\big] = \bm{0}$ provides a moment condition to solve $\bm{\theta}$, which is usually implemented using numerical optimization algorithms such as the Newton-Raphson algorithm.
\hfill \halmos
\end{example}

The prediction-powered $M$-estimator \citep{angelopoulos2023prediction} is defined as the solution to the rectified empirical risk minimization problem,
\begin{align}
\widehat{\bm{\theta}} = \argmin_{\bm{\theta} \in \rR^d} \bigg[ \frac{1}{n-s} \sum_{i=1}^{n-s} \Big(l(\bm{X}_i, Y_i; \bm{\theta}) - l(\bm{X}_i, f^{(s)}(\bm{X}_i); \bm{\theta})\Big) + \frac{1}{m} \sum_{j=1}^m l(\tilde{\bm{X}}_j, f^{(s)}(\tilde{\bm{X}}_j); \bm{\theta}) \bigg]. \label{eqn:Mestimator}
\end{align}

To simplify the exposition, we maintain the same mild regularity conditions as in \citet{angelopoulos2023ppi}, including their smoothness, moment, and consistency conditions, without restating them here.

It is easy to see that $\widehat{\bm{\theta}}$ as defined in \eqref{eqn:Mestimator} is a consistent estimator. To see this, we focus on the right-hand side of \eqref{eqn:Mestimator} when both $(n-s) \to +\infty$ and $m \to +\infty$.
\begin{multline*}
\frac{1}{n-s} \sum_{i=1}^{n-s} \Big(l(\bm{X}_i, Y_i; \bm{\theta}) - l(\bm{X}_i, f^{(s)}(\bm{X}_i); \bm{\theta})\Big) + \frac{1}{m} \sum_{j=1}^m l(\tilde{\bm{X}}_j, f^{(s)}(\tilde{\bm{X}}_j); \bm{\theta}) \\
\xrightarrow{p} \ \bE\Big[l(\bm{X}_i, Y_i; \bm{\theta})\Big] - \bE\Big[l(\bm{X}_i, f^{(s)}(\bm{X}_i); \bm{\theta})\Big] + \bE\Big[l(\tilde{\bm{X}}_i, f^{(s)}(\tilde{\bm{X}}_i); \bm{\theta})\Big] = \bE\Big[l(\bm{X}_i, Y_i; \bm{\theta})\Big],
\end{multline*} where the equality is because $\bm{X}_i$ from the small labeled dataset and $\tilde{\bm{X}}_j$ from the large unlabeled dataset are sampled from the same distribution.
In the examples above, the outcome enters the loss only through a term that is linear in $\bm{\theta}$, so the difference $l(\bm{X}, Y; \bm{\theta}) - l(\bm{X}, f^{(s)}(\bm{X}); \bm{\theta})$ is linear in $\bm{\theta}$ and the rectified objective in \eqref{eqn:Mestimator} is convex in $\bm{\theta}$. Since the population criterion has a unique minimizer, Theorem~2.7 of \citet{newey1994large} ensures that $\widehat{\bm{\theta}}$ is consistent.

Because $\widehat{\bm{\theta}}$ is consistent, we focus on the asymptotic covariance matrix of $\widehat{\bm{\theta}}$. The asymptotic covariance matrix of $\widehat{\bm{\theta}}$ is given by the sandwich variance formula.

\begin{lemma}[\citet{angelopoulos2023ppi} Theorem~1]
\label{lem:Mestimator:variance} The asymptotic variance of $\widehat{\bm{\theta}}$ is
\begin{align*}
\Var(\widehat{{\bm{\theta}}}) = \bm{H}^{-1}\bm{V}\bm{H}^{-1},
\end{align*} where
\begin{align*}
\bm{H} = & \bE\Big[\bm{H}_{\bm{\theta}} l(\bm{X},Y;\bm{\theta}^*)\Big], \\
\bm{V} = & \frac{1}{n-s} \Var\big(\bm{\psi}(\bm{X},Y;\bm{\theta}^*) - \bm{\psi}(\bm{X},f^{(s)}(\bm{X});\bm{\theta}^*)\big) + \frac{1}{m} \Var\big(\bm{\psi}(\tilde{\bm{X}},f^{(s)}(\tilde{\bm{X}});\bm{\theta}^*)\big).
\end{align*} Here, $\bm{H}_{\bm{\theta}} l(\bm{X},Y;\bm{\theta}^*)$ stands for the Hessian matrix of $l(\bm{X},Y;\bm{\theta})$ evaluated at $\bm{\theta}^*$.
\end{lemma}

In the case when $m$ is much larger than $n-s$, the second term can be ignored, and we focus our attention primarily on the first term. Note that the asymptotic variance of $\widehat{{\bm{\theta}}}$ is a covariance matrix $\Var(\widehat{{\bm{\theta}}})$.
Because covariance matrices are not totally ordered, a common strategy from the optimal experimental design literature is to choose a scalar criterion of $\Var(\widehat{{\bm{\theta}}})$ and minimize that scalarization \citep{atkinson2007optimum, fedorov2013theory, pukelsheim2006optimal, silvey2013optimal}.
We consider minimizing the trace of the covariance matrix $\tr\big(\Var(\widehat{{\bm{\theta}}})\big)$, which is usually referred to as the A-optimal objective.\footnote{An alternative is the determinant (D-optimal) scalarization, which minimizes the volume of the confidence ellipsoid; because $\bm{H}$ does not depend on $f^{(s)}$, it reduces to minimizing $\det(\Var(\bm{\psi}(\bm{X},Y;\bm{\theta}^*) - \bm{\psi}(\bm{X},f^{(s)}(\bm{X});\bm{\theta}^*)))^{1/d}/(n-s)$: a geometric average variance per dimension. The D-optimal scalarization admits the same power-law curve and allocation analysis as the A-optimal scalarization; we omit the parallel development.} Minimizing the trace of the covariance matrix corresponds to minimizing the mean squared error of the estimator vector $\bE[\|\widehat{\bm{\theta}} - \bm{\theta}^*\|_2^2]$; see \citet{boyd2004convex}.

Recall from Lemma~\ref{lem:Mestimator:variance} that the asymptotic covariance is $\bm{H}^{-1}\bm{V}\bm{H}^{-1}$. Using the cyclic property of the trace operator, we can express the total variance objective as:
\begin{align*}
\mathrm{tr}\big(\Var(\widehat{\bm{\theta}})\big)
= \mathrm{tr}\big(\bm{H}^{-1}\bm{V}\bm{H}^{-1}\big)
= \mathrm{tr}\big((\bm{H}^{-1})^2\bm{V}\big),
\end{align*} Because the Hessian term $\bm{H}$ is a population quantity that does not depend on the fine-tuned function $f^{(s)}(\cdot)$, we can think of $(\bm{H}^{-1})^2$ as a fixed weighting matrix, and our objective is
\begin{align}
\label{eqn:MestimatorTrace}
\frac{1}{n-s} \mathrm{tr}\Big((\bm{H}^{-1})^2 \Var\big(\bm{\psi}(\bm{X},Y;\bm{\theta}^*) - \bm{\psi}(\bm{X},f^{(s)}(\bm{X});\bm{\theta}^*)\big)\Big).
\end{align} We denote the numerator in \eqref{eqn:MestimatorTrace} as the scalarized variance metric $\nu(s)$
\begin{align}
\nu(s) = \mathrm{tr}\Big((\bm{H}^{-1})^2 \Var\big(\bm{\psi}(\bm{X},Y;\bm{\theta}^*) - \bm{\psi}(\bm{X},f^{(s)}(\bm{X});\bm{\theta}^*)\big)\Big). \label{eqn:scalarizedmetric}
\end{align}
 Next, we show that minimizing the objective \eqref{eqn:MestimatorTrace}, by definition, weakly dominates any alternative fine-tuning objective.
We also show in Example~\ref{exa:two-region} below that the inequality can be strict.

\begin{theorem}
\label{thm:EstimandAware}
Fix $s \in (0,n)$ and let $\mathcal{C}$ be a class of fine-tuned LLM predictors independent of the $n-s$ rectification samples and the $m$ unlabeled samples.
Assume $\bm{H}$ is non-singular.
For each $\varphi \in \mathcal{C}$, assume $\Var\big(\bm{\psi}(\bm{X},Y;\bm{\theta}^*) - \bm{\psi}(\bm{X},\varphi(\bm{X});\bm{\theta}^*)\big)$ is finite.
Let $\widehat{\bm{\theta}}(\varphi)$ denote the rectified $M$-estimator \eqref{eqn:Mestimator} with $f^{(s)}$ replaced by $\varphi$.
Let $\nu_{\varphi}(s)$ denote the scalarized variance metric \eqref{eqn:scalarizedmetric} with $f^{(s)}$ replaced by $\varphi$.
Let
\begin{align*}
\bm{\Sigma}_{\varphi} = \frac{1}{n-s}\, \bm{H}^{-1}\, \Var\big(\bm{\psi}(\bm{X},Y;\bm{\theta}^*) - \bm{\psi}(\bm{X},\varphi(\bm{X});\bm{\theta}^*)\big)\, \bm{H}^{-1}
\end{align*}
denote the leading term of the covariance $\Var(\widehat{{\bm{\theta}}})$ in Lemma~\ref{lem:Mestimator:variance} in the regime $m/(n-s) \to \infty$.

Let $\varphi^* \in \argmin_{\varphi \in \mathcal{C}} \nu_{\varphi}(s)$. Then for every $\varphi^{\circ} \in \mathcal{C}$,
\begin{align*}
\mathrm{tr}\big(\bm{\Sigma}_{\varphi^*}\big) = \frac{\nu_{\varphi^*}(s)}{n-s} \leq \frac{\nu_{\varphi^{\circ}}(s)}{n-s} = \mathrm{tr}\big(\bm{\Sigma}_{\varphi^{\circ}}\big),
\end{align*}
with equality if and only if $\varphi^{\circ}$ also minimizes $\nu_{\varphi}(s)$ over $\mathcal{C}$.
\end{theorem}

The proof of Theorem~\ref{thm:EstimandAware} can be found in Online Appendix~\ref{appsec:proof}.
In Theorem~\ref{thm:EstimandAware}, $\varphi^{\circ}$ can be the LLM predictor fine-tuned under MSE, cross-entropy, or any other conventional objective.
Recall that, in the mean estimation setup, Remark~\ref{rmk:objectives} shows that the residual variance objective is the same as the MSE objective under some conditions.
In the more general $M$-estimation setup, however, the inequality in Theorem~\ref{thm:EstimandAware} can be strict.
Intuitively, this is because the objective function is usually not aligned with the second stage rectification.

\begin{example}[A Strict Gap]
\label{exa:two-region}
Let $W$ be a $\mathrm{Bernoulli}(\frac{1}{2})$ random variable and let $Y=W$.
So with probability one half, $(W, Y) = (0, 0)$, and with probability one half, $(W, Y) = (1, 1)$.
Consider the following quantity of interest $\beta = \bE[WY] = \frac{1}{2}$.

Suppose that, instead of using a flexible LLM predictor, we restrict ourselves to predicting constants. %
If we train this constant predictor using the MSE objective, then the best constant predictor is $\frac{1}{2}$.
On the other hand, if we train this constant predictor using the scalarized variance metric in \eqref{eqn:scalarizedmetric}, then the best constant predictor is $1$.

On the scalarized variance metric, we can calculate a non-zero gap between the two constant predictors $\nu_{1} = 0 < \nu_{\frac{1}{2}} = \frac{1}{16}$. So the inequality in Theorem~\ref{thm:EstimandAware} can be strict.
The reason is that MSE objective trains the constant predictor to fit both $W=1$ and $W=0$ cases, whereas only the $W=1$ case influences the target estimand.
\hfill \halmos
\end{example}

Although Theorem~\ref{thm:EstimandAware} is specific to A-optimal scalarization, we can easily adapt it to D-optimal scalarization or other scalarizations.
In particular, in the linear regression setting of Section~\ref{sec:mest-case}, we estimate a single coefficient $\beta = \bm{c}^{\top}\bm{\theta}^*$.
Theorem~\ref{thm:EstimandAware} applies to the scalarization of $\bm{c}^{\top}\bm{\Sigma}_{\varphi}\bm{c}$, showing that the estimator obtained in this way attains the smallest asymptotic variance and Wald interval width for the reported coefficient $\beta$.

It is natural to model $\nu(s)$ using the same power-law scaling relationship $\nu(s) = a s^{-\alpha} + b$, where $\alpha, a > 0, b \geq 0$ are task-specific constants.
When the constants are known, minimizing the trace objective \eqref{eqn:MestimatorTrace} is equivalent to solving the univariate optimization problem
\begin{align*}
\min_{s \in (0,n)} \ \frac{a s^{-\alpha} + b}{n-s}.
\end{align*}
This recovers the exact form of the optimization problem in \eqref{eqn:Minimization}.
Consequently, both the optimal sample allocation rule from Theorem~\ref{thm:OptimalAllocation} and the monotonicity of Proposition~\ref{prop:monotonicity_abn} directly hold.
Proposition~\ref{prop:variance_reduction_conditions} is specific to mean estimation and no longer holds in $M$-estimation.

\subsubsection*{Inference.}
Finally, we state the inference method provided by \citet{angelopoulos2023ppi} to construct confidence intervals for $M$-estimation. We will make use of the prediction-powered $M$-estimator $\widehat{\bm{\theta}}$ defined in \eqref{eqn:Mestimator}. First, we estimate the sample covariance matrix
\begin{align*}
\widehat{\bm{V}}_{f^{(s)}} = \frac{1}{m-1} \sum_{j=1}^m \bigg(\bm{\psi}(\tilde{\bm{X}}_j,f^{(s)}(\tilde{\bm{X}}_j);\widehat{\bm{\theta}}) - \bar{\bm{\psi}}\bigg) \bigg(\bm{\psi}(\tilde{\bm{X}}_j,f^{(s)}(\tilde{\bm{X}}_j);\widehat{\bm{\theta}}) - \bar{\bm{\psi}}\bigg)^\top
\end{align*} in the large unlabeled dataset, where $\bar{\bm{\psi}} = \frac{1}{m}\sum_{j=1}^m\bm{\psi}(\tilde{\bm{X}}_j,f^{(s)}(\tilde{\bm{X}}_j);\widehat{\bm{\theta}})$. Second, we estimate the sample covariance matrix
\begin{align*}
\widehat{\bm{V}}_{Y-f^{(s)}} = \frac{1}{n-s-1} \sum_{i=1}^{n-s} \bigg(\bm{\Delta}_i - \bar{\bm{\Delta}}\bigg) \bigg(\bm{\Delta}_i - \bar{\bm{\Delta}}\bigg)^\top
\end{align*} in the small labeled dataset, where $\bm{\Delta}_i = \bm{\psi}(\bm{X}_i,Y_i;\widehat{\bm{\theta}}) - \bm{\psi}(\bm{X}_i,f^{(s)}(\bm{X}_i);\widehat{\bm{\theta}})$ and $\bar{\bm{\Delta}} = \frac{1}{n-s}\sum_{i=1}^{n-s}\bm{\Delta}_i$. Third, we estimate the sample Hessian matrix
\begin{align*}
\widehat{\bm{H}} = \frac{1}{n-s} \sum_{i=1}^{n-s} \bm{H}_{\bm{\theta}}l(\bm{X}_i,Y_i;\widehat{\bm{\theta}})
\end{align*} in the labeled dataset. Finally, we calculate
\begin{align}\nonumber
\widehat{\bm{\Sigma}} = \widehat{\bm{H}}^{-1} \bigg(\frac{1}{n-s}\widehat{\bm{V}}_{Y-f^{(s)}} + \frac{1}{m}\widehat{\bm{V}}_{f^{(s)}}\bigg) \widehat{\bm{H}}^{-1}.
\end{align} For any $\delta \in (0,1)$ and any dimension $l \in [d]$, we calculate the following confidence interval
\begin{align*}
\bigg[\widehat{\theta}_l - z_{1-\frac{\delta}{2}}\sqrt{\widehat{\bm{\Sigma}}_{ll}}, \ \widehat{\theta}_l + z_{1-\frac{\delta}{2}}\sqrt{\widehat{\bm{\Sigma}}_{ll}}\bigg].
\end{align*} Theorem~1 of \citet{angelopoulos2023ppi} suggests that the confidence interval constructed in this way is asymptotically valid.

\section{Empirical Analysis: Mean Estimation}
\label{sec:empirical}

In this section, we present empirical evidence for the mean-estimation setting. After introducing our empirical setup in Section~\ref{ssec:empirical_setup}, we conduct three empirical analyses. First, in Section~\ref{sssec:scaling_law_validation}, we assess the goodness-of-fit of the fine-tuning scaling law. Second, we test the optimal sample allocation rule in Section~\ref{sssec:optimal_allocation}. Third, we compare our proposed method to several benchmarks in Section~\ref{sssec:compare-estimators}, highlighting the variance reduction and cost saving benefits of using the variance-based loss function in fine-tuning. Section~\ref{ssec:robustness_summary} summarizes additional empirical evidence and robustness checks.

\subsection{Empirical Setup}
\label{ssec:empirical_setup}

Textual product reviews are a common source of information for assessing perceived quality in many markets. Such reviews encode rich but noisy signals about latent product attributes that are difficult to measure directly. A central challenge for firms is to aggregate these textual evaluations into reliable quality measures.

This challenge is particularly important in settings where high-quality evaluations are costly. In such settings, each additional labeled observation is expensive, and improving statistical efficiency directly reduces the number of human evaluations needed to achieve a target level of precision. As a result, variance reduction is not only the technical goal; it has direct implications for how quickly and reliably firms can form aggregate assessments of product quality under fixed evaluation budgets.

\paragraph{Dataset and Estimand.} We study this problem using the Wine Reviews dataset, consisting of wine reviews written by professional wine critics at \textit{Wine Enthusiast} magazine and publicly available on Kaggle \citep{wine_reviews_kaggle}. This dataset has also been used in prior work on prediction-powered inference and surrogate outcomes \citep{ji2025predictions}.

The dataset comprises $129{,}971$ wine reviews, each containing a free-form natural language description of sensory characteristics, such as aroma, flavor, balance, and finish, alongside structured metadata including price. After deleting duplicated reviews and records without a valid price field, we have $90{,}610$ unique reviews for our empirical analysis. The outcome variable $Y$ is an overall rating given by expert reviewers, ranging from 80 to 100 points (as the magazine only publishes reviews scoring 80 or above). This metric reflects a standardized professional rubric widely adopted in the industry. Each observation's input $\bm{X}$ has two components: the free-form review text and the wine's retail price. Two representative examples are provided in Table~\ref{tab:wine_summary_examples}.

To get the population mean of ratings, we randomly partition the dataset into two disjoint sets: one with $80{,}610$ unique reviews as the \textit{target population}, another with $10{,}000$ reviews to estimate the fine-tuning scaling law in Section~\ref{sssec:scaling_law_validation}. The population mean rating over the target population is $88.493$, and is treated as the ground truth throughout the empirical analysis.

\begin{table}[ht]
\centering
\footnotesize
\caption{Examples from the Wine Reviews dataset}
\vspace{2mm}
\begin{tabular}{l p{11cm}}
\toprule
Review & This is a walk backward after the impressive 2012. Almost impenetrably black, the flavors converge around espresso and bitter chocolate, yet the tannins have a green edge. The wine simply feels flat in the mouth with no life to it. \\
Price & \$28 \\
Rating ($Y$) & 85 \\
\midrule
Review  & This is one of the great Rieslings from the Wachau, a wonderful panoply of ripe, tropical fruit, pierced with flint, spice and minerality. It is rich and opulent, while never losing sight of the core tautness of a fine Riesling. \\
Price  & {\$70} \\
Rating ($Y$) & 95 \\
\bottomrule
\end{tabular}
\label{tab:wine_summary_examples}
\end{table}

\paragraph{PPI Estimator Construction.} In subsequent experiments in Section~\ref{sssec:optimal_allocation} and Section~\ref{sssec:compare-estimators}, for each independent replication and each budget setting with $n$ labeled samples, we draw the labeled dataset $\sN$ i.i.d.\ \emph{with replacement} from the target population (the superpopulation convention, under which the sample mean's variance scales exactly as $1/n$ and the labeled draws are i.i.d.); part of $\sN$ is used for LLM fine-tuning and the remainder for PPI rectification. All remaining reviews in the target population are treated as unlabeled data, denoted as $\sM$, and are used for PPI estimator construction.\footnote{The unlabeled set is the census excluding the drawn indices. The sample-savings identity below is exact for the i.i.d.\ labeled draws. The census term's dependence on the drawn indices is shared by every compared estimator; the reported intervals use the standard independent-sample approximation.} Importantly, the $10{,}000$ reviews used for scaling law estimation in Section~\ref{sssec:scaling_law_validation} are never used in fine-tuning or PPI, ensuring a clean separation between scaling law estimation and downstream inference.

\paragraph{LLM Fine-Tuning.}
We map each review to a $3{,}072$-dimensional embedding using the frozen \texttt{text-embedding-3-large} model and train an MLP regression head consisting of two linear layers with a ReLU activation between them to predict its rating. This setup is closely related to task-head fine-tuning and linear probing \citep{alain2016probing, kumar2022finetuning}. The mean-estimation study uses only review text; price is incorporated in the regression-coefficient case study in Section~\ref{sec:mest-case}. To obtain $f^{(s)}$, we randomly split the $s$ labeled samples allocated to fine-tuning into $80\%$ for training and $20\%$ for validation, and train the regression head using Adam. Following Section~\ref{ssec:loss_func}, we minimize the mini-batch residual variance
$\mathcal{L}_{\mathrm{var}} = \frac{1}{k}\sum_{i=1}^k (r_i-\bar r)^2$,
where $k$ is the batch size, $r_i=Y_i-f^{(s)}(\bm X_i)$, and
$\bar r=\frac{1}{k}\sum_{i=1}^k r_i$. We also examine LoRA fine-tuning of a decoder-only model as a robustness check in Online Appendix~\ref{app:lora}.

\subsection{Scaling Law Validation}
\label{sssec:scaling_law_validation}

We assess the scaling law $\Var(Y-f^{(s)}(\bm X))=as^{-\alpha}+b$ using the $10{,}000$ labeled reviews reserved for scaling-law estimation. We randomly divide these reviews into a fine-tuning pool and a shared evaluation set of $5{,}000$ reviews each. For varying allocations $s$ from the fine-tuning pool, we train models using the variance-based objective and evaluate their residual variance on the shared evaluation set. Each allocation $s$ includes the internal $80/20$ training/validation split described above.

\begin{figure}[htb!]
\centering
\setlength{\fboxrule}{0.1pt}

\begin{minipage}{0.98\linewidth}
\centering

\begin{subfigure}{0.48\linewidth}
\centering
\includegraphics[width=\linewidth]{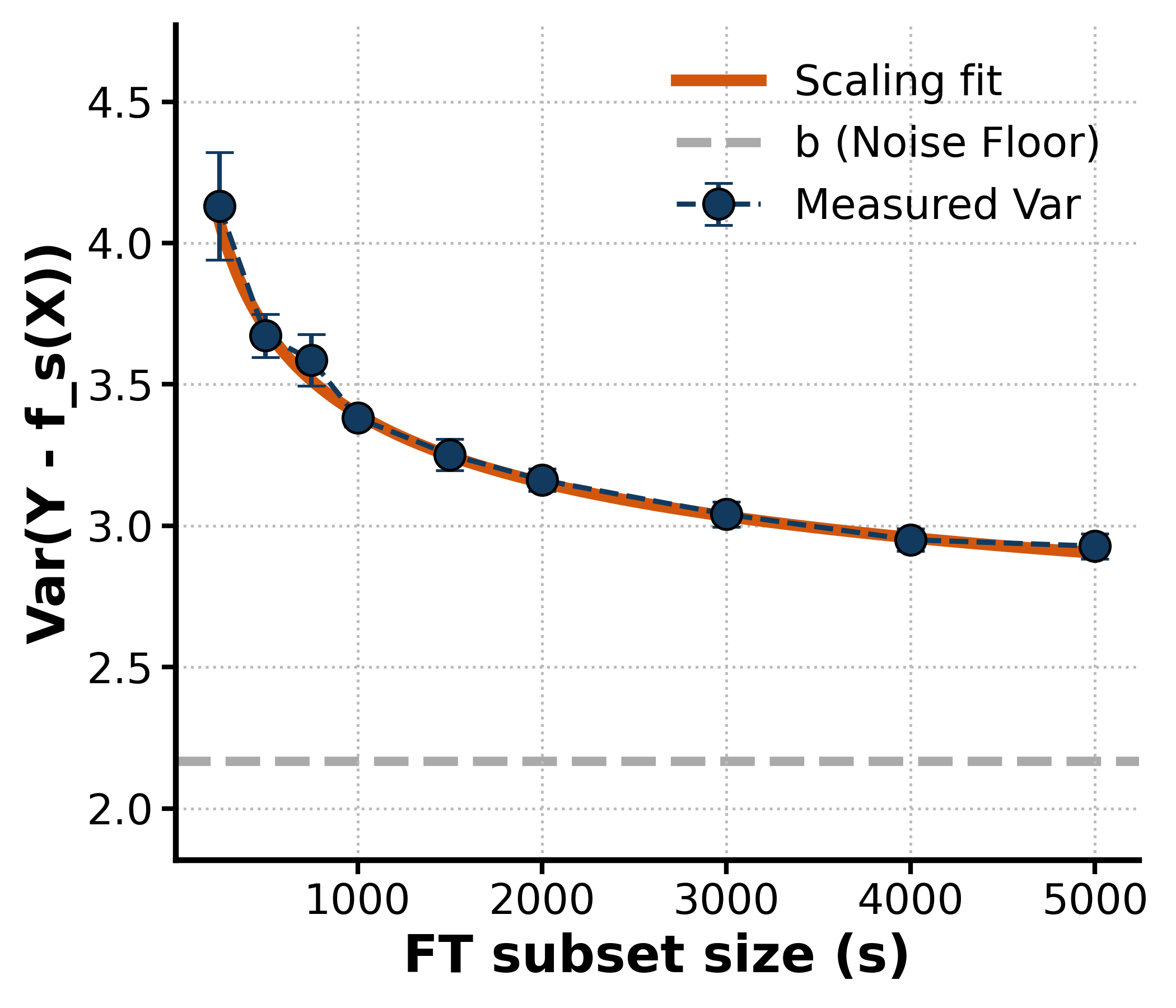}
\caption{Power-law fit in linear space}
\label{fig:wine-linear-scaling}
\end{subfigure}
\hfill
\begin{subfigure}{0.48\linewidth}
\centering
\includegraphics[width=\linewidth]{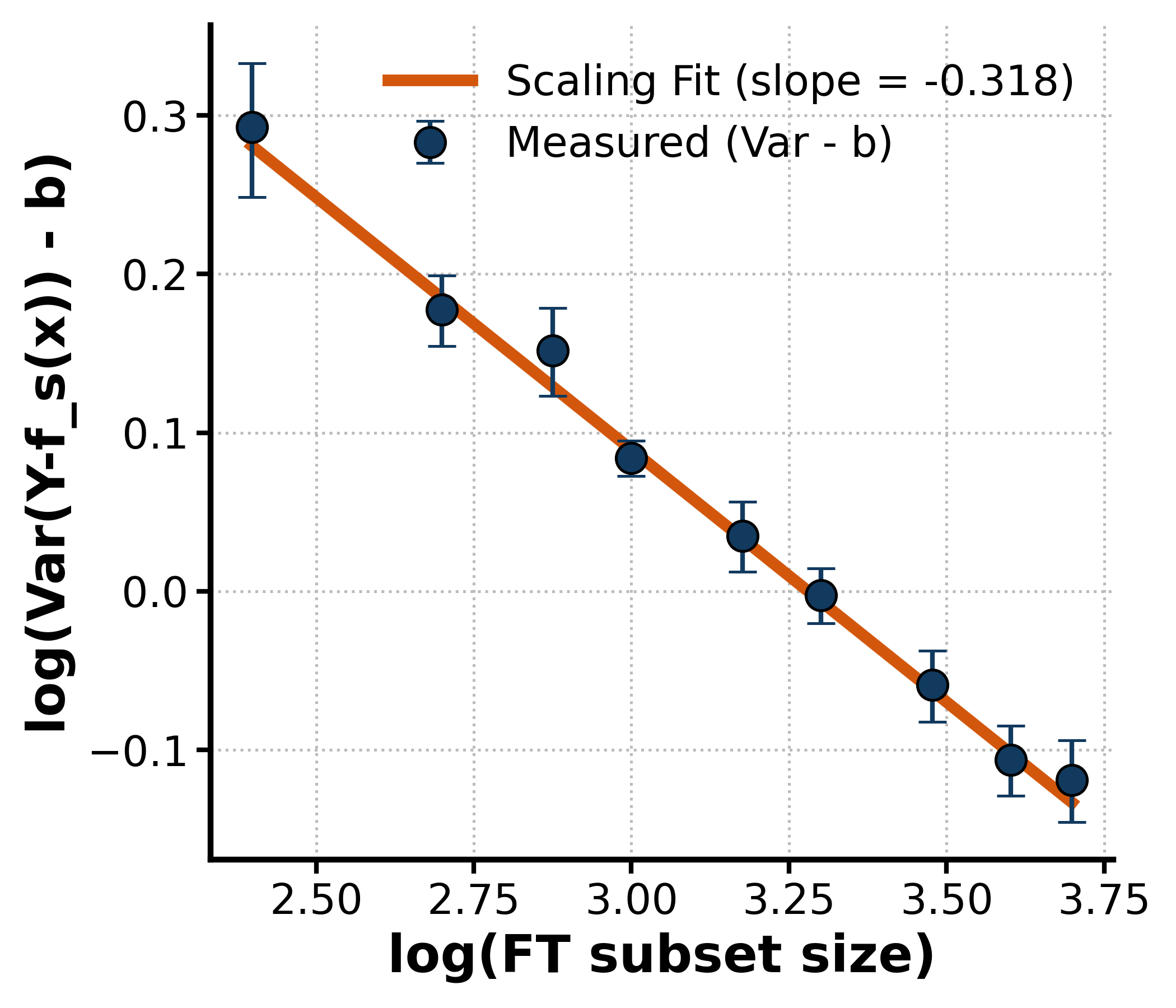}
\caption{Linear fit in log-log space}
\label{fig:wine-loglog-scaling}
\end{subfigure}
\caption{
Empirical validation of the variance-based scaling law. The estimated parameters are $\widehat a = 11.06$, $\widehat \alpha = 0.318$, and $\widehat b = 2.17$ (adjusted $R^2 = 0.999$); points are medians over ten training replicates per size with $95\%$ error bars ($1.96$ standard errors over the replicates)
}
\label{fig:wine-scaling-both}
\end{minipage}%
\end{figure}

Figure~\ref{fig:wine-scaling-both} shows that the power law closely fits the measured response. Online Appendix~\ref{appssec:robustness_check} assesses the robustness of the resulting allocation through two complementary checks. First, a hierarchical bootstrap examines sensitivity to data sampling and training randomness, showing that the implied allocation ratio $s^*/n$ remains tightly concentrated despite substantial variation in individual parameter estimates. Second, to assess practical use with limited pilot data and computation, we fit the law using a single training run at each of a few small sample sizes. These one-shot fits yield noisier allocations but incur only modest estimated efficiency losses because the allocation objective is flat near its minimum.

\subsection{Optimal Allocation of Labeled Samples}
\label{sssec:optimal_allocation}

Next, we empirically evaluate the impact of sample allocation on estimation efficiency. Our primary goal is to validate that the allocation trade-off follows the predicted pattern and that our theoretical optimal allocation rule (Theorem~\ref{thm:OptimalAllocation}) accurately identifies the empirical minimizer. We utilize the scaling law parameters $(\widehat a,\widehat\alpha,\widehat b)=(11.06, 0.318, 2.17)$ estimated above as fixed inputs to derive the theoretical optimal allocation $s^*$, which is then tested against empirical performance.

We consider an inference setting where $n=10000$ labeled samples are observed to construct the mean rating estimators. Using this labeled dataset $\sN$, we vary the allocation fraction $s/n\in\{5\%,10\%,\dots,95\%\}$ assigned to fine-tuning, while reserving the remaining $n-s$ labeled samples for PPI rectification. For each allocation ratio, we conduct $5$ independent replications of the full fine-tuning and rectification pipeline and report the average width of the resulting $95\%$ confidence intervals.

\begin{figure}[ht]
\centering
\includegraphics[width=0.8\linewidth]{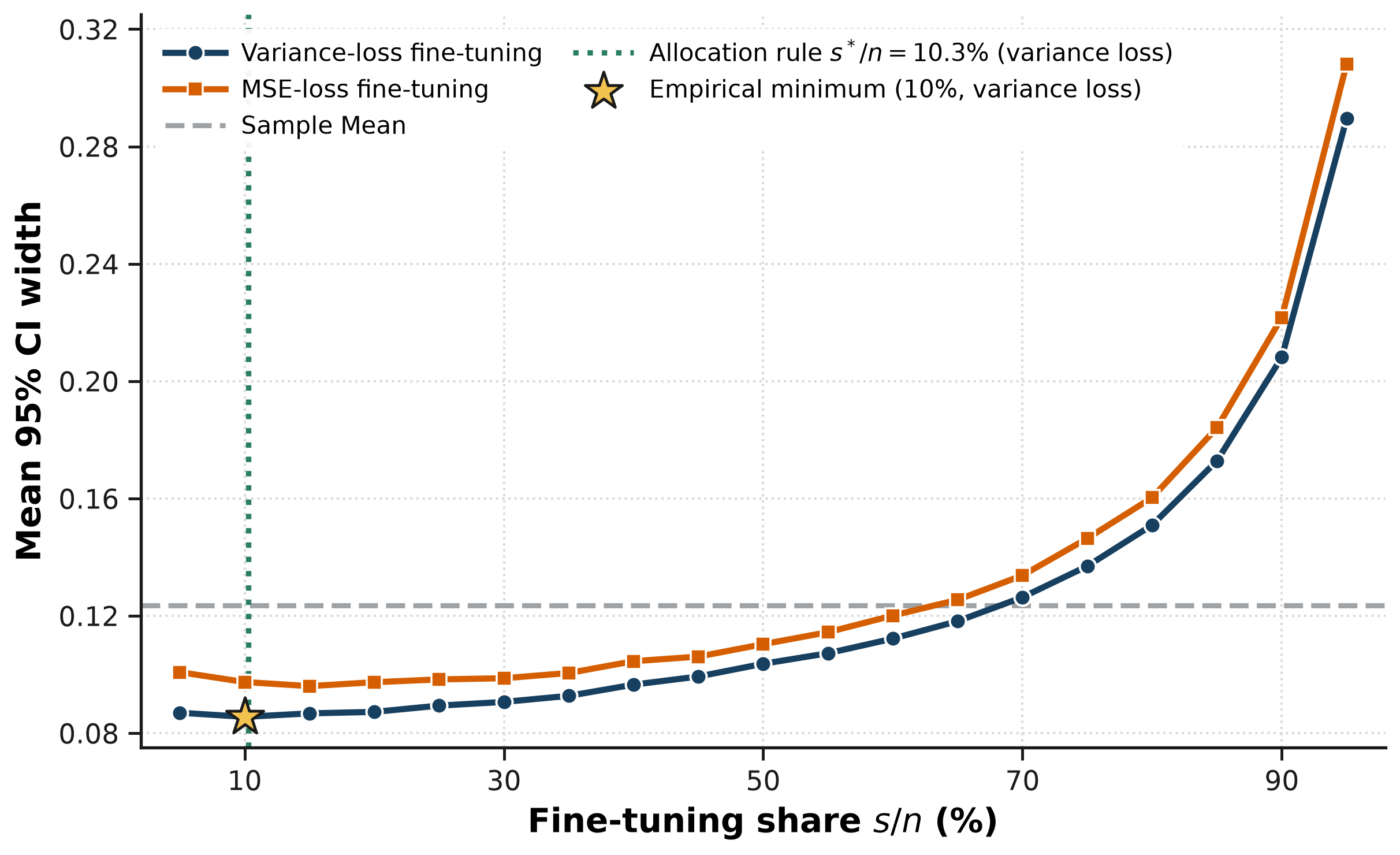}
\caption{Mean estimated $95\%$ CI width of the FT+PPI estimator across sample allocations, for the variance-loss and MSE-loss fine-tuning objectives.}
\label{fig:allocation}
\end{figure}

The results are shown in Figure~\ref{fig:allocation}, where the dashed line is the performance of the \textsc{Sample Mean} estimator. First, Figure~\ref{fig:allocation} confirms that the theoretically optimal allocation ratio $s^*/n \approx 10.3\%$ closely matches the empirical optimum, which lies at $10\%$ (evaluated at $5\%$ increments; the rule's allocation falls within the empirically flat $5\%$--$20\%$ region of the objective). In this region, the estimator achieves the narrowest confidence intervals, corresponding to the highest statistical efficiency. This observation empirically validates Theorem~\ref{thm:OptimalAllocation} and demonstrates the practical utility of our proposed variance-based scaling law.

Second, variance-loss fine-tuning yields narrower mean confidence intervals than MSE-loss fine-tuning across all evaluated allocations. Even at its empirically best allocation ($15\%$), the MSE-loss method's mean interval width is $12\%$ larger than that of the variance-loss method at the rule-selected allocation.

\begin{figure}[ht!]
\centering
\setlength{\fboxrule}{0.1pt}
\includegraphics[width=0.7\linewidth]{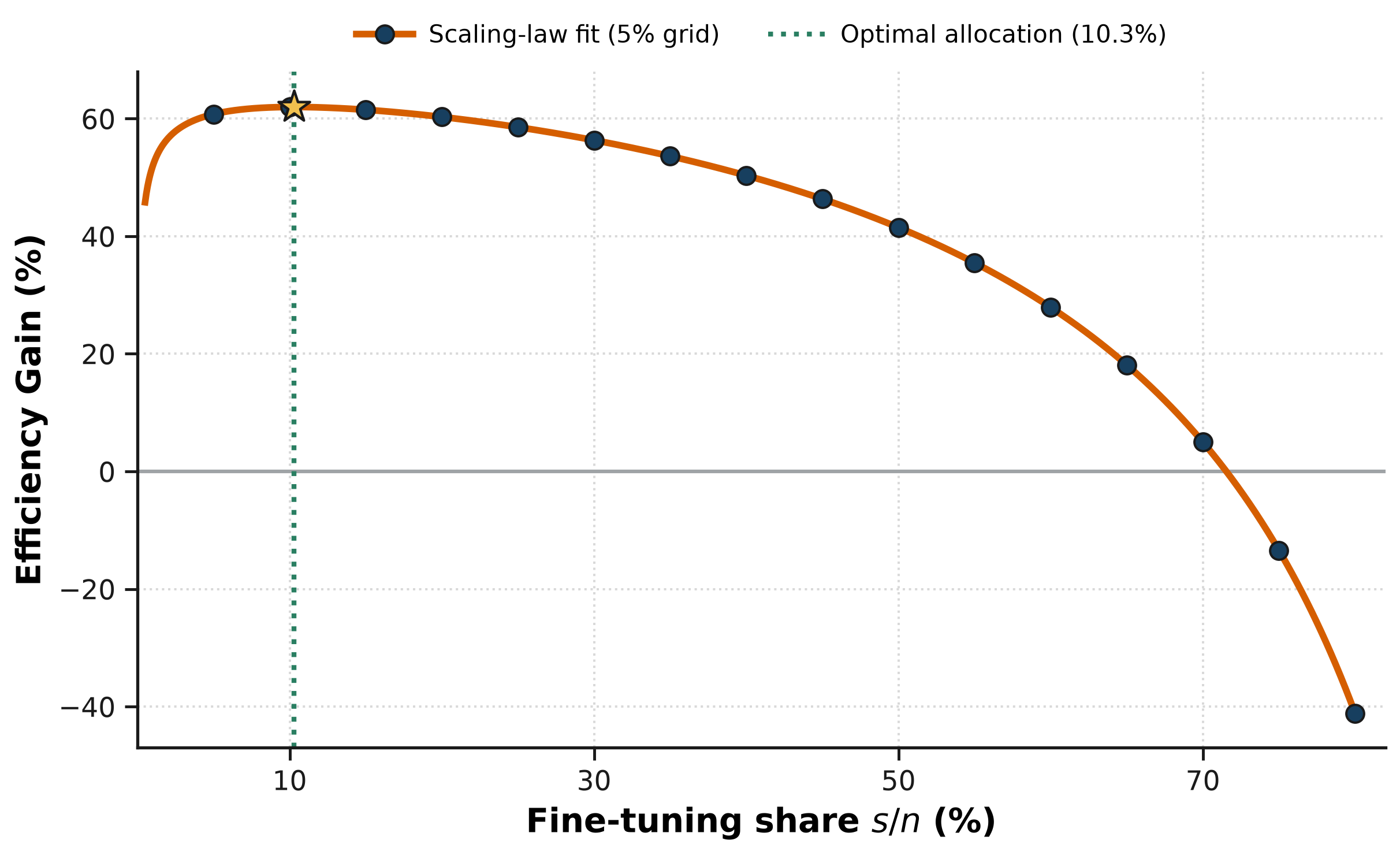}
\caption{Relative variance reduction of \textsc{FT+PPI} versus the \textsc{Sample Mean}, computed from the fitted scaling law.}
\label{fig:efficiency-gain}
\end{figure}

Third, Proposition~\ref{prop:ppi_vs_mean_R2} establishes that \textsc{FT+PPI} has lower variance than the \textsc{Sample Mean} when $R^2(s)>s/n$. Figure~\ref{fig:efficiency-gain} plots the corresponding relative variance reduction, $(R^2(s)-s/n)/(1-s/n)$. The gain remains positive over the evaluated allocations up to approximately $70\%$, beyond which allocating more labels to fine-tuning yields higher variance than the sample mean.

\subsection{Comparison of Different Estimators}
\label{sssec:compare-estimators} We evaluate the performance of our proposed framework against standard benchmarks across varying labeled data budgets, $n \in \{1000, 3000, 5000, 10000\}$. This range allows us to assess the method's effectiveness in both data-scarce and data-rich regimes.

We compare the following five estimators:
\begin{itemize}
    \item \textsc{Sample Mean:} The classical unbiased estimator averaging over  $n$ labeled samples.
    \item \textsc{FT-Only}: An MSE-trained surrogate fitted on all $n$ labels, using the same fine-tuning method as \textsc{FT+PPI (MSE Loss)}. Its average prediction over the unlabeled census is used without rectification. %
    \item \textsc{PPI-Only} (Prompting): The base model \texttt{gpt-5.4-nano} is prompted with the review text and rating range ($80$--$100$), without fine-tuning. Predictions are then rectified. Online Appendix~\ref{appsec:frontier} also examines prompting with three stronger models: \texttt{gpt-5.6-terra}, \texttt{gpt-5.6-sol}, and \texttt{gpt-5.6-luna}.
    \item \textsc{FT+PPI (MSE Loss):} A hybrid approach where the model is fine-tuned with MSE loss on a subset of data, and the remaining data is used for PPI. We select the allocation ratio by a grid search at $5\%$ intervals.
    \item \textsc{FT+PPI (Var Loss):} Our proposed method. The model is fine-tuned using the variance-based objective, and samples are split according to the theoretically optimal allocation ratio $s^*/n$.
\end{itemize}

\subsubsection*{Estimation Accuracy.} Table~\ref{tab:comparison-estimators-by-N} reports the Root Mean Squared Error (RMSE), Mean Absolute Error (MAE), and Mean Absolute Percentage Error (MAPE) for the population mean estimate.

\begin{table}[htb!]
\centering
\footnotesize
\caption{Estimation accuracy across estimators under different labeled sample sizes $n$}
\vspace{2mm}
\begin{tabular}{c l c c c}
\toprule
\textbf{$n$} & \textbf{Estimator}
& \textbf{RMSE ($\times 10^{-2}$)}
& \textbf{MAE ($\times 10^{-2}$)}
& \textbf{MAPE ($\times 10^{-2}\%$)} \\
\midrule

\multirow{5}{*}{$1000$}
& \textsc{Sample Mean}     & $10.062\;(0.432)$ & $8.018\;(0.352)$ & $9.060\;(0.397)$ \\
& \textsc{FT-Only}     & $32.462\;(0.469)$ & $31.308\;(0.496)$ & $35.380\;(0.561)$ \\
& \textsc{PPI-Only}     & \bm{$7.092\;(0.306)$} & \bm{$5.621\;(0.250)$} & \bm{$6.352\;(0.283)$} \\
& \textsc{FT+PPI (MSE Loss)}     & $9.767\;(0.432)$ & $7.653\;(0.351)$ & $8.649\;(0.397)$ \\
& \textsc{FT+PPI (Var Loss)}     & $7.155\;(0.261)$ & $5.819\;(0.241)$ & $6.575\;(0.272)$ \\
\midrule
\multirow{5}{*}{$3000$}
& \textsc{Sample Mean}     & $5.783\;(0.226)$ & $4.600\;(0.203)$ & $5.198\;(0.229)$ \\
& \textsc{FT-Only}     & $15.770\;(0.394)$ & $14.187\;(0.398)$ & $16.032\;(0.450)$ \\
& \textsc{PPI-Only}     & $4.051\;(0.154)$ & $3.316\;(0.135)$ & $3.747\;(0.152)$ \\
& \textsc{FT+PPI (MSE Loss)}     & $4.912\;(0.215)$ & $3.900\;(0.173)$ & $4.407\;(0.195)$ \\
& \textsc{FT+PPI (Var Loss)}     & \bm{$3.660\;(0.143)$} & \bm{$2.956\;(0.125)$} & \bm{$3.340\;(0.141)$} \\
\midrule
\multirow{5}{*}{$5000$}
& \textsc{Sample Mean}     & $4.574\;(0.203)$ & $3.596\;(0.163)$ & $4.064\;(0.185)$ \\
& \textsc{FT-Only}     & $11.276\;(0.347)$ & $9.689\;(0.334)$ & $10.948\;(0.377)$ \\
& \textsc{PPI-Only}     & $3.380\;(0.148)$ & $2.659\;(0.121)$ & $3.004\;(0.136)$ \\
& \textsc{FT+PPI (MSE Loss)}     & $3.560\;(0.143)$ & $2.860\;(0.123)$ & $3.231\;(0.139)$ \\
& \textsc{FT+PPI (Var Loss)}     & \bm{$3.020\;(0.125)$} & \bm{$2.398\;(0.106)$} & \bm{$2.710\;(0.120)$} \\
\midrule
\multirow{5}{*}{$10000$}
& \textsc{Sample Mean}     & $3.374\;(0.121)$ & $2.766\;(0.112)$ & $3.126\;(0.126)$ \\
& \textsc{FT-Only}     & $7.573\;(0.274)$ & $6.172\;(0.254)$ & $6.974\;(0.287)$ \\
& \textsc{PPI-Only}     & $2.473\;(0.104)$ & $1.991\;(0.085)$ & $2.250\;(0.096)$ \\
& \textsc{FT+PPI (MSE Loss)}     & $2.381\;(0.094)$ & $1.892\;(0.084)$ & $2.138\;(0.094)$ \\
& \textsc{FT+PPI (Var Loss)}     & \bm{$2.177\;(0.082)$} & \bm{$1.752\;(0.075)$} & \bm{$1.980\;(0.084)$} \\
\bottomrule
\end{tabular}
\vspace{2mm}
\begin{tablenotes}
\linespread{1}\selectfont
\item Note: Results are based on $300$ independent replications, with standard errors in parentheses. Bold indicates the lowest error for each budget and metric. Coverage of nominal $95\%$ intervals is $0.94$--$0.97$ across the rectified methods and budgets.
\end{tablenotes}
\label{tab:comparison-estimators-by-N}
\end{table}

First, \textsc{FT-Only} has the largest errors despite fine-tuning on all $n$ labels: its RMSE is $2.2$--$3.2$ times that of \textsc{Sample Mean}. Systematic negative bias accounts for $50\%$--$93\%$ of its MSE, highlighting the need for rectification when using surrogate predictions for measurement.

Second, \textsc{FT+PPI (Var Loss)} achieves the lowest estimation errors at the three larger budgets. At $n=1{,}000$, \textsc{PPI-Only} has smaller but statistically insignificant errors. Note that the main FT+PPI allocation rule considers $s>0$, while PPI-Only uses a different closed-weight model.

\subsubsection*{Sample Savings.}
Under our sampling design, the \textsc{Sample Mean}'s variance scales as $1/n$. Variance reduction,
$1-\Var(\widehat{\mu}_{\text{FT+PPI}})/\Var(\widehat{\mu}_{\textsc{Sample Mean}})$,
therefore equals the proportion of labels saved relative to this baseline at equal precision. Table~\ref{tab:sample-savings-ftppi} reports these results.

\begin{table}[htb!]
\centering
\footnotesize
\caption{Variance reductions (\%) across estimators}
\vspace{2mm}
\begin{tabular}{c c c c}
\toprule
\textbf{$n$}
& \makecell{FT+PPI (Var Loss)\\vs. \textsc{Sample Mean}}
& \makecell{FT+PPI (MSE Loss)\\vs. \textsc{Sample Mean}}
& \makecell{FT+PPI:\\Var vs. MSE Loss} \\
\midrule
$1000$  & $47.84\;(4.43)$ & $7.83\;(14.47)$ & $42.27\;(8.87)$ \\
$3000$  & $53.02\;(3.24)$ & $31.22\;(4.54)$ & $31.46\;(5.87)$ \\
$5000$  & $53.91\;(2.15)$ & $37.05\;(3.14)$ & $26.65\;(4.35)$ \\
$10000$ & $51.90\;(1.51)$ & $38.36\;(1.92)$ & $21.91\;(2.89)$ \\
\bottomrule
\end{tabular}
\vspace{2mm}
\begin{tablenotes}
\linespread{1}\selectfont
\item Note: Entries are mean percentage reductions across $300$ independent replications, with standard deviations in parentheses. Reductions are computed within each replication from the estimators' respective variance estimates.
\item Scaling-law fitting uses a separate, one-time $10{,}000$-review reserve that is not counted in $n$; Online Appendix~\ref{appsec:ramp-up} describes within-budget scaling-law estimation.
\end{tablenotes}
\label{tab:sample-savings-ftppi}
\end{table}

Table~\ref{tab:sample-savings-ftppi} shows that \textsc{FT+PPI (Var Loss)} reduces estimator variance by $48\%$--$54\%$ relative to \textsc{Sample Mean} and by $22\%$--$42\%$ relative to MSE-based fine-tuning, which highlights a nontrivial limitation of the standard fine-tuning practice. The former corresponds to requiring roughly half as many labels as the sample mean for comparable precision. At the three larger budgets, estimated variance is also $11\%$--$22\%$ lower than under \textsc{PPI-Only}.

\subsection{Additional Evidence and Robustness Checks}
\label{ssec:robustness_summary}

We conduct the following checks to assess the sensitivity of the main findings. Full implementation details and results are reported in Online Appendices~\ref{appsec:practical_implementation}--\ref{appsec:additional_empirics}.
\begin{itemize}
\item \textit{Scaling-law estimation.} To assess how estimation noise and limited pilot training affect allocation, we use bootstrap resampling and one-shot fits with a single training run at each of a few small sizes. Bootstrap allocations remain tightly concentrated, while noisier one-shot fits incur modest losses in the implied interval-width objective because it is flat near its optimum (Online Appendix~\ref{appssec:robustness_check}).

\item \textit{Outcome centering.} To assess whether a simple preprocessing step narrows the training-objective gap, we center the training targets. On this dataset, centering reduces the estimator-variance gap to $0.5\%$--$2.4\%$. However, this heuristic does not in general make MSE and residual-variance training equivalent and we construct a simple example to illustrate the gap (Online Appendix~\ref{app:mechanism}).

\item \textit{Inference design.} To assess whether label reuse or estimator-side calibration removes the training-objective gap, we examine cross-fitting and PPI++ power tuning. Variance-trained models retain lower estimated estimator variance within matched cross-fitting folds and after power tuning; cross-fitting also reduces RMSE relative to a single split, at higher training cost (Online Appendices~\ref{appssec:crossfit} and~\ref{appssec:ppi++}).

\item \textit{Additional datasets.} To assess whether the findings extend beyond wine reviews, we repeat the comparison on clothing reviews and a survey panel. Variance-loss training yields lower estimated estimator variance than MSE in both applications, although direct rectification of strong existing surrogates is more efficient at smaller budgets (Online Appendix~\ref{appsec:applications}).

\item \textit{Surrogate quality.} To examine how stronger zero-shot predictions affect the value of further fine-tuning, we compare four prompting models under a common affine calibration class and evaluate richer fine-tuning classes for \texttt{gpt-5.6-sol}. The rule selects $30$--$250$ calibration labels across the tested models and budgets, with stronger models yielding larger variance reductions relative to the sample mean; for \texttt{gpt-5.6-sol} at $n=10{,}000$, the extended rule favors affine calibration over a richer MLP (Online Appendices~\ref{appsec:frontier} and~\ref{appsec:finetuning_class}).

\item \textit{Cross-entropy training.} To compare with a common alternative for predicting integer-valued ratings, we train a classification head with cross-entropy loss. Variance-loss training yields lower RMSE at the three smaller budgets, while the two methods perform similarly at $n=10{,}000$ (Online Appendix~\ref{appsec:ce}).

\item \textit{Classical text representation baseline.} To assess whether the LLM-based pipeline improves inference over a traditional text predictor, we compare it with ridge regression on TF-IDF features. The LLM-embedding pipeline yields lower estimated estimator variance at both tested budgets (Online Appendix~\ref{appsssec:tf_idf}).

\item \textit{LoRA fine-tuning.} To examine fine-tuning beyond a prediction head on frozen embeddings, we repeat the objective comparison with LoRA. At $s=1{,}135$ and $n=10{,}000$, variance-loss training yields $11.6\%$ lower held-out residual variance in a matched comparison (Online Appendix~\ref{app:lora}).
\end{itemize}

\section{Empirical Analysis: \texorpdfstring{$M$}{M}-Estimation}
\label{sec:mest-case}

We evaluate the $M$-estimation extension through a price-rating regression application and a controlled synthetic study of when the training objectives differ.

\subsection{Empirical Setup}
\label{ssec:mest-estimator}
\label{ssec:mest-training}

Using the same target population as in Section~\ref{sec:empirical}, let $Y$ denote rating and $p$ denote log price standardized to have mean zero and unit variance. Our target estimand is the slope coefficient $\beta^*$ in the population linear projection of rating on standardized log price:
\[
(\beta_0^*,\beta^*)
=\arg\min_{\beta_0,\beta}
\bE\!\left[(Y-\beta_0-\beta p)^2\right].
\]
The population slope is $\beta^*=1.9613$.

We apply the prediction-powered regression estimator from Section~\ref{sec:M_estimation}, with sandwich variance estimates from Lemma~\ref{lem:Mestimator:variance} and Wald confidence intervals. We use $n=10{,}000$ labeled observations and $300$ independent replications. For this regression, the slope's influence weight is $p$, so the estimand-aware objective is $\Var\!\left(p[Y-f(\bm X)]\right).$
Errors at larger absolute standardized log prices therefore receive greater weight.

We use the fine-tuning specification $f^{(s)}(\bm X)=c_0+c_1p+c_2f_{\mathrm{zs}}(\bm X)$, where $f_{\mathrm{zs}}$ is the text-only zero-shot prediction from \texttt{gpt-5.6-sol}. The coefficients are fitted using $s$ labeled samples. Online Appendix~\ref{appssec:surrogate} examines a richer MLP specification using text-embedding components, price, and the zero-shot prediction, as well as a zero-shot benchmark with price included in the prompt.

We fit the slope-specific scaling law $\Var\!\left(p\left[Y-f^{(s)}(\bm X)\right]\right)=as^{-\alpha}+b$ using eleven fine-tuning sample sizes ranging from $50$ to $3{,}000$ labels. The resulting allocation rule selects $s^*=318$. We compare conventional MSE training with minimization of the empirical slope-specific variance under the same fine-tuning specification.

\subsection{Results}
\label{ssec:mest-results}

We compare four estimators.
The \textsc{OLS} benchmark uses ordinary least squares on all labeled observations.
The \textsc{PPI-Only} benchmark rectifies the zero-shot predictions of \texttt{gpt-5.6-sol},
prompted with the review text and the price.
The two \textsc{FT+PPI} estimators use the fine-tuning specification from
Section~\ref{ssec:mest-training}, differing only in the training objective.
All four estimators have access to standardized log price, although they incorporate it through
different functional forms.
Table~\ref{tab:mest-main} reports the resulting estimation accuracy.

\begin{table}[ht!]\centering
\footnotesize
\caption{Estimation accuracy for the price-rating regression coefficient at $n = 10{,}000$}
\vspace{1mm}
\begin{tabular}{lccc}
\toprule
\textbf{Estimator} & \textbf{RMSE ($\times 10^{-2}$)} & \textbf{MAE ($\times 10^{-2}$)} & \textbf{MAPE (\%)} \\
\midrule
\textsc{OLS} & $2.436\;(0.100)$ & $1.956\;(0.084)$ & $0.997\;(0.043)$ \\
\textsc{PPI-Only} & $1.679\;(0.066)$ & $1.341\;(0.058)$ & $0.684\;(0.030)$ \\
\textsc{FT+PPI (MSE Loss)} & $1.420\;(0.055)$ & $1.155\;(0.048)$ & $0.589\;(0.024)$ \\
\textsc{FT+PPI (Var Loss)} & $\bm{1.404\;(0.051)}$ & $\bm{1.145\;(0.047)}$ & $\bm{0.584\;(0.024)}$ \\
\bottomrule
\end{tabular}
\vspace{2mm}
\begin{tablenotes}
\linespread{1}\selectfont
\item Note: Results are based on $300$ independent replications, with standard errors in parentheses. Both \textsc{FT+PPI} methods use $s=318$ development labels. Coverage of nominal $95\%$ intervals is $0.96$--$0.99$. Bold indicates the lowest error in each column.
\end{tablenotes}
\label{tab:mest-main}
\end{table}

Both \textsc{FT+PPI} estimators improve on \textsc{OLS} and \textsc{PPI-Only}, with confidence intervals approximately one third narrower than those of \textsc{OLS}. The two training objectives yield nearly identical performance: their differences are small relative to replication uncertainty. This application illustrates a boundary case in which conventional and estimand-aware training nearly coincide.

\subsection{When Do the Objectives Differ? A Controlled Synthetic Study}
\label{ssec:mest-synthetic}

To examine when the objectives differ, we construct a two-region design with heterogeneous influence weights and a constrained surrogate that must trade off approximation accuracy between the regions. Conventional MSE emphasizes the larger, low-influence region, whereas estimand-aware training prioritizes errors that matter more for the slope. Across $100$ paired training trials, the latter reduces the slope-specific variance criterion by approximately $31\%$. This gap persists with a free intercept and centered targets, so it cannot be removed by an additive-level adjustment (Online Appendix~\ref{app:synthetic}).

Table~\ref{tab:mest-synthetic} reports $200$ paired downstream trials using the same $s=300$ labeled samples and power-tuned rectification procedure. Estimand-aware training yields $16.0\%$ narrower confidence intervals than MSE training, with close-to-nominal coverage. \textsc{OLS} remains more accurate in this constrained design, which isolates the effect of the training objective.

\begin{table}[ht!]
\centering
\footnotesize
\caption{Controlled synthetic study: estimation accuracy for the census slope at $n=1,200$}
\vspace{1mm}
\begin{tabular}{lccc}
\toprule
\textbf{Estimator}
& \textbf{RMSE ($\times 10^{-2}$)}
& \textbf{MAE ($\times 10^{-2}$)}
& \textbf{MAPE (\%)} \\
\midrule
\textsc{OLS}
& $2.976\;(0.148)$
& $2.394\;(0.125)$
& $3.303\;(0.173)$ \\

\textsc{FT+PPI (MSE Loss)}
& $3.839\;(0.185)$
& $3.063\;(0.164)$
& $4.227\;(0.226)$ \\

\textsc{FT+PPI (Var Loss)}
& $3.259\;(0.165)$
& $2.599\;(0.139)$
& $3.586\;(0.192)$ \\
\bottomrule
\end{tabular}

\vspace{2mm}
\begin{tablenotes}
\linespread{1}\selectfont
\item Note: Results are based on $200$ paired trials, with standard errors in parentheses. Both \textsc{FT+PPI} methods use $s=300$ labeled samples and the same power-tuned rectification procedure. Coverage of nominal $95\%$ intervals is $0.94$--$0.96$.
\end{tablenotes}
\label{tab:mest-synthetic}
\end{table}

The objective gap becomes negligible when either the surrogate can represent the two regions separately or the high-influence cluster is removed, so that the entangled feature no longer has to serve two regions and the surrogate class becomes nearly correctly specified (Online Appendix~\ref{app:synthetic}). Together with the real-data results, these variants show when estimand-aware training is valuable: when a constrained surrogate must allocate approximation accuracy across observations with different influence on the downstream estimand.

\section{Conclusions}
\label{sec:conclusion} In this paper, we propose a fine-tuning then rectification framework for leveraging LLMs as human surrogates under limited human labeled data. We show that the statistical efficiency of the downstream rectification estimator is governed primarily by the residual variance of the LLM surrogates. Consequently, we propose a variance-based fine-tuning objective that outperforms the standard fine-tuning objective of mean-squared error. Furthermore, we leverage empirical scaling laws to derive a data-driven procedure for optimally allocating scarce labeled samples between fine-tuning and rectification. Empirical results confirm that our fine-tuning then rectification framework outperforms using either fine-tuning or rectification alone, as well as using the standard fine-tuning objective of MSE.
Empirical results also confirm that our theoretically derived allocation closely matches the empirical optimum.
Beyond mean estimation, the same framework applied to the price-rating regression coefficient delivers confidence intervals narrower than \textsc{OLS}.

We view this work as an initial step toward a systematic understanding of how to jointly optimize LLM model improvement and statistical correction. We discuss a ramp-up procedure for estimating the scaling-law parameters and a procedure to handle independent but non-identical fine-tuning data in Online Appendices~\ref{appsec:ramp-up} and~\ref{sec:external_ft_extension}, respectively.
These procedures are sometimes effective in practice and offer fertile grounds for future research.
For instance, we currently model the non-identical setting simply through changes in scaling law parameters, yet more theoretical and empirical analysis of such transfer-learning dynamics remains to be explored.
Similarly, the optimality of the data-driven ramp-up strategy could be further investigated.
As LLMs become increasingly integrated into market research, developing such rigorous methodologies for combining residual variance reduction (via fine-tuning) with bias correction (via rectification) will be essential for conducting market research at scale.

\ACKNOWLEDGMENT{Part of this research was performed while the authors were visiting the Institute for Mathematical and Statistical Innovation (IMSI), which is supported by the National Science Foundation (Grant No. DMS-2425650). The authors thank X.Y. Han, Lihua Lei, Lars van der Laan, Stefan Wager, Ruohan Zhan, and Tijana Zrnic, and seminar and workshop participants at Stanford University, University of California Los Angeles, University of Chicago, University of Miami, and Washington University in Saint Louis for their insightful comments that have greatly improved this paper.}

\ECSwitch

\ECHead{Online Appendix}

\makeatletter
\renewcommand{\theHsection}{ec.\thesection}
\renewcommand{\theHsubsection}{ec.\thesubsection}
\renewcommand{\theHfigure}{ec.\thefigure}
\renewcommand{\theHtable}{ec.\thetable}

\setcounter{equation}{0}
\renewcommand{\theequation}{ec.\arabic{equation}}
\renewcommand{\theHequation}{ec.\theequation}
\makeatother

\section{Proofs of Main Results}
\label{appsec:proof}

\subsection{Proof of Theorem~\ref{thm:OptimalAllocation}}

\proof{Proof of Theorem~\ref{thm:OptimalAllocation}.} We consider the optimization problem as in \eqref{eqn:Minimization}. Denote
\begin{align*}
V(s) = \frac{a s^{-\alpha} + b}{n-s}.
\end{align*} We then find the first order derivative,
\begin{align*}
V'(s) = & \frac{-\alpha a s^{-\alpha-1} (n-s) + (a s^{-\alpha}+b)}{(n-s)^2} \\
= & \frac{- \alpha a n s^{-\alpha-1} + (\alpha+1) a s^{-\alpha} + b}{(n-s)^2}.
\end{align*} Setting $V'(s) = 0,$ we have
\begin{align}
\alpha a n s^{-\alpha-1} - (\alpha+1) a s^{-\alpha} - b = 0. \label{eqn:FOC}
\end{align}

Let $h(s) = \alpha a n s^{-\alpha-1} - (\alpha+1) a s^{-\alpha}.$ Then we know $h'(s) = -\alpha (\alpha+1)\big(\frac{n}{s}-1\big)as^{-\alpha-1} < 0$. This means that the left-hand side of \eqref{eqn:FOC} is monotone decreasing. Note that when $s \to 0^+$, the left-hand side of \eqref{eqn:FOC} can be expressed as $a s^{-\alpha} \big(\frac{\alpha n}{s} - \alpha - 1\big) - b \to +\infty$. When $s = n$, the left-hand side of \eqref{eqn:FOC} can be expressed as $-an^{-\alpha}-b < 0$. So there exists a unique solution $s^*$ between $0$ and $n$ that solves equation \eqref{eqn:FOC}.
\hfill \halmos
\endproof

\subsection{Proof of Proposition~\ref{prop:monotonicity_abn}}

\proof{Proof of Proposition~\ref{prop:monotonicity_abn}.}

For clarity, write $s^* = s^*(a,b,\alpha,n)$ for the unique solution in $(0,n)$ to
\begin{align}
F(s;a,b,\alpha,n)
:=\alpha a n s^{-\alpha-1} - (\alpha+1)a s^{-\alpha} - b = 0,
\label{eq:FOC}
\end{align} where $a>0$, $b\ge 0$, $\alpha>0$, and $n>0$. Differentiating with respect to $s$, we have,
\[
\frac{\partial F}{\partial s}
= a\alpha(\alpha+1)s^{-\alpha-2}(s-n) < 0 \quad\text{for all }s\in(0,n),
\]
so $F(\cdot)$ is strictly decreasing and $s^*$ is well-defined and unique.

Rewriting the first-order condition \eqref{eq:FOC} gives
\begin{align}
\alpha n
= (\alpha+1)s^* + \frac{b}{a}\,(s^*)^{\alpha+1},
\qquad\text{equivalently}\qquad
n = \frac{\alpha+1}{\alpha}s^* + \frac{b}{a\alpha}(s^*)^{\alpha+1}.
\label{eq:n_identity}
\end{align}

Differentiate $F$ with respect to $a$, $b$, and $n$:
\[
\frac{\partial F}{\partial a}
= \alpha n s^{-\alpha-1} - (\alpha+1)s^{-\alpha},\qquad
\frac{\partial F}{\partial b} = -1,\qquad
\frac{\partial F}{\partial n} = \alpha a s^{-\alpha-1}.
\]

At $s=s^*$, use $F(s^*;a,b,\alpha,n)=0$ to rewrite
\[
b
= \alpha a n (s^*)^{-\alpha-1} - (\alpha+1)a (s^*)^{-\alpha}
= a\frac{\partial F}{\partial a}(s^*),
\]
so $\partial F/\partial a(s^*) = b/a \ge 0$.

By the implicit function theorem, for any parameter $\theta\in\{a,b,n\}$,
\[
\frac{\partial s^*}{\partial \theta}
= -\frac{\partial F/\partial \theta}{\partial F/\partial s}\Bigg|_{s=s^*}.
\]

Since $\partial F/\partial s<0$ on $(0,n)$, we obtain:

\smallskip
\noindent\emph{(i) Dependence on $a$.}
\[
\frac{\partial s^*}{\partial a}
= -\frac{(b/a)}{\partial F/\partial s}
\begin{cases}
>0,& b>0,\\[2pt]
=0,& b=0,
\end{cases}
\]
so $s^*$ is nondecreasing in $a$ and strictly increasing when $b>0$. The same sign carries over to $s^*/n$ since $n$ is fixed in this derivative.

\smallskip
\noindent\emph{(ii) Dependence on $b$.}
\[
\frac{\partial s^*}{\partial b}
= -\frac{-1}{\partial F/\partial s}
= \frac{1}{\partial F/\partial s} < 0,
\]
so $s^*$ and $s^*/n$ are strictly decreasing in $b$.

\smallskip
\noindent\emph{(iii) Dependence on $n$.}
\[
\frac{\partial s^*}{\partial n}
= -\frac{\alpha a (s^*)^{-\alpha-1}}{\partial F/\partial s} >0,
\]
so $s^*$ is strictly increasing in $n$.

\smallskip
\noindent\emph{(iv) Special case $b=0$.} If $b=0$, \eqref{eq:n_identity} simplifies to
\[
\alpha n = (\alpha+1)s^*
\quad\Rightarrow\quad
\frac{s^*}{n} = \frac{\alpha}{\alpha+1},
\]
which is exactly constant in $a$ and $n$.

\smallskip
\noindent\emph{(v) Dependence of $s^*/n$ on $n$.} Starting from \eqref{eq:n_identity}, write
\[
n = c_1 s^* + c_2 (s^*)^{\alpha+1},
\qquad
c_1 := \frac{\alpha+1}{\alpha}>0,\quad
c_2 := \frac{b}{a\alpha}\ge 0.
\]
Differentiate both sides with respect to $n$ (treating $s^*=s^*(n)$):
\[
1
= \left(c_1 + c_2(\alpha+1)(s^*)^{\alpha}\right)
\frac{\mathrm d s^*}{\mathrm d n},
\]
so
\[
\frac{\mathrm d s^*}{\mathrm d n}
= \frac{1}{c_1 + c_2(\alpha+1)(s^*)^{\alpha}}
= \frac{a\alpha}{(a + b(s^*)^{\alpha})(\alpha+1)}.
\]

Now
\[
\frac{\mathrm d}{\mathrm d n}\Big(\frac{s^*}{n}\Big)
= \frac{n\,\mathrm d s^*/\mathrm d n - s^*}{n^2}.
\]
Using $n = c_1 s^* + c_2 (s^*)^{\alpha+1}$ from \eqref{eq:n_identity}, a short calculation yields
\[
n\frac{\mathrm d s^*}{\mathrm d n} - s^*
= -\,\frac{\alpha b (s^*)^{\alpha+1}}{(a + b(s^*)^{\alpha})(\alpha+1)}.
\]
Therefore, for $b>0$,
\[
\frac{\mathrm d}{\mathrm d n}\Big(\frac{s^*}{n}\Big)
= -\,\frac{\alpha b (s^*)^{\alpha+1}}{n^2 (a + b(s^*)^{\alpha})(\alpha+1)} < 0,
\]
so $s^*/n$ is strictly decreasing in $n$.

Moreover, since $a + b(s^*)^{\alpha} \ge a$ and $s^*<n$, we bound
\[
\Bigg|\frac{\mathrm d}{\mathrm d n}\Big(\frac{s^*}{n}\Big)\Bigg|
\le
\frac{\alpha b (s^*)^{\alpha+1}}{n^2 a (\alpha+1)}
\le
\frac{\alpha b}{\alpha+1}\,\frac{n^{\alpha+1}}{n^2 a}
=
\frac{\alpha}{\alpha+1}\,\frac{b}{a}\,n^{\alpha-1},
\]
which is \eqref{eq:dx_dn_bound}.

Finally, we consider the limit as $n \to \infty$ for $b>0$. Dividing \eqref{eq:n_identity} by $n$ gives
\[
1 = \frac{\alpha+1}{\alpha}\frac{s^*}{n} + \frac{b}{a\alpha}\frac{(s^*)^{\alpha+1}}{n}.
\]
Let $L = \lim_{n\to\infty} s^*/n$. Since $s^*/n \in (0,1)$, $L$ must be finite. Suppose for the sake of contradiction that $L > 0$. Then as $n \to \infty$, $s^* \to \infty$ linearly with $n$. The term $\frac{(s^*)^{\alpha+1}}{n} = \frac{s^*}{n} (s^*)^\alpha$ behaves like $L \cdot (Ln)^\alpha$, which tends to $\infty$ (since $\alpha > 0$). This implies the right-hand side of the equation diverges to infinity while the left-hand side is fixed at 1, a contradiction. Thus, we must have $L=0$, i.e., $\lim_{n \to \infty} s^*/n = 0$.

\hfill \halmos
\endproof

\subsection{Proof of Proposition~\ref{prop:ppi_vs_mean_R2}}
\proof{Proof of Proposition~\ref{prop:ppi_vs_mean_R2}.} Under the infinite-$m$ approximation, the variance of the estimator is
\[
\Var(\widehat{\mu})
= \frac{1}{n-s}\Var\!\left(Y-f^{(s)}(\bm{X})\right).
\]
Comparing this to the variance of the sample mean:
\[
\Var(\widehat{\mu})<\Var(\widehat{\mu}_{\mathrm{mean}})
\quad\Longleftrightarrow\quad
\frac{\Var\!\left(Y-f^{(s)}(\bm{X})\right)}{n-s}
<\frac{\Var(Y)}{n}.
\]
Multiplying both sides by $(n-s)/\Var(Y)$ (noting $n-s > 0$) yields
\[
\frac{\Var\!\left(Y-f^{(s)}(\bm{X})\right)}{\Var(Y)}
< \frac{n-s}{n} = 1-\frac{s}{n}.
\]
Recalling the definition $1-R^{2}(s) = \Var(Y-f^{(s)}(\bm{X}))/\Var(Y)$, the inequality becomes
\[
1-R^{2}(s) < 1-\frac{s}{n}
\quad\Longleftrightarrow\quad
R^{2}(s) > \frac{s}{n}.
\]
\hfill \halmos
\endproof

\subsection{Proof of Proposition~\ref{prop:variance_reduction_conditions}}

\proof{Proof of Proposition~\ref{prop:variance_reduction_conditions}.}

\textit{(i)} Under the asymptotic assumption ($m \to \infty$), we have $\Var(\widehat\mu_{\mathrm{FT+PPI}}) = (a s^{-\alpha}+b)/(n-s)$. The condition $\Var(\widehat\mu_{\mathrm{FT+PPI}}) < \Var(\widehat\mu_{\mathrm{mean}})$ is equivalent to
\[
\frac{a s^{-\alpha}+b}{n-s} < \frac{\sigma^{2}}{n}
\quad\Longleftrightarrow\quad
n(a s^{-\alpha}+b) < \sigma^2(n-s)
\quad\Longleftrightarrow\quad
q(s) > 0.
\]
Thus, an improvement exists if and only if the maximum of $q(s)$ over the interval $(0, n)$ is positive.

\textit{(ii)} To derive the explicit condition, we analyze the extrema of $q(s)$. Differentiating $q(s)$ with respect to $s$:
\[
q'(s) = -\sigma^{2} + a\alpha n s^{-\alpha-1}, \qquad
q''(s) = -a\alpha(\alpha+1)n s^{-\alpha-2}.
\]
Since $a, \alpha, n, s > 0$, we have $q''(s) < 0$ for all $s > 0$, implying $q(s)$ is strictly concave. The unique unconstrained maximizer $s_0$ is found by setting $q'(s) = 0$:
\[
a\alpha n s^{-\alpha-1} = \sigma^{2}
\quad\Longrightarrow\quad
s_{0} = \left(\frac{a\alpha n}{\sigma^{2}}\right)^{\!1/(\alpha+1)}.
\]
Since $q$ is strictly concave, the condition $\sup_{s \in (0,n)} q(s) > 0$ is satisfied if and only if the global maximum is positive, i.e., $q(s_0) > 0$ (noting that if $s_0 \ge n$, strict concavity with $q'(s_0)=0$ implies $q$ is \emph{increasing} on $(0,n)$, so $\sup_{(0,n)} q = q(n^-)$; since $q(n) = -\text{(positive)} < 0$ in that regime, feasibility again reduces to the stated condition, which forces $s_0 < n$).

Using the first-order condition $a n s_{0}^{-\alpha} = (\sigma^{2}/\alpha)s_{0}$, we evaluate the peak value:
\[
\begin{aligned}
q(s_{0})
&= \sigma^{2}n - \sigma^{2}s_{0} - n\bigl(a s_{0}^{-\alpha}+b\bigr) \\
&= \sigma^{2}n - \sigma^{2}s_{0} - \frac{\sigma^{2}}{\alpha}s_{0} - n b \\
&= \sigma^{2}n - \sigma^{2}s_{0}\left(1+\frac{1}{\alpha}\right) - n b.
\end{aligned}
\]
The condition $q(s_{0}) > 0$ is therefore equivalent to
\[
\sigma^{2}n - n b > \sigma^{2}s_{0}\left(1+\frac{1}{\alpha}\right)
\quad\Longleftrightarrow\quad
1 - \frac{b}{\sigma^{2}} > \frac{s_{0}}{n}\left(1+\frac{1}{\alpha}\right).
\]
Substituting $s_{0}/n = (a\alpha n^{-\alpha}/\sigma^{2})^{1/(\alpha+1)}$ yields the final condition:
\[
\frac{b}{\sigma^{2}} < 1 - \left(1+\frac{1}{\alpha}\right) \left(\frac{a\alpha n^{-\alpha}}{\sigma^{2}}\right)^{\!1/(\alpha+1)}.
\]
\hfill \halmos
\endproof

\subsection{Proof of Lemma~\ref{lem:Mestimator:variance}}

\proof{Proof of Lemma~\ref{lem:Mestimator:variance}.} Denote $\bm{\psi}(\bm{X},Y;\bm{\theta}) = \nabla_{\bm{\theta}} l(\bm{X},Y;\bm{\theta})$. Denote
\begin{align*}
\bm{\eta}(\bm{\theta}) = \frac{1}{n-s} \sum_{i=1}^{n-s} \Big(\bm{\psi}(\bm{X}_i, Y_i; \bm{\theta}) - \bm{\psi}(\bm{X}_i, f^{(s)}(\bm{X}_i); \bm{\theta})\Big) + \frac{1}{m} \sum_{j=1}^m \bm{\psi}(\tilde{\bm{X}}_j, f^{(s)}(\tilde{\bm{X}}_j); \bm{\theta}).
\end{align*}

Then the first order condition of \eqref{eqn:Mestimator} can be written as
\begin{align*}
\bm{\eta}(\widehat{\bm{\theta}}) = \frac{1}{n-s} \sum_{i=1}^{n-s} \Big(\bm{\psi}(\bm{X}_i, Y_i; \widehat{\bm{\theta}}) - \bm{\psi}(\bm{X}_i, f^{(s)}(\bm{X}_i); \widehat{\bm{\theta}})\Big) + \frac{1}{m} \sum_{j=1}^m \bm{\psi}(\tilde{\bm{X}}_j, f^{(s)}(\tilde{\bm{X}}_j); \widehat{\bm{\theta}}) = 0.
\end{align*} We can apply the Taylor's expansion on $\bm{\eta}(\widehat{\bm{\theta}})$ around the true parameter $\bm{\theta}^*$, such that
\begin{align*}
\widehat{\bm{\theta}} - \bm{\theta}^* \approx -\Big[\bm{J}_{\bm{\eta}}(\bm{\theta}^*)\Big]^{-1} \bm{\eta}(\bm{\theta}^*),
\end{align*} where $\bm{J}_{\bm{\eta}}(\bm{\theta}^*)$ stands for the Jacobian matrix of $\bm{\eta}(\cdot)$ evaluated at $\bm{\theta}^*$. Applying the Delta method, we know that the asymptotic variance of $\widehat{\bm{\theta}}$ is
\begin{align*}
\Var(\widehat{\bm{\theta}}) = \bE\big[\bm{J}_{\bm{\eta}}(\bm{\theta}^*)\big]^{-1} \Var\Big(\bm{\eta}(\bm{\theta}^*)\Big) \bE\big[\bm{J}_{\bm{\eta}}(\bm{\theta}^*)\big]^{-1}.
\end{align*}

Note that
\begin{multline*}
\bE\big[\bm{J}_{\bm{\eta}}(\bm{\theta}^*)\big] = \bE\Big[\bm{H}_{\bm{\theta}} l(\bm{X},Y;\bm{\theta}^*) - \bm{H}_{\bm{\theta}} l(\bm{X},f^{(s)}(\bm{X});\bm{\theta}^*) + \bm{H}_{\bm{\theta}} l(\tilde{\bm{X}},f^{(s)}(\tilde{\bm{X}});\bm{\theta}^*)\Big] \\
= \bE\Big[\bm{H}_{\bm{\theta}} l(\bm{X},Y;\bm{\theta}^*)\Big],
\end{multline*} where $\bm{H}_{\bm{\theta}} l(\bm{X},Y;\bm{\theta}^*)$ stands for the Hessian matrix of $l(\bm{X},Y;\bm{\theta})$ evaluated at $\bm{\theta}^*$, and the second equality is because $\bm{X}_i$ from the small labeled dataset and $\tilde{\bm{X}}_j$ from the large unlabeled dataset are sampled from the same distribution.

Note also that,
\begin{align*}
\Var\Big(\bm{\eta}(\bm{\theta}^*)\Big) = \frac{1}{n-s} \Var\big(\bm{\psi}(\bm{X},Y;\bm{\theta}^*) - \bm{\psi}(\bm{X},f^{(s)}(\bm{X});\bm{\theta}^*)\big) + \frac{1}{m} \Var\big(\bm{\psi}(\tilde{\bm{X}},f^{(s)}(\tilde{\bm{X}});\bm{\theta}^*)\big).
\end{align*} Because we have a large unlabeled dataset, the second term is approximately zero when $m$ is much larger than $n-s$.

Putting the above together, the asymptotic distribution of $\widehat{\bm{\theta}}$ is
\begin{align*}
\widehat{\bm{\theta}} \ \sim \ \cN(\bm{\theta}^*, \bm{H}^{-1}\bm{V}\bm{H}^{-1}),
\end{align*} where
\begin{align*}
\bm{H} = & \bE\Big[\bm{H}_{\bm{\theta}} l(\bm{X},Y;\bm{\theta}^*)\Big], \\
\bm{V} = & \frac{1}{n-s} \Var\big(\bm{\psi}(\bm{X},Y;\bm{\theta}^*) - \bm{\psi}(\bm{X},f^{(s)}(\bm{X});\bm{\theta}^*)\big) + \frac{1}{m} \Var\big(\bm{\psi}(\tilde{\bm{X}},f^{(s)}(\tilde{\bm{X}});\bm{\theta}^*)\big)
\end{align*}
\hfill \halmos
\endproof

\subsection{Proof of Theorem~\ref{thm:EstimandAware}}

\proof{Proof of Theorem~\ref{thm:EstimandAware}.}
 Fix $\varphi \in \mathcal{C}$.
Conditional on the fine-tuning data and the training randomness, $\varphi$ is a fixed function independent of the $n-s$ rectification samples and the $m$ unlabeled samples, so Lemma~\ref{lem:Mestimator:variance} applies with $\varphi$ in place of $f^{(s)}$ and gives
\begin{align*}
\Var\big(\widehat{\bm{\theta}}(\varphi)\big) = \bm{H}^{-1} \bigg( \frac{1}{n-s} \Var\big(\bm{\psi}(\bm{X},Y;\bm{\theta}^*) - \bm{\psi}(\bm{X},\varphi(\bm{X});\bm{\theta}^*)\big) + \frac{1}{m} \Var\big(\bm{\psi}(\tilde{\bm{X}}, \varphi(\tilde{\bm{X}}); \bm{\theta}^*)\big) \bigg) \bm{H}^{-1}.
\end{align*}
In the regime $m/(n-s) \to \infty$, the second term vanishes.

The cyclic property of the trace gives
\begin{align*}
\mathrm{tr}\big(\bm{\Sigma}_{\varphi}\big) = \frac{1}{n-s}\, \mathrm{tr}\Big((\bm{H}^{-1})^2 \Var\big(\bm{\psi}(\bm{X},Y;\bm{\theta}^*) - \bm{\psi}(\bm{X},\varphi(\bm{X});\bm{\theta}^*)\big) \Big) = \frac{\nu_{\varphi}(s)}{n-s},
\end{align*}
so $\varphi \mapsto \mathrm{tr}(\bm{\Sigma}_{\varphi})$ attains its minimum over $\mathcal{C}$ exactly at the minimizer of $\nu_{\varphi}(s)$.
The inequality and its equality condition follow from the definition of $\varphi^*$.
\hfill \halmos
\endproof

\section{Practical Implementation and Robustness of the Allocation Rule}
\label{appsec:practical_implementation}
This section examines the practical implementation of the allocation rule when the scaling-law
parameters are unknown. We first assess the stability of the fitted response curve and the allocation
it implies, and then discuss how to combine the optimal sample allocation with cross-fitting type sample splitting, and finally describe a data-driven ramp-up procedure for estimating the curve from limited
pilot data.

\subsection{Scaling-Law Robustness and Allocation Stability}
\label{appssec:robustness_check}

This subsection evaluates the stability of the scaling-law-based allocation rule through
two complementary analyses. We first quantify sensitivity to data sampling and training randomness
using a hierarchical bootstrap, and then examine a more data-limited setting in which the response
curve is fitted from only a small number of early development sizes. At the smallest deployment budget, $n=1{,}000$, the allocation rule selects $s^*=141$, slightly below the $250$-label floor of the measured development grid; the corresponding criterion value therefore relies on a short extrapolation of the fitted scaling law rather than interpolation. At all larger budgets, the rule-selected allocations lie within the measured range.

\subsubsection{Bootstrap Robustness of the Scaling-Law Fit}
\label{appssec:scaling_bootstrap}

Let $\omega$ denote any scalar quantity derived from the scaling-law fit, such as
$a$, $\alpha$, $b$, or the implied allocation ratio $s^*/n$. We decompose the total variability of $\omega$ using the Law of Total Variance:
\begin{equation}
\Var(\omega) = \underbrace{\Var_D\bigl(\mathbb{E}_{\epsilon}[\omega \mid D]\bigr)}_{\text{Data Sampling}} + \underbrace{\mathbb{E}_D\bigl(\Var_{\epsilon}(\omega \mid D)\bigr)}_{\text{Training Randomness}},
\end{equation} where $D$ represents the specific labeled dataset split and $\epsilon$ captures stochasticity arising from model initialization and optimization. The first term quantifies sensitivity to the specific sampling of the fine-tuning data, while the second term captures variability induced by stochastic optimization when the dataset is held fixed.

We approximate this decomposition using a two-level resampling design. Specifically, from the held-out pool we first draw $B_D=10$ independent datasets $\{D_1,\ldots,D_{10}\}$ of size $5{,}000$. For each $D_j$, we rerun the fine-tuning and scaling-law fitting under $B_\epsilon=3$ different random seeds, yielding $\omega_{j,k}$ for $k=1,\ldots,3$ and a total of $B_D\times B_\epsilon=30$ fitted outcomes. To form the reported robustness intervals, we apply a \emph{hierarchical} bootstrap that respects the two-level design: each of $1{,}200$ draws resamples the $10$ datasets with replacement and then the seeds within each selected dataset, and recomputes the summary statistic (median parameters and the implied $s^*/n$). The $95\%$ robustness interval is given by the 2.5th and 97.5th percentiles of the resulting distribution. The law-of-total-variance decomposition attributes $18\%$--$46\%$ of each quantity's variability to data sampling and the remainder to training randomness (the allocation ratio is the most data-limited at $46\%$).

\begin{figure}[ht]
\centering
\includegraphics[width=0.6\linewidth]{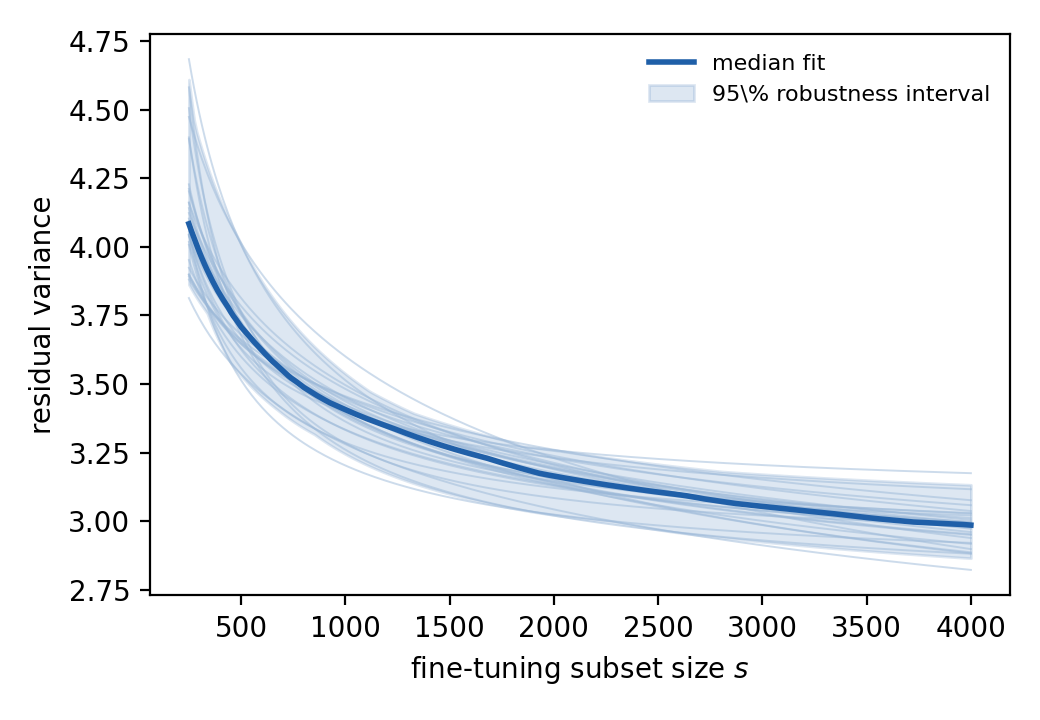}
\caption{Bootstrap robustness of the scaling-law fit based on $5{,}000$-sample development pools. Light curves correspond to individual bootstrap replicates, the solid line denotes the median fit, and the shaded region indicates the central $95\%$ robustness interval.}
\label{fig:scaling_robustness}
\end{figure}

Figure~\ref{fig:scaling_robustness} visualizes the robustness of the fitted scaling law under bootstrap resampling. Each light curve corresponds to one bootstrap replicate, while the solid line represents the median scaling law fit across replicates. The shaded region indicates the central $95\%$ robustness interval. Despite noticeable dispersion in the individual parameter estimates $(a,\alpha,b)$, the fitted curves remain tightly clustered across the full range of fine-tuning subset sizes. This demonstrates a high degree of robustness --- the fitted power law captures a stable empirical relationship between development size and residual variance in this setting.

The results are further quantified in Table~\ref{tab:scaling_robustness}, which summarizes the bootstrap robustness of the estimated parameters and the implied optimal allocation ratio across $1{,}200$ replicates. While individual parameters $(a, \alpha, b)$ show substantial dispersion, the optimal allocation ratio $s^*/n$ (computed at $n = 10{,}000$) remains tightly concentrated --- the decision-relevant quantity, the implied allocation ratio, is considerably more stable than the individual parameter estimates. Its median of $10.4\%$ sits next to the full-grid fit's $10.3\%$; the allocation objective is flat across this range, so either allocation costs under $1\%$ of interval width.

\begin{table}[ht]
\centering
\scriptsize
\caption{Robustness of the scaling-law fit}
\label{tab:scaling_robustness}
\vspace{2mm}
\begin{tabular}{l c c}
\toprule
\textbf{Quantity} & \textbf{Median} & \textbf{95\% Robustness Interval} \\
\midrule
Scaling parameter $a$      & $14.265$ & $[10.057,\;63.832]$ \\
Scaling parameter $\alpha$ & $0.391$  & $[0.251,\;0.531]$ \\
Variance floor $b$         & $2.338$  & $[0.990,\;2.434]$ \\
\midrule
Optimal allocation $s^*/n$ & $0.104$  & $[0.094,\;0.107]$ \\
\bottomrule
\end{tabular}\par
\vspace{1mm}
\begin{tablenotes}[flushleft]
\linespread{1}\selectfont
\item Note: Hierarchical bootstrap with $1{,}200$ draws from $10$ datasets and three seeds per dataset; $s^*/n$ is evaluated at $n=10{,}000$.
\end{tablenotes}
\end{table}

These findings provide strong empirical evidence that the scaling law offers a robust foundation for allocation decisions. The stability of $s^*/n$ confirms that the resulting allocation rule is resilient to both data sampling and training randomness, supporting its reliability in practical estimation and inference tasks.

\subsubsection{One-Shot Fits from Early Sizes}
\label{appssec:oneshot}
To assess whether a few small training sizes suffice to guide allocation, we fit the scaling law using a single training run at each of the four smallest sizes, $s\in\{250,500,750,1000\}$, with the same fixed validation split. We repeat this exercise for each of the ten runs. For each run, we fit a reference scaling law to the other nine runs and use it to evaluate the selected allocation.

We compare the predicted confidence interval width at the selected allocation with the minimum width predicted by the same reference law. The selected allocation ratios vary substantially: at $n=1{,}000$, they range from $10\%$ to $28\%$. However, the corresponding increases in predicted interval width are modest. For each budget, we compute the median percentage increase across the ten runs; these medians range from $0.3\%$ to $5.3\%$ across the tested budgets, and the largest increase across all runs and budgets is $10.4\%$.

Extending the fitting grid to six sizes up to $2{,}000$ reduces the median increase to below $3\%$ at every budget, with a maximum of $7.9\%$. Thus, even when the estimated allocation differs from the reference optimum, inference precision changes relatively little because the interval-width objective is flat near its minimum.

\subsection{A Cross-Fitting Type Sample Splitting}
\label{appsec:CrossFitting}

Now we discuss how to combine the optimal sample allocation with cross-fitting type sample splitting. Suppose that the optimal sample allocation suggests using $s$ samples for fine-tuning and $n-s$ samples for PPI. We find the smallest integers $p$ and $q$ such that
\begin{align*}
\frac{s}{n} = \frac{p}{q},
\end{align*} by dividing both $s$ and $n$ by their greatest common divisor. If $p$ and $q$ are still too large, we approximately find a reasonable pair of integers $p$ and $q$ such that $\frac{s}{n} \approx \frac{p}{q}$.

After we obtain a pair of integers $p$ and $q$, we split the entire sample into $q$ folds and number them as $\sN^{(1)}$, $\sN^{(2)}$, ..., $\sN^{(q)}$. First, we use samples from the first $p$ folds $\sN^{(1)}$, ..., $\sN^{(p)}$ for fine-tuning and then use samples from the remaining $q-p$ folds $\sN^{(p+1)}$, ..., $\sN^{(q)}$ for PPI. Second, we use samples from $\sN^{(2)}$, ..., $\sN^{(p+1)}$ for fine-tuning and then use samples from the remaining $q-p$ folds $\sN^{(p+2)}$, ..., $\sN^{(1)}$ for PPI. We repeat this rotation for $q$ times and then aggregate the $q$ estimators $\widehat{\mu}^{(1)}$, ..., $\widehat{\mu}^{(q)}$ by taking their average. This procedure ensures that, across different splits, each sample contributes to both fine-tuning and PPI.
In Online Appendix~\ref{appssec:crossfit}, we empirically study the impacts of the training objective and the number of folds.

\begin{example}
\label{exa:CrossFitting} Suppose that we wish to use $\frac{2}{9} \approx 22.2\%$ samples for fine-tuning and $\frac{7}{9} \approx 77.8\%$ samples for PPI. To combine this with cross-fitting type sample splitting, we split the entire sample into $9$ folds and number them as $\sN^{(1)}$, $\sN^{(2)}$, ..., $\sN^{(9)}$. First, we use samples from $\sN^{(1)}$ and $\sN^{(2)}$ for fine-tuning and then use samples from $\sN^{(3)}$, ..., $\sN^{(9)}$ for PPI to obtain an estimator $\widehat{\mu}^{(1)}$. Second, we use samples from $\sN^{(2)}$ and $\sN^{(3)}$ for fine-tuning and then use samples from $\sN^{(4)}$, ..., $\sN^{(1)}$ for PPI to obtain an estimator $\widehat{\mu}^{(2)}$. We repeat this rotation for $9$ times and then aggregate the $9$ estimators $\widehat{\mu}^{(1)}$, ..., $\widehat{\mu}^{(9)}$ by taking their average.
\end{example}

\subsection{Data-Driven Ramp-Up Procedure}
\label{appsec:ramp-up}

In this subsection, we propose a data-driven ramp-up procedure that sequentially estimates the
scaling law and determines when to stop allocating additional labeled observations to surrogate
development.

The procedure reserves a fixed validation subset of size $n_{\rv}$ for estimating the response
curve. This reserve is distinct from any internal validation split used for checkpoint selection within
a development run. Consequently, after reserving these observations, the effective labeled budget
available for surrogate development and PPI rectification is
\[
n_{\mathrm{eff}} = n-n_{\rv}.
\]
If $s$ observations from this effective budget are allocated to surrogate development, the remaining
rectification sample has size $n_{\mathrm{eff}}-s$. The relevant allocation objective is therefore
\[
\frac{a s^{-\alpha}+b}{n_{\mathrm{eff}}-s},
\]
rather than $(a s^{-\alpha}+b)/(n-s)$. Thus, the allocation selected by the ramp-up procedure is
optimal conditional on the reserved validation sample.

We first describe the procedure informally. We reserve a fixed validation subset for estimating the
response curve. From the remaining labeled observations, we progressively enlarge a nested
development set. At each stage, we train the surrogate at the current development size, evaluate its
residual variance on the fixed validation subset, and update the fitted scaling law. Once the current
development size reaches or exceeds the allocation implied by the fitted curve, we stop the ramp-up
procedure and use all remaining labeled observations for PPI rectification.

Formally, let the procedure consider $k\geq 3$ candidate development sizes
\[
0<s_1<s_2<\cdots<s_k<n_{\mathrm{eff}}.
\]
We randomly select $n_{\rv}$ labeled observations from $\sN$ to form the fixed validation subset
$\sN_{\rv}$. From the remaining observations $\sN\setminus\sN_{\rv}$, we construct a nested
sequence
\[
\emptyset
\subset \sN_1
\subset \sN_2
\subset \cdots
\subset \sN_k
\subset \sN\setminus\sN_{\rv},
\]
where $|\sN_\ell|=s_\ell$. For example, one may first draw $\sN_k$ and then obtain the smaller
development sets by nested subsampling.

At stage $\ell\in[k]$, we develop a surrogate $f^{(s_\ell)}$ using $\sN_\ell$ and evaluate its
residual variance on the fixed validation subset. Define
\[
r_i^{(\ell)}
=
Y_i-f^{(s_\ell)}(\bm X_i),
\qquad
(\bm X_i,Y_i)\in\sN_{\rv},
\]
and let
\[
\bar r_\ell
=
\frac{1}{n_{\rv}}
\sum_{(\bm X_i,Y_i)\in\sN_{\rv}}
r_i^{(\ell)}.
\]
The corresponding validation estimate of the residual variance is
\[
\widehat V_\ell
=
\frac{1}{n_{\rv}-1}
\sum_{(\bm X_i,Y_i)\in\sN_{\rv}}
\left(
r_i^{(\ell)}-\bar r_\ell
\right)^2.
\]

Using the observations collected through stage $\ell$, we fit the scaling law by solving
\[
(\widehat a_\ell,\widehat\alpha_\ell,\widehat b_\ell)
=
\arg\min_{a,\alpha,b}
\sum_{\iota=1}^{\ell}
\left(
\widehat V_\iota
-
a s_\iota^{-\alpha}
-
b
\right)^2,
\]
subject to $a,\alpha>0$ and $b\geq 0$. Because the model contains three unknown parameters,
we begin fitting the curve once $\ell\geq 3$.

Let $\widehat s_\ell^*$ denote the allocation obtained from
Equation~\eqref{eqn:NumericalSolve} after replacing $(a,\alpha,b)$ with
$(\widehat a_\ell,\widehat\alpha_\ell,\widehat b_\ell)$ and replacing the total budget $n$ with
the effective budget $n_{\mathrm{eff}}$. If $\ell<k$ and
$\widehat s_\ell^*>s_\ell$, the procedure continues to stage $\ell+1$. Otherwise, it stops at the
current development size and sets $s=s_\ell$. Because the procedure has already used the
observations in $\sN_\ell$ for surrogate development, $s_\ell$ is the implementable stopping
allocation on the prespecified grid. The remaining observations
\[
\sN\setminus(\sN_{\rv}\cup\sN_\ell)
\]
are then used for PPI rectification.

The fixed validation subset is used only to estimate the response curve and is not reused for
surrogate development or PPI rectification. A more sample-efficient alternative could replace this
reserved subset with cross-validated estimates of residual variance along the nested development
sequence, thereby allowing all labels to be reused. We do not implement this variant here and leave
a formal analysis of the resulting dependence structure for future work.

\section{Supplementary Analyses for Mean Estimation}
\label{appsec:additional_empirics}

This section provides supplementary evidence for the mean-estimation analysis in
Section~\ref{sec:empirical}. We examine four dimensions: diagnostics and boundary conditions
for the training objective, alternative training and inference designs, the choice and quality of the
surrogate model, and additional mean-estimation applications beyond the main Wine Reviews corpus.

\subsection{Training-Objective Diagnostics and Boundary Conditions}
\label{appssec:loss_comparison}

This subsection examines when and why variance-loss training differs from conventional MSE training, and how the comparison changes under alternative training specifications. We consider three analyses: a residual-variance diagnostic on the validation set, a sensitivity analysis of outcome centering, and a comparison with cross-entropy training.

\subsubsection{Residual-Variance Comparison}
\label{appssec:res_var}

We analyze the behavior of the fine-tuned LLM on the validation set to explain the performance differences observed in the main analysis. As established in Proposition~\ref{prop:ppi_vs_mean_R2}, the efficiency of the PPI estimator is primarily governed by the magnitude of the residual variance $\Var(Y - f^{(s)}(\bm{X}))$. For a fixed allocation $s$ and a sufficiently large unlabeled sample, a lower residual variance directly reduces the leading variance term of the PPI estimator.

\begin{figure}[ht!]
\centering
\setlength{\fboxrule}{0.1pt}
\includegraphics[width=0.7\linewidth]{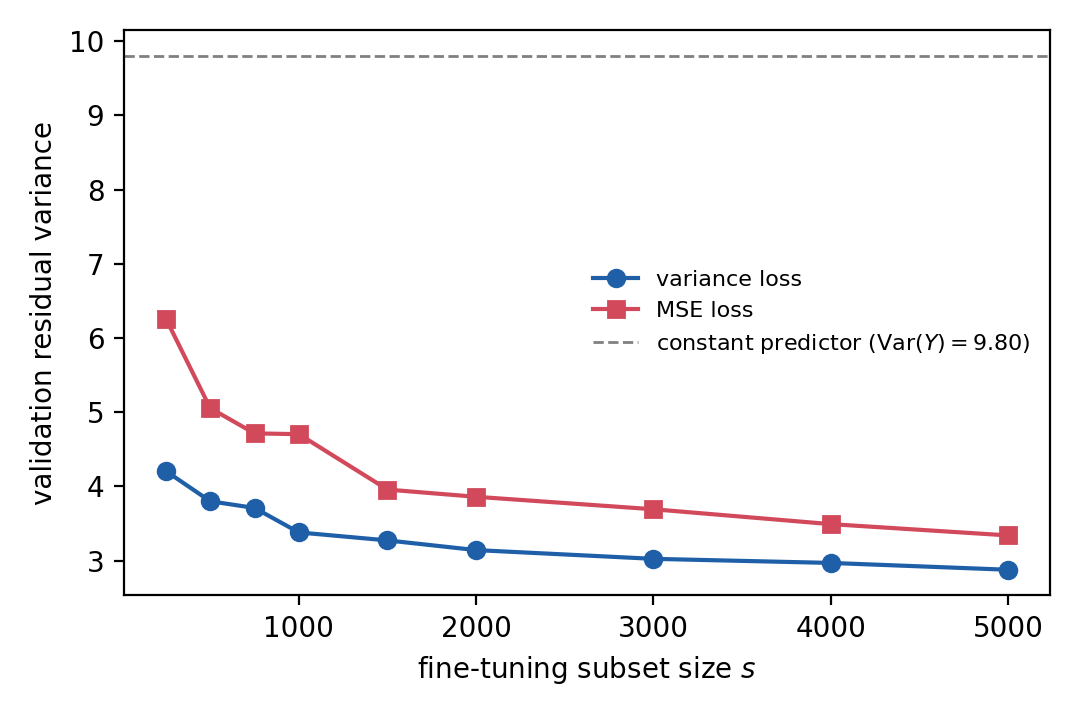}
\caption{
Validation residual variance under variance-based and MSE-based fine-tuning, with the constant-predictor baseline ($\mathrm{Var}(Y)$) for reference; current recipe (frozen \texttt{text-embedding-3-large} + MLP head).
}
\label{fig:loss-var-vs-mse}
\end{figure}

Figure~\ref{fig:loss-var-vs-mse} compares the validation residual variance of LLMs fine-tuned with our proposed variance loss ($\mathcal{L}_{\mathrm{var}}$) versus those trained with the standard MSE loss across varying subset sizes $s$. For reference, we include the residual variance of a constant predictor, which equals $\Var(Y)$. The results demonstrate that the variance-based objective consistently achieves lower residual variance than MSE. This advantage is particularly pronounced in the small-$s$ regime. Whereas MSE penalizes both the residual mean and residual variance, the variance loss directly targets the latter, which governs the leading PPI variance term.

\subsubsection{Sensitivity to Outcome Centering}
\label{app:mechanism}

We assess target centering as a preprocessing heuristic for MSE-based fine-tuning. We rerun the full empirical comparison under three treatments of the outcome level, using matched random seeds throughout: (A) targets centered by the training-sample mean; (B) raw targets on the $80$--$100$ scale; and (C) the main specification, with targets measured from the scale minimum ($Y-80$ here and $Y-1$ in the $1$--$5$ applications). Because the variance loss is invariant to additive shifts of the outcome, these transformations leave its training objective unchanged and isolate the sensitivity of MSE training to outcome-level handling.

Under specification (A), the variance-loss and MSE estimators are nearly indistinguishable, with differences in estimator variance of only $0.5\%$--$2.4\%$ across budgets. This preprocessing step is therefore effective in this dataset and training pipeline. Under the deployed specification (C), the variance-loss estimator attains the $22\%$--$42\%$ reductions in estimator variance reported in the main analysis. These results indicate that the empirical gap is sensitive to outcome-level handling during training. Under specification (B), early stopping occurs before the large intercept is adequately learned, producing a poorly performing MSE surrogate. We treat this specification as a diagnostic of level handling rather than as a practically relevant benchmark.\footnote{The three outcome-level specifications were evaluated under an earlier without-replacement sampling protocol with matched seeds. The deployed specification was subsequently rerun under the with-replacement protocol used in the main analysis and yields the headline variance reductions reported there.}

Centering alone does not generally make the training objectives equivalent. For centered targets, the identity
\[
\widehat{\bE}\bigl[(Y-f_\theta(X))^2\bigr]
=\widehat{\Var}\bigl(Y-f_\theta(X)\bigr)
+\bigl(\widehat{\bE}[f_\theta(X)]\bigr)^2
\]
holds, where both empirical moments use the same sample averages. If the initialized predictions have mean zero, the objectives and their gradients coincide at initialization, but the mean prediction can change during training.

To see that even the optimized fits can differ under regularization, consider two equally weighted observations $(X,Y)=(0,-1),(2,1)$, whose outcomes are already centered. Let $f_{w,b}(X)=wX+b$, with both $w$ and $b$ trainable, and apply the same ridge penalty $w^2+b^2$ to both objectives:
\[
\begin{aligned}
J_{\mathrm{MSE}}(w,b)&=(1-w)^2+(w+b)^2+w^2+b^2,\\
J_{\mathrm{Var}}(w,b)&=(1-w)^2+w^2+b^2.
\end{aligned}
\]
The unique global minimizers and minimum objective values are
\[
\begin{aligned}
(w_{\mathrm{MSE}},b_{\mathrm{MSE}})&=(2/5,-1/5),
&\min J_{\mathrm{MSE}}&=3/5,\\
(w_{\mathrm{Var}},b_{\mathrm{Var}})&=(1/2,0),
&\min J_{\mathrm{Var}}&=1/2.
\end{aligned}
\]
At these solutions, the unpenalized MSEs are $2/5$ and $1/2$, respectively, while the downstream residual variances satisfy
\[
\widehat{\Var}(Y-f_{w_{\mathrm{MSE}},b_{\mathrm{MSE}}}(X))
=\frac{9}{25}
>\frac14
=\widehat{\Var}(Y-f_{w_{\mathrm{Var}},b_{\mathrm{Var}}}(X)).
\]
Thus, centering need not eliminate the residual-variance gap even when both fits reach their global optima. The example includes a penalty on the trainable intercept; Remark~\ref{rmk:objectives}'s equivalence applies when the additive level is unrestricted and unpenalized. This example establishes a limitation of centering as a general remedy. For general $M$-estimation, heterogeneous influence on the estimand can also prevent a global level adjustment from aligning the objectives, as illustrated in Section~\ref{ssec:mest-synthetic} and Online Appendix~\ref{app:synthetic}.

\subsubsection{Alternative Training Loss: Cross-Entropy}
\label{appsec:ce}

For integer-valued ratings, a common alternative is to formulate prediction as a $21$-class classification problem and train with cross-entropy loss. At inference, we convert the predicted class probabilities to a scalar rating using their probability-weighted expectation. We compare \textsc{FT+PPI} using this classification model with \textsc{FT+PPI (Var Loss)}, holding the features, $80/20$ split, and allocations fixed; each model uses early stopping on its own validation objective. Table~\ref{tab:ce-loss} reports RMSE for the population mean estimate. Variance-loss training yields lower RMSE at the three smaller budgets, while the two methods perform similarly at $n=10{,}000$.

\begin{table}[ht!]\centering
\footnotesize
\caption{Cross-entropy training versus the variance loss}
\vspace{1mm}
\begin{tabular}{ccc}
\toprule
$n$ & \makecell{\textbf{RMSE (Cross-entropy)}\\\textbf{($\times 10^{-2}$)}} & \makecell{\textbf{RMSE (Var Loss)}\\\textbf{($\times 10^{-2}$)}} \\
\midrule
$1{,}000$ & $10.038$ & $7.873$ \\
$3{,}000$ & $4.303$ & $2.567$ \\
$5{,}000$ & $3.037$ & $2.816$ \\
$10{,}000$ & $2.471$ & $2.514$ \\
\bottomrule
\end{tabular}
\vspace{2mm}
\begin{tablenotes}
\linespread{1}\selectfont
\item Note: Results use $20$ paired replications.
\end{tablenotes}
\label{tab:ce-loss}
\end{table}

\subsection{Alternative Training and Inference Designs}
\label{app:designs}

This subsection examines whether the training-objective comparison persists under changes to the training architecture and inference design. We consider three alternatives: LoRA adaptation of an open-weight language model, label reuse through cross-fitting, and estimator-side power tuning through PPI++.

\subsubsection{Architecture Robustness: LoRA Fine-Tuning}
\label{app:lora}

The main study trains an MLP head on frozen embeddings. To examine whether the objective comparison persists when the language model itself is fine-tuned, we use LoRA fine-tuning of \texttt{Qwen2.5-1.5B-Instruct}, with rank-$8$ adapters on the attention projections, a scalar prediction head, and text-only prompts. At $s=1{,}135$ and $n=10{,}000$, we compare MSE with the mini-batch residual-variance loss using the same fine-tuning draw, initialization, and training-data order. Each objective selects its training duration using its own five-fold validation criterion, followed by one refit on all $s$ fine-tuning observations.

Table~\ref{tab:mean-lora} reports held-out residual variance and the resulting PPI estimation errors. Variance-loss training reduces residual variance from $3.626$ to $3.205$, an $11.6\%$ reduction, and yields lower RMSE, MAE, and MAPE. This comparison uses one matched training pair at a single allocation.

\begin{table}[ht!]\centering
\footnotesize
\caption{LoRA fine-tuning: residual variance and PPI estimation accuracy}
\vspace{1mm}
\begin{tabular}{lcccc}
\toprule
\textbf{Estimator} & \makecell{\textbf{Residual}\\\textbf{variance}} & \makecell{\textbf{RMSE}\\\textbf{($\times 10^{-2}$)}} & \makecell{\textbf{MAE}\\\textbf{($\times 10^{-2}$)}} & \makecell{\textbf{MAPE}\\\textbf{($\times 10^{-2}\%$)}} \\
\midrule
\textsc{FT+PPI (MSE Loss)} & $3.626$ & $2.187$ & $1.743$ & $1.969$ \\
\textsc{FT+PPI (Var Loss)} & $3.205$ & $1.951$ & $1.580$ & $1.786$ \\
\bottomrule
\end{tabular}
\vspace{2mm}
\begin{tablenotes}
\linespread{1}\selectfont
\item Note: $n=10{,}000$ and $s=1{,}135$. Residual variance is evaluated on target-population reviews excluded from fine-tuning. Estimation errors summarize $60$ paired redraws of the $n-s$ rectification observations, conditional on the fitted surrogates. Metric units match Table~\ref{tab:comparison-estimators-by-N}.
\end{tablenotes}
\label{tab:mean-lora}
\end{table}

\subsubsection{Cross-Fitting: Label Reuse and the Training Objective}
\label{appssec:crossfit}

Following the cross-prediction approach of \citet{zrnic2024cross}, we consider a $K$-fold cross-fitted estimator that reuses every labeled observation for both surrogate development and rectification:
\[
\widehat{\mu}_{\mathrm{CF}}
=
\frac{1}{m}\sum_{j=1}^{m}\bar f(\tilde{\bm X}_j)
+
\frac{1}{n}\sum_{i=1}^{n}
\left(
Y_i-f^{(-k(i))}(\bm X_i)
\right),
\]
where $f^{(-k(i))}$ denotes the fold model trained on the $K-1$ folds that exclude observation $i$, and
\[
\bar f(\bm x)
=
\frac{1}{K}\sum_{k=1}^{K} f^{(-k)}(\bm x)
\]
denotes the average prediction across the $K$ fold models. Because every labeled observation enters a rectification fold, the correction term is scaled by $1/n$ rather than $1/(n-s)$. Standard cross-fitting therefore removes the labeled-sample allocation trade-off analyzed in Theorem~\ref{thm:OptimalAllocation}; the remaining design choices concern the training objective used for the fold models and the number of folds $K$.

We evaluate these choices under the same $n=10{,}000$ protocol as Table~\ref{tab:comparison-estimators-by-N}, using $300$ paired replications on the same labeled-data draws. We consider $K\in\{3,4,5\}$ and, within each $K$, use identical fold partitions for the MSE-trained and variance-trained models. The single-split benchmark rows reproduce Table~\ref{tab:comparison-estimators-by-N}. Table~\ref{tab:crossfit-obj} reports the results.

\begin{table}[ht!]
\centering
\footnotesize
\caption{Cross-fitting on the main corpus: training objective at each fold count}
\vspace{1mm}
\begin{tabular}{llccc}
\toprule
\textbf{Design}
& \textbf{Training Objective}
& \textbf{RMSE ($\times 10^{-2}$)}
& \textbf{MAE ($\times 10^{-2}$)}
& \textbf{MAPE ($\times 10^{-2}\%$)} \\
\midrule
\textsc{Sample Mean}
& ---
& $3.374\;(0.121)$
& $2.766\;(0.112)$
& $3.126\;(0.126)$ \\

\textsc{PPI-Only}
& ---
& $2.473\;(0.104)$
& $1.991\;(0.085)$
& $2.250\;(0.096)$ \\

FT+PPI split
& MSE
& $2.381\;(0.094)$
& $1.892\;(0.084)$
& $2.138\;(0.094)$ \\

FT+PPI split
& Variance
& $2.177\;(0.082)$
& $1.752\;(0.075)$
& $1.980\;(0.084)$ \\
\midrule

Cross-fitting $K=3$
& MSE
& $1.909\;(0.083)$
& $1.509\;(0.068)$
& $1.706\;(0.076)$ \\

& Variance
& \bm{$1.734\;(0.081)$}
& \bm{$1.372\;(0.061)$}
& \bm{$1.551\;(0.069)$} \\

Cross-fitting $K=4$
& MSE
& $1.908\;(0.084)$
& $1.491\;(0.069)$
& $1.685\;(0.078)$ \\

& Variance
& \bm{$1.665\;(0.068)$}
& \bm{$1.335\;(0.058)$}
& \bm{$1.508\;(0.065)$} \\

Cross-fitting $K=5$
& MSE
& $1.861\;(0.081)$
& $1.452\;(0.067)$
& $1.641\;(0.076)$ \\

& Variance
& \bm{$1.687\;(0.078)$}
& \bm{$1.297\;(0.062)$}
& \bm{$1.466\;(0.071)$} \\
\bottomrule
\end{tabular}

\begin{tablenotes}
\linespread{1}\selectfont
\item Note: Standard errors are reported in parentheses. Results use $n=10{,}000$ and $300$ paired replications on the same labeled-data draws as Table~\ref{tab:comparison-estimators-by-N}. Within each $K$, the MSE and variance-loss estimators use identical fold partitions. The variance-loss single-split estimator uses the rule-selected allocation $s^*=1{,}027$, while the MSE single-split estimator selects its allocation over a grid at $5\%$ intervals. Boldface identifies the lower error within each MSE-versus-variance cross-fitting pair at a fixed $K$.
\end{tablenotes}
\label{tab:crossfit-obj}
\end{table}

Three findings emerge. First, the training-objective comparison persists within the cross-fitted design. Holding the folds fixed, variance-trained models achieve approximately $9\%$--$13\%$ lower RMSE than their MSE-trained counterparts across $K=3,4,5$. Paired comparisons of absolute estimation errors yield $t=2.2$, $2.5$, and $2.5$, respectively. The variance-trained folds also have $5.9\%$--$7.0\%$ lower estimated estimator variance across the three fold counts. The contrast is larger in a more data-limited $n=1{,}000$ check, where variance-trained folds reduce estimated estimator variance by $24\%$--$26\%$.

Second, label reuse materially improves statistical efficiency at the larger budget. Relative to the variance-trained single-split estimator, variance-trained cross-fitting reduces RMSE by approximately $20\%$--$24\%$ across the three fold counts. This gain comes at substantially higher computational cost. A $K$-fold implementation trains $K$ models, each using $(K-1)n/K$ labeled observations, corresponding to approximately $19$ to $39$ times as much training-example exposure as the single $s^*=1{,}027$-label fit. Cross-fitting therefore trades additional training compute for improved label efficiency, while the variance-based objective continues to improve the fold models within the cross-fitted design.

Third, the realized errors differ little across $K=3,4,5$, suggesting limited sensitivity to the fold count over the range examined here.

\subsubsection{Power Tuning (PPI++)}
\label{appssec:ppi++}

To separate the effect of the training objective from estimator-side calibration, we evaluate a full $2\times2$ design: surrogates trained with either MSE or the variance loss are deployed under both plain PPI ($\lambda=1$) and PPI++ with an estimated power parameter $\widehat{\lambda}$ following \citet{angelopoulos2023ppi}. Table~\ref{tab:twobytwo} reports the resulting estimator variances.

The estimated power parameters reveal a clear difference between the two trained surrogates. The variance-trained surrogate requires relatively little power adjustment, with mean $\widehat{\lambda}=0.97$ and $0.95$ at $n=1{,}000$ and $10{,}000$, respectively. The MSE-trained surrogate exhibits larger departures from the unit-weight solution, with mean $\widehat{\lambda}=0.61$ and $0.82$. Consistent with this pattern, power tuning substantially improves the MSE-trained estimator: relative to plain PPI, it recovers approximately $43\%$ and $52\%$ of the initial estimator-variance gap between the two training objectives at the two budgets. Nevertheless, after both surrogates are power-tuned, the variance-trained estimator retains $33.3\%$ and $14.1\%$ lower estimator variance at $n=1{,}000$ and $10{,}000$, respectively. Thus, estimator-side calibration partially substitutes for training-objective alignment in this experiment, but does not eliminate the remaining efficiency difference.

At the population level and in the large-unlabeled-sample regime, oracle power tuning changes the relevant residual criterion from
$\Var(Y-f(\bm X))$ to
\[
\min_{\lambda}\Var\!\left(Y-\lambda f(\bm X)\right).
\]
Accordingly, the unit-weight optimality results in the main text apply directly to plain PPI, whereas PPI++ introduces an additional calibration dimension that must itself be estimated from the labeled data. The comparison here therefore evaluates how much of the empirical training-objective gap can be recovered through this additional estimator-side adjustment.

\begin{table}[ht!]
\centering
\footnotesize
\caption{Estimator variance under plain PPI and PPI++}

\begin{tabular}{clccc}
\toprule
$n$
& \textbf{Training Objective}
& \textbf{PPI Variance}
& \textbf{PPI++ Variance}
& \textbf{Mean $\widehat{\lambda}$} \\
\midrule

$1{,}000$
& MSE
& $9.38\times10^{-3}$
& $7.58\times10^{-3}$
& $0.61$ \\

& Variance
& $5.17\times10^{-3}$
& $\bm{5.06\times10^{-3}}$
& $0.97$ \\

\midrule

$10{,}000$
& MSE
& $6.11\times10^{-4}$
& $5.42\times10^{-4}$
& $0.82$ \\

& Variance
& $4.79\times10^{-4}$
& $\bm{4.65\times10^{-4}}$
& $0.95$ \\

\bottomrule
\end{tabular}

\begin{tablenotes}
\linespread{1}\selectfont
\item Note: Reported values are mean estimated estimator variances across $20$ paired replications. Plain PPI uses $\lambda=1$, whereas PPI++ estimates the power parameter $\widehat{\lambda}$ from the labeled data. The final column reports the mean estimated $\widehat{\lambda}$ for the corresponding PPI++ estimator. Boldface identifies the lowest estimator variance within each labeled-data budget.
\end{tablenotes}

\label{tab:twobytwo}
\end{table}

\subsection{Surrogate Model Choice and Quality}
\label{appsec:surrogate_models}

This subsection studies how downstream efficiency and the preferred deployment change with three features of the surrogate pipeline: the text representation used for supervised adaptation, the quality of the census-wide surrogate available before adaptation, and the complexity of the fine-tuning class.

\subsubsection{Classical Text Baseline: TF-IDF}
\label{appsssec:tf_idf}

We first assess whether the gains from the main LLM-embedding pipeline persist relative to a classical sparse-text representation. We replace the pretrained embedding with a $20{,}000$-feature TF-IDF representation with bigrams and fit ridge regression as the prediction model. At labeled-data budgets of $n=1{,}000$ and $n=10{,}000$, the LLM-embedding pipeline attains $43\%$ and $26\%$ lower estimator variance, respectively, than the TF-IDF baseline. Thus, in this application, the LLM-embedding pipeline remains substantially more statistically efficient than a conventional sparse-text alternative.

\subsubsection{Zero-Shot Model Quality and the Extended Allocation Rule}
\label{appsec:frontier}

The main study's \textsc{PPI-Only} benchmark uses the zero-shot predictions of \texttt{gpt-5.4-nano}, which have a correlation of $0.741$ with the human ratings. We additionally evaluate stronger closed-weight zero-shot models to examine a practically important question: when a high-quality surrogate is already available census-wide at essentially zero marginal label cost, how much labeled data, if any, should still be allocated to supervised adaptation?

We cast this question as an allocation problem over an expanded set of candidate surrogate classes. The raw zero-shot predictor constitutes an explicit $s=0$ deployment option. For $s>0$, we consider several forms of supervised adaptation, including an affine calibration of the zero-shot prediction, a combined MLP using both embedding and zero-shot information, and the main-text embedding-head FT+PPI pipeline. In particular, the affine class
\[
f(\bm X)=c_0+c_2 f_{\mathrm{zs}}(\bm X)
\]
contains the raw zero-shot predictor as the special case $(c_0,c_2)=(0,1)$.

For each candidate class $\mathcal C$, we estimate its residual-variance response curve $\widehat{\nu}_{\mathcal C}(s)$ and evaluate the corresponding allocation objective
\[
\frac{\widehat{\nu}_{\mathcal C}(s)}{n-s}.
\]
The extended rule then selects the feasible class-allocation pair $(\mathcal C,s)$ that minimizes this criterion, including the raw zero-shot corner at $s=0$. This formulation allows the amount of supervised adaptation to adjust endogenously to both the quality of the starting surrogate and the complexity of the fine-tuning class, rather than presuming that additional fine-tuning is always beneficial.

We next isolate the role of the quality of the census-wide zero-shot surrogate. We obtain text-only zero-shot predictions for all $90{,}610$ reviews from four closed-weight models spanning two model generations: \texttt{gpt-5.4-nano}, used in the main study, with correlation $0.741$ with the human ratings; \texttt{gpt-5.6-terra}, with correlation $0.860$; \texttt{gpt-5.6-sol}, with correlation $0.863$; and \texttt{gpt-5.6-luna}, with correlation $0.885$. The three 2026 models are prompted zero-shot with ten reviews per prompt.\footnote{\texttt{gpt-5.6-luna} was evaluated at its default reasoning setting because the low-reasoning configuration used for the other models was unavailable. Its correlation should therefore be interpreted with this configuration difference in mind. A price-included variant of \texttt{gpt-5.6-sol} is reported separately in Online Appendix~\ref{appssec:mest_supp}.}

To isolate zero-shot model quality from the complexity of the fine-tuning class, we hold this class fixed across models. Specifically, for each zero-shot surrogate we use the affine calibration
\[
f(\bm X)
=
c_0+c_2f_{\mathrm{zs}}(\bm X),
\]
estimated using $s$ labeled samples, and apply the allocation rule to select $s$. We evaluate labeled-data budgets ranging from $n=1{,}000$ to $n=20{,}000$, using the full corpus as the target population, labels drawn with replacement, and $20$ replications per budget. Across all models and budgets, the selected calibration allocation remains small, with $s^*$ between $30$ and $250$ labels and between $30$ and $100$ labels for three of the four models.

Table~\ref{tab:zs-models} summarizes the resulting efficiency gains. Within this set of models, higher zero-shot correlation is associated with larger variance reductions relative to the sample mean. The savings increase from approximately $52\%$--$54\%$ for \texttt{gpt-5.4-nano} to $77\%$--$78\%$ for \texttt{gpt-5.6-luna}, with relatively little variation in the percentage savings across the labeled-data budgets considered.

\begin{table}[ht!]
\centering
\footnotesize
\caption{Zero-shot model quality and estimator efficiency under affine calibration}

\begin{tabular}{lcc}
\toprule
\textbf{Zero-Shot Model}
& \textbf{Corr.\ with Human Ratings}
& \makecell{\textbf{Variance reduction}\\\textbf{w.r.t.\ \textsc{Sample Mean}}} \\
\midrule
\texttt{gpt-5.4-nano}
& $0.741$
& $52\%$--$54\%$ \\

\texttt{gpt-5.6-terra}
& $0.860$
& $72\%$--$74\%$ \\

\texttt{gpt-5.6-sol}
& $0.863$
& $73\%$--$74\%$ \\

\texttt{gpt-5.6-luna}
& $0.885$
& $77\%$--$78\%$ \\
\bottomrule
\end{tabular}

\begin{tablenotes}
\linespread{1}\selectfont
\item Note: Variance reductions compare the deployed estimator with \textsc{Sample Mean} at the same labeled-data budget. Reported ranges are across budgets from $n=1{,}000$ to $n=20{,}000$, with $20$ replications per budget, labels drawn with replacement, and the full $90{,}610$-review corpus used as the target population. For every model, the fine-tuning class is the affine calibration $f=c_0+c_2f_{\mathrm{zs}}$. The allocation rule selects between $30$ and $250$ labeled samples in every reported setting.
\end{tablenotes}

\label{tab:zs-models}
\end{table}

Two conclusions emerge. First, improving the quality of the census-wide zero-shot surrogate translates into substantial gains in downstream statistical efficiency: within the common affine fine-tuning class, the estimator-variance reduction increases systematically with zero-shot model quality. Second, even for the weaker zero-shot model, the allocation rule assigns only a relatively small number of labels to calibration. As the zero-shot surrogate strengthens, the preferred deployment therefore shifts toward lightweight supervised adaptation rather than extensive model retraining. The next analysis holds the zero-shot model fixed and instead varies the complexity of the fine-tuning class.

\subsubsection{Choosing the Fine-Tuning Class}
\label{appsec:finetuning_class}

Holding the text-only zero-shot surrogate \texttt{gpt-5.6-sol} fixed, we next vary the complexity of the fine-tuning class and allow each class to operate at its own rule-selected allocation. The raw zero-shot predictor corresponds to the $s=0$ corner and has residual variance $2.98$. The affine calibration reduces the residual variance to $2.53$, an improvement of approximately $15\%$ over the raw corner. Its allocation objective is also relatively flat over $s\in[30,500]$, which contains the small calibration allocations selected by the extended rule in Table~\ref{tab:zs-models}. We then consider a richer MLP class that combines the zero-shot prediction with $64$ principal components of the text embedding. This class selects $s^*=1{,}175$ and reduces residual variance by $12.4\%$ relative to the raw zero-shot predictor.

Among these candidate deployments, the affine calibration achieves the lowest measured residual-variance criterion while requiring substantially fewer labeled samples than the richer MLP class. After accounting for both surrogate quality and the labels consumed by adaptation, the extended class-allocation rule therefore selects the affine calibration at $n=10{,}000$. This comparison illustrates that a more flexible fine-tuning class need not be preferred when a strong census-wide surrogate is already available and the additional flexibility requires substantially more labeled data.

We also examine how the zero-shot prediction is incorporated into richer fine-tuning classes. A direct concatenation of the scalar zero-shot prediction with the raw $3{,}072$-dimensional embedding performs poorly in the small-development regime: under the measured training procedure, its residual-variance curve remains above that of the raw zero-shot predictor until approximately $s=3{,}000$. Consequently, this specification would favor the $s=0$ corner throughout much of the scarce-label range, despite the presence of potentially complementary information in the embedding.

We consider two alternative parameterizations. First, an additive specification,
\[
f(\bm X)
=
f_{\mathrm{zs}}(\bm X)+g(\bm X),
\]
with $g$ initialized at zero, begins exactly at the zero-shot predictor and contains it as the special case $g\equiv0$. This specification improves residual variance by $8.6\%$ relative to the raw zero-shot corner at $n=10{,}000$, but adapts more slowly than the dimension-reduced MLP specification described above, which achieves a $12.4\%$ reduction under the same protocol. Second, reducing the embedding to $64$ principal components before combining it with the zero-shot prediction substantially improves performance in the limited-label regime. These comparisons indicate that, when a strong zero-shot surrogate is available, the parameterization of the fine-tuning class can materially affect how efficiently additional labeled data are used. Full response curves for these implementation variants are provided in the replication materials.

\subsection{Additional Mean-Estimation Applications}
\label{appsec:applications}

This subsection evaluates the mean-estimation pipeline beyond the main Wine Reviews corpus. We consider two additional settings: a second review-rating application and a digital-twin survey panel with a strong census-wide prior-wave surrogate.

\subsubsection{Customer-Review Ratings}
\label{app:clothing}

We first replicate the mean-estimation analysis on a corpus of $22{,}634$ women's e-commerce clothing reviews with $1$--$5$ star ratings. We reserve $4{,}000$ reviews for estimating the response curve and use the remaining reviews as the target population, whose mean rating is $4.186$. The FT+PPI estimators follow the main study's training configuration, using frozen \texttt{text-embedding-3-large} representations, an MLP prediction head, and checkpoint selection based on each model's own validation objective. As a strong prompting benchmark, the \textsc{PPI-Only} estimator rectifies zero-shot predictions from \texttt{gpt-5.6-sol}, which have correlation $0.860$ with the human ratings. Its prompts include item metadata in addition to the review text.

For the variance-trained surrogate, the fitted response curve is
\[
80.84\,s^{-1.213}+0.393,
\]
with adjusted $R^2=0.963$ over the five measured development sizes. Applying the allocation rule yields $s^*=218$, $321$, and $462$ at labeled-data budgets $n=1{,}000$, $2{,}000$, and $4{,}000$, respectively. Table~\ref{tab:clothing} reports the resulting estimation performance.

\begin{table}[ht!]
\centering
\footnotesize
\caption{Customer-review corpus: estimation accuracy across estimators}
\begin{tabular}{llccc}
\toprule
$n$ & \textbf{Estimator}
& \textbf{RMSE ($\times 10^{-2}$)}
& \textbf{MAE ($\times 10^{-2}$)}
& \textbf{MAPE (\%)} \\
\midrule

$1{,}000$
& \textsc{Sample Mean}
& $3.286\;(0.328)$
& $2.905\;(0.352)$
& $0.694\;(0.084)$ \\

& \textsc{PPI-Only}
& $2.293\;(0.341)$
& $1.886\;(0.299)$
& $0.451\;(0.071)$ \\

& \textsc{FT+PPI (MSE Loss)}
& $3.728\;(0.445)$
& $3.079\;(0.482)$
& $0.736\;(0.115)$ \\

& \textsc{FT+PPI (Var Loss)}
& $2.539\;(0.463)$
& $2.041\;(0.347)$
& $0.488\;(0.083)$ \\

\midrule

$2{,}000$
& \textsc{Sample Mean}
& $2.504\;(0.462)$
& $1.923\;(0.368)$
& $0.459\;(0.088)$ \\

& \textsc{PPI-Only}
& $1.534\;(0.135)$
& $1.384\;(0.152)$
& $0.331\;(0.036)$ \\

& \textsc{FT+PPI (MSE Loss)}
& $2.118\;(0.386)$
& $1.625\;(0.311)$
& $0.388\;(0.074)$ \\

& \textsc{FT+PPI (Var Loss)}
& $1.332\;(0.270)$
& $0.963\;(0.211)$
& $0.230\;(0.050)$ \\

\midrule

$4{,}000$
& \textsc{Sample Mean}
& $2.569\;(0.368)$
& $2.024\;(0.363)$
& $0.484\;(0.087)$ \\

& \textsc{PPI-Only}
& $1.440\;(0.220)$
& $1.163\;(0.195)$
& $0.278\;(0.047)$ \\

& \textsc{FT+PPI (MSE Loss)}
& $1.834\;(0.299)$
& $1.467\;(0.252)$
& $0.350\;(0.060)$ \\

& \textsc{FT+PPI (Var Loss)}
& $1.353\;(0.180)$
& $1.151\;(0.163)$
& $0.275\;(0.039)$ \\
\bottomrule
\end{tabular}
\begin{tablenotes}
\linespread{1}\selectfont
\item Note: Reported values are averages across $20$ replications relative to the target-population mean $4.186$, with standard errors in parentheses. The two FT+PPI estimators use the rule-selected allocations $s^*=218$, $321$, and $462$. The variance-loss estimator reduces estimated estimator variance relative to the MSE-trained estimator by $40.1\%$, $35.4\%$, and $29.7\%$ across the three budgets. Relative to \textsc{Sample Mean}, the corresponding reductions are $41.9\%$, $47.2\%$, and $40.9\%$ for the variance-loss estimator and $62.4\%$, $55.1\%$, and $37.4\%$ for \textsc{PPI-Only}. The \textsc{PPI-Only} benchmark uses \texttt{gpt-5.6-sol} with item metadata included in the prompt.
\end{tablenotes}

\label{tab:clothing}
\end{table}

Three findings are noteworthy. First, the advantage of variance-loss training over MSE training carries over to the second review corpus. Across the three labeled-data budgets, the variance-loss estimator reduces estimated estimator variance by $29.7\%$--$40.1\%$ relative to the MSE-trained estimator, a magnitude comparable to that observed in the main Wine Reviews application.

Second, the strong zero-shot benchmark performs particularly well at the smaller budgets. In terms of estimated estimator variance, \textsc{PPI-Only} is more efficient than the variance-trained FT+PPI estimator at $n=1{,}000$ and $n=2{,}000$. Realized point-estimation errors need not follow the same ranking in every finite set of replications; for example, at $n=2{,}000$ the variance-trained estimator has lower realized RMSE even though the estimated variance favors \textsc{PPI-Only}.

Third, the variance comparison reverses at the largest budget considered. At $n=4{,}000$, the variance-trained FT+PPI estimator reduces estimator variance by $40.9\%$ relative to \textsc{Sample Mean}, compared with $37.4\%$ for \textsc{PPI-Only}. This application therefore illustrates that the preferred deployment can depend on the available labeling budget: direct rectification of a strong pre-existing surrogate can be attractive when labels are scarce, while supervised adaptation can become more competitive as the labeled-data budget increases.

\subsubsection{Digital-Twin Survey Panel (Twin-2K-500)}
\label{app:survey}

We next consider the Twin-2K survey panel, which contains a census of $2{,}058$ respondents. The estimand is the population mean of wave-4 policy support, with $\mu^*=3.720$ on a $1$--$5$ scale. This setting provides a strong census-wide surrogate at essentially zero marginal labeling cost: each respondent's prior-wave answer is observed before the wave-4 outcome and has correlation $0.816$ with the target response.

The \textsc{PPI-Only} benchmark directly rectifies this prior-wave surrogate. The FT+PPI estimators use an MLP head whose inputs include persona-embedding principal components, demographics, and the prior-wave response, with checkpoint selection based on each model's own validation objective. We evaluate labeled-data budgets $n\in\{250,500,1{,}000\}$ over $20$ replications. Development sizes are chosen from the measured set $s\in\{30,60,100,250\}$, using $s=100$, $250$, and $250$ at the three budgets, respectively. Table~\ref{tab:survey} reports the resulting estimation performance.

\begin{table}[ht!]
\centering
\footnotesize
\caption{Digital-twin survey panel: estimation accuracy across estimators}
\begin{tabular}{llccc}
\toprule
$n$ & \textbf{Estimator}
& \textbf{RMSE ($\times 10^{-2}$)}
& \textbf{MAE ($\times 10^{-2}$)}
& \textbf{MAPE (\%)} \\
\midrule

$250$
& \textsc{Sample Mean}
& $8.603\;(1.166)$
& $6.997\;(1.149)$
& $1.881\;(0.309)$ \\

& \textsc{PPI-Only (prior wave)}
& $5.251\;(0.591)$
& $4.402\;(0.657)$
& $1.183\;(0.177)$ \\

& \textsc{FT+PPI (MSE Loss)}
& $8.452\;(1.055)$
& $7.138\;(1.039)$
& $1.919\;(0.279)$ \\

& \textsc{FT+PPI (Var Loss)}
& $8.798\;(1.250)$
& $6.841\;(1.269)$
& $1.839\;(0.341)$ \\

\midrule

$500$
& \textsc{Sample Mean}
& $6.245\;(0.728)$
& $5.309\;(0.755)$
& $1.427\;(0.203)$ \\

& \textsc{PPI-Only (prior wave)}
& $4.732\;(0.669)$
& $3.878\;(0.622)$
& $1.043\;(0.167)$ \\

& \textsc{FT+PPI (MSE Loss)}
& $4.621\;(0.726)$
& $3.817\;(0.597)$
& $1.026\;(0.161)$ \\

& \textsc{FT+PPI (Var Loss)}
& $4.288\;(0.603)$
& $3.448\;(0.584)$
& $0.927\;(0.157)$ \\

\midrule

$1{,}000$
& \textsc{Sample Mean}
& $4.830\;(0.646)$
& $3.866\;(0.664)$
& $1.039\;(0.179)$ \\

& \textsc{PPI-Only (prior wave)}
& $3.358\;(0.650)$
& $2.484\;(0.518)$
& $0.668\;(0.139)$ \\

& \textsc{FT+PPI (MSE Loss)}
& $4.038\;(0.581)$
& $3.227\;(0.557)$
& $0.868\;(0.150)$ \\

& \textsc{FT+PPI (Var Loss)}
& $2.350\;(0.243)$
& $2.057\;(0.261)$
& $0.553\;(0.070)$ \\
\bottomrule
\end{tabular}
\begin{tablenotes}
\linespread{1}\selectfont
\item Note: Reported values are averages across $20$ replications relative to the census mean $3.720$, with standard errors in parentheses. The FT+PPI estimators use development sizes $s=100$, $250$, and $250$ at $n=250$, $500$, and $1{,}000$, respectively. Labels are drawn with replacement from the census, following the same superpopulation sampling convention as the main study. Across the three budgets, the variance-loss estimator has $5.9\%$, $4.9\%$, and $7.2\%$ lower estimated estimator variance than the corresponding MSE-trained estimator. Empirical coverage of the nominal $95\%$ confidence intervals ranges from $0.90$ to $1.00$.
\end{tablenotes}
\label{tab:survey}
\end{table}

Three findings emerge. First, this application illustrates the value of directly rectifying a strong pre-existing surrogate when labeled outcomes are scarce. At $n=250$ and $n=500$, the prior-wave \textsc{PPI-Only} estimator has lower estimated estimator variance than the variance-trained FT+PPI estimator. The realized error metrics need not follow the same ranking in every finite set of replications; for example, at $n=500$ the variance-trained FT+PPI estimator has lower realized RMSE even though the estimated variance favors direct rectification of the prior-wave surrogate.

Second, the comparison changes at the largest budget. At $n=1{,}000$, the variance-trained FT+PPI estimator substantially improves on direct rectification of the prior-wave surrogate, with RMSE $2.35\times10^{-2}$ compared with $3.36\times10^{-2}$. Together with the smaller-budget results, this pattern shows that the preferred deployment can depend on the available labeling budget: a strong census-wide surrogate can make direct rectification attractive when labels are scarce, while supervised adaptation becomes more competitive as additional labeled data become available.

Third, the training-objective comparison has the same qualitative direction as in the text applications. The variance-loss estimator has $5.9\%$, $4.9\%$, and $7.2\%$ lower estimated estimator variance than the MSE-trained estimator across the three budgets. Although these differences are smaller than those observed in the review-rating applications, they provide evidence that the downstream variance criterion remains relevant in a distinct data environment with a non-LLM, census-wide surrogate.

\section{Supplementary Analyses for \texorpdfstring{$M$}{M}-Estimation}
\label{appsec:mestimation}

This section provides supplementary analyses for the $M$-estimation study in
Section~\ref{sec:mest-case}. We first examine the price-rating regression case study through
additional surrogate-specification checks. We then present the
controlled synthetic design used to isolate the mechanism behind the objective gap and the
conditions under which that gap disappears.

\subsection{Additional Checks for the Price-Rating Regression Case Study}
\label{appssec:mest_supp}

This subsection provides supplementary checks for the price-rating regression case
study in Section~\ref{sec:mest-case}. We examine the fine-tuning and input specifications,
including the affine fine-tuning specification, a richer MLP alternative, and a price-in-prompt benchmark.

For the price-rating application, we fit the slope-specific variance criterion across eleven fine-tuning sample sizes,
\[
s\in\{50,100,150,250,375,500,750,1000,1500,2000,3000\}.
\]
At each size, we fit the fine-tuning specification in Section~\ref{ssec:mest-training} over $10$ replications and use the median price-weighted residual variance on the scaling-law validation split to fit the power law. The fitted law has adjusted $R^2=0.98$ and selects $s^*=318$ fine-tuning labels at $n=10{,}000$. The allocation objective remains within $1\%$ of its minimum for allocations between approximately $180$ and $550$ labels.

\subsubsection{Fine-Tuning and Input Specifications}
\label{appssec:surrogate}
To give the surrogate explicit flexibility along the price dimension that enters the slope
estimand, we include standardized log price directly in the fine-tuning specification. Specifically, the
main case study uses
\[
f^{(s)}(\bm X)
=
c_0+c_1p+c_2f_{\mathrm{zs}}(\bm X),
\]
where $p$ denotes standardized log price and $f_{\mathrm{zs}}(\bm X)$ is the text-only zero-shot
prediction from \texttt{gpt-5.6-sol}. This low-dimensional specification combines the census-wide
text signal with an explicit price component, with only three coefficients fitted using the $s$
labeled samples.

We also evaluate a richer $66$-feature MLP specification that combines $64$ principal components of the text
embedding, price, and the zero-shot prediction. At its own rule-selected allocation, this richer
specification performs worse than the affine fine-tuning specification across the reported estimation metrics. We therefore
use the affine fine-tuning specification in the main case study and omit the detailed MLP results for brevity.

Separately, we evaluate a zero-shot variant of \texttt{gpt-5.6-sol} in which price is supplied
directly in the prompt. This predictor has correlation $0.879$ with the human ratings. Unlike the
text-only zero-shot prediction used as the language-model component of the main fine-tuning specification, this
variant gives the language model access to price at the prompting stage. We therefore report it as
a separate benchmark rather than pooling it with the text-only zero-shot models, and identify this
additional input explicitly wherever the benchmark is used.

\subsection{Controlled Synthetic Analysis of the Objective Gap}
\label{app:synthetic}

This subsection uses a controlled synthetic design to isolate the mechanism underlying the
difference between estimand-aware and conventional training. The design creates a setting in which
a constrained surrogate must trade approximation accuracy between observations with different
influence on the target estimand. This generates a clear objective gap whose source can be examined
directly. We then remove the features that generate this trade-off to verify the conditions under
which the two training objectives become equivalent.

\paragraph{Design.}
We construct a two-region data-generating process consisting of a large low-influence bulk and a
smaller high-influence cluster. The outcome is
\[
Y
=
g_L(Z)
+
\kappa g_H(Z)
+
\delta U
+
\varepsilon,
\]
where
\[
g_L(Z)=\tanh(1.5-Z),
\qquad
g_H(Z)=\tanh(Z-z_{\mathrm{hi}}+1).
\]
The first component primarily captures variation in the low-influence bulk, whereas the second
primarily captures variation in the high-influence cluster. The auxiliary variable $U$ is an
independent persona-like signal available to the surrogate, and
$\varepsilon$ is homoskedastic outcome noise.

The key restriction is imposed through the surrogate class. Rather than observing the two
region-specific components separately, the surrogate receives an entangled feature that combines
them, so that a common coefficient must trade approximation accuracy across the two regions.
Conventional MSE training weights this trade-off primarily according to the relative frequency of
observations in each region. In contrast, the estimand-aware objective assigns greater weight to
observations that contribute more strongly to the reported slope. The estimand is the census
regression slope, and Table~\ref{tab:synth-dgp} summarizes the generating parameters.

\begin{table}[ht!]
\centering
\footnotesize
\caption{Generating parameters of the synthetic design}
\begin{tabular}{llc}
\toprule
\textbf{Component} & \textbf{Role} & \textbf{Value} \\
\midrule
Target-population size
& census size
& $4{,}000$ \\

Cluster location $z_{\mathrm{hi}}$
& center of the high-influence cluster
& $3.5$ \\

Cluster mass
& share of the census in the cluster
& $15\%$ \\

Cluster signal weight $\kappa$
& outcome weight of the high-influence component
& $3.0$ \\

Outcome noise $\sigma$
& homoskedastic noise scale
& $0.8$ \\

Nuisance dimensions
& uninformative embedding coordinates
& $5$ \\

Persona-like coefficient $\delta$
& coefficient on the auxiliary surrogate signal
& $1.0$ \\

Labeled samples $s$
& training size for each objective
& $300$ \\
\bottomrule
\end{tabular}
\label{tab:synth-dgp}
\end{table}

\paragraph{Objective gap and mechanism.}
Across $100$ paired training trials, estimand-aware training attains a slope-specific variance criterion
approximately $31\%$ lower than conventional MSE training. The conventional surrogate includes a
free, unpenalized intercept, and centering the training outcomes leaves the gap essentially
unchanged, at $31.8\%$ with a paired $t$-statistic of $60$. The separation therefore does not arise
from the additive-level effect that can distinguish the two objectives in finite-sample
mean-estimation training. The corresponding downstream slope-estimation results are reported in
Table~\ref{tab:mest-synthetic} of the main text.

The mechanism instead comes from how the two objectives allocate approximation error across the
two regions. Conventional MSE training places greater emphasis on the low-influence bulk because
it contains most observations. Estimand-aware training accepts more residual error in that region
in exchange for substantially reducing error in the smaller high-influence cluster, approximately
halving the cluster-level bias in the fitted design. Because residual errors from high-influence
observations contribute more strongly to the variance of the reported slope, this reallocation
reduces the downstream variance criterion even when it need not improve conventional prediction
accuracy by the same amount.

\paragraph{Removing the preconditions.}
We consider two variants that separately remove the features responsible for the objective gap.
First, we give the surrogate separate access to the two region-specific components rather than only
their entangled representation. This removes the forced approximation trade-off because the
surrogate can fit the low- and high-influence regions independently. Second, we remove the high-influence cluster (drawing $Z$ uniformly on $[-1,1]$) while retaining the
restricted surrogate representation. The entangled feature then no longer has to serve two regions, so
the surrogate class becomes nearly correctly specified; the influence weights remain heterogeneous in
this variant.

In both variants, the objective gap becomes essentially zero. Thus, the separation between
estimand-aware and conventional training requires both heterogeneous influence on the downstream
estimand and a misspecified surrogate class that cannot simultaneously fit the relevant regions.
When the misspecification is removed, as in both variants, the two objectives yield nearly identical
solutions; when the influence is constant, the mean-estimation equivalence of Remark~\ref{rmk:objectives}
applies. This boundary regime is consistent with the near-equivalence observed for the real-corpus
price-rating case study in Section~\ref{sec:mest-case}.

\section{Extension to Independent but Non-Identical Fine-Tuning Data}
\label{sec:external_ft_extension}

In the main paper, we have focused on a single labeled dataset $\sN$ and a single unlabeled dataset $\sM$ drawn from the target distribution, and studied how to optimally allocate the $n$ target labels between fine-tuning and PPI.

In many real applications, however, practitioners also have access to additional labeled data that are \emph{independent} but not necessarily \emph{identically distributed} with the target task. For example, in a survey study, it is common to have a large archive of previous questionnaires and responses across products, features, and customer segments, and then to field a new survey on a specific question of interest. One can first fine-tune an LLM on the archive of prior surveys and responses, and then use that fine-tuned LLM as a starting point for a new product study with limited labels on the new questionnaire; see, e.g., \citet{brand2023using}. This is an ``independent but non-identical'' setting: the prior surveys are not drawn from the same distribution as the new question, but they provide useful prior information about how to map text responses to numerical outcomes.

In the following, we show that our framework naturally accommodates such extra fine-tuning data, and that the optimal sample allocation on the new task remains valid under mild independence assumptions.

\paragraph{Setup with external fine-tuning data.}

We extend the main setup as follows. In addition to the target-task labeled and unlabeled datasets
\[
\sN = \{(X_i,Y_i)\}_{i=1}^n,\qquad
\sM = \{\tilde X_j\}_{j=1}^m,
\]
drawn i.i.d.\ from $(X,Y)\sim\cF$ and $X\sim\cF_X$ as before, we assume there is an extra labeled dataset
\[
\sH = \{(X^{\mathrm{ext}}_k, Y^{\mathrm{ext}}_k)\}_{k=1}^{N_{\mathrm{ext}}},
\]
with an \emph{arbitrary} joint distribution (not necessarily equal to $\cF$), such that
\[
\sH \;\indep\; (\sN,\sM).
\]
The data in $\sH$ may come from previous survey questions, different products, or other related tasks.

A training algorithm $\cA$ takes as input the external dataset $\sH$ and $s$ labeled examples from the new task, say $\sN_{\mathrm{FT}}\subset\sN$ with $|\sN_{\mathrm{FT}}|=s$, and outputs a fine-tuned predictor
\[
f^{(s)} = \cA(\sH,\sN_{\mathrm{FT}}): \cX\to\cY.
\]
The remaining $n-s$ target labels
\[
\sN_{\mathrm{PPI}} = \sN\setminus\sN_{\mathrm{FT}}
\]
are reserved for PPI. The PPI estimator on the new task is then
\begin{align}
\widehat{\mu}^{(s)}
=\frac{1}{n-s}\sum_{(X_i,Y_i)\in\sN_{\mathrm{PPI}}}\!\bigl(Y_i-f^{(s)}(X_i)\bigr)
+\frac{1}{m}\sum_{j=1}^{m} f^{(s)}(\tilde X_j).
\label{eq:ppi_external}
\end{align}

The key structural assumption is that the data used to build $f^{(s)}$, namely $(\sH,\sN_{\mathrm{FT}})$, are independent of the data used for PPI, namely $(\sN_{\mathrm{PPI}},\sM)$; their marginal distributions can differ.

\begin{proposition}[Unbiasedness and variance with external fine-tuning data]
\label{prop:external_ft} Assume:
\begin{enumerate}[(a)]
\item $(X_i,Y_i)$ are i.i.d.\ from $(X,Y)\sim\cF$ for $i=1,\dots,n$; $\tilde X_j$ are i.i.d.\ from $X\sim\cF_X$ for $j=1,\dots,m$, independent of $\{(X_i,Y_i)\}_{i=1}^n$;

\item $\sH$ is an arbitrary labeled dataset, independent of $(\sN,\sM)$;

\item for a given $s$, the predictor $f^{(s)}$ is a measurable function of $(\sH,\sN_{\mathrm{FT}})$ only, with $\sN_{\mathrm{FT}}$ and $\sN_{\mathrm{PPI}}$ forming a partition of $\sN$ of sizes $s$ and $n-s$.
\end{enumerate} Then the estimator $\widehat{\mu}^{(s)}$ defined in~\eqref{eq:ppi_external} satisfies
\begin{align*}
\bE\!\left[\widehat{\mu}^{(s)}\right]
&= \bE[Y],\\[3pt]
\Var\!\left(\widehat{\mu}^{(s)}\right)
&= \frac{1}{n-s}\,\bE\!\left[\Var\bigl(Y-f^{(s)}(X)\,\big|\,f^{(s)}\bigr)\right]
  + \frac{1}{m}\,\bE\!\left[\Var\bigl(f^{(s)}(X)\,\big|\,f^{(s)}\bigr)\right],
\end{align*} where the conditional variances are taken with respect to the target distribution $\cF$ of $(X,Y)$ and the outer expectations are over the training data $(\sH, \sN_{\mathrm{FT}})$; because the estimator is conditionally unbiased, no additional between-training-run term appears.
\end{proposition}

\proof{Proof of Proposition~\ref{prop:external_ft}.} We condition on the fine-tuning data $(\sH, \sN_{\mathrm{FT}})$, which fixes the predictor $f^{(s)}$. By assumption (a) and the independence assumption (b), the datasets $\sN_{\mathrm{PPI}}$ and $\sM$ remain independent of $f^{(s)}$ and consist of i.i.d.\ draws from $\cF$ and $\cF_X$, respectively.

First, we establish unbiasedness. Using the linearity of expectation and the fact that $f^{(s)}$ is fixed conditionally:
\begin{align*}
\bE[\widehat{\mu}^{(s)} \mid f^{(s)}]
&= \frac{1}{n-s}\sum_{(X_i,Y_i)\in\sN_{\mathrm{PPI}}} \bE[Y_i - f^{(s)}(X_i) \mid f^{(s)}] + \frac{1}{m}\sum_{j=1}^m \bE[f^{(s)}(\tilde X_j) \mid f^{(s)}] \\
&= \bE[Y - f^{(s)}(X) \mid f^{(s)}] + \bE[f^{(s)}(X) \mid f^{(s)}] \\
&= \bE[Y] - \bE[f^{(s)}(X)] + \bE[f^{(s)}(X)] \\
&= \bE[Y].
\end{align*} By the Law of Iterated Expectations, $\bE[\widehat{\mu}^{(s)}] = \bE[\bE[\widehat{\mu}^{(s)} \mid f^{(s)}]] = \bE[\bE[Y]] = \bE[Y]$.

Next, we derive the variance. Applying the Law of Total Variance:
\[
\Var(\widehat{\mu}^{(s)}) = \bE[\Var(\widehat{\mu}^{(s)} \mid f^{(s)})] + \Var(\bE[\widehat{\mu}^{(s)} \mid f^{(s)}]).
\]
From the unbiasedness result above, the conditional expectation $\bE[\widehat{\mu}^{(s)} \mid f^{(s)}] = \bE[Y]$ is a constant (the true population mean). Therefore, the second term $\Var(\bE[\widehat{\mu}^{(s)} \mid f^{(s)}]) = 0$.

We now compute the first term. Conditional on $f^{(s)}$, the summands in \eqref{eq:ppi_external} are independent. Thus, the conditional variance is:
\begin{align*}
\Var(\widehat{\mu}^{(s)} \mid f^{(s)})
&= \Var\left( \frac{1}{n-s}\sum_{i} (Y_i - f^{(s)}(X_i)) \;\middle|\; f^{(s)} \right) + \Var\left( \frac{1}{m}\sum_{j} f^{(s)}(\tilde X_j) \;\middle|\; f^{(s)} \right) \\
&= \frac{1}{n-s}\Var(Y-f^{(s)}(X) \mid f^{(s)}) + \frac{1}{m}\Var(f^{(s)}(X) \mid f^{(s)}).
\end{align*} Taking the expectation over $f^{(s)}$ --- so the variance terms of the proposition are the expected conditional variances, and additive-level variation across training draws does not enter --- we obtain the final result.
\hfill \halmos
\endproof

Proposition~\ref{prop:external_ft} shows that our variance decomposition in \eqref{eqn:VarianceDecomposition} holds without change when the fine-tuning procedure uses additional independent data from other tasks. All that matters for the PPI variance on the new task is how well the resulting $f^{(s)}$ predicts $(X,Y)\sim\cF$.

\paragraph{Scaling law and optimal allocation with external data.}

In the main development, we assume that for the \emph{target} task there exist constants $a>0$, $\alpha>0$, and $b\ge 0$ such that the residual variance of $f^{(s)}$ obeys the scaling law
\[
\Var\bigl(Y-f^{(s)}(X)\bigr)
\approx a s^{-\alpha}+b,
\]
when $f^{(s)}$ is obtained by fine-tuning on $s$ independent target labels. In the extended setting above, the same assumption is interpreted \emph{conditional on} the external data $\sH$: for a given external fine-tuning procedure and resulting starting model, there exist $a,\alpha,b$ such that the residual variance on the new task decays as $s$ increases, and the quality of the external data is reflected in these parameters. For example, a rich archive of related surveys may lower the floor $b$ or steepen the decay $\alpha$, though the direction is not guaranteed.

Under this interpretation, the variance approximation in the large-$m$ regime remains
\[
\Var\!\left(\widehat{\mu}^{(s)}\right)
\approx \frac{a s^{-\alpha}+b}{n-s},
\]
and the optimal sample allocation problem on the \emph{new} task is unchanged: we still choose $s\in(0,n)$ to minimize $(a s^{-\alpha}+b)/(n-s)$. This is exactly the same objective as \eqref{eqn:Minimization}. All previous results, including Theorem~\ref{thm:OptimalAllocation} and Propositions~\ref{prop:monotonicity_abn}--\ref{prop:variance_reduction_conditions}, hold here.

In other words, our method continues to require labeled data from the new task for both fine-tuning (to move along the scaling curve) and for PPI (to debias the predictions). Additional independent labeled data from other tasks can be freely used to improve the starting point of the model; their effect is summarized by a more favorable scaling law on the new task. As long as the scaling law is estimated using labeled data from the new task (via the ramp-up procedure in Online Appendix~\ref{appsec:ramp-up}), the same allocation rule solves the same leading-variance problem for that task, regardless of how much extra independent fine-tuning data were available upstream.


\begin{thebibliography}{58}
\providecommand{\natexlab}[1]{#1}
\providecommand{\url}[1]{\texttt{#1}}
\providecommand{\urlprefix}{URL }

\bibitem[{Alain \protect\BIBand{} Bengio(2016)}]{alain2016probing}
Alain G, Bengio Y (2016) Understanding intermediate layers using linear
  classifier probes. \emph{arXiv preprint arXiv:1610.01644} .

\bibitem[{Alfaro \protect\BIBand{} Corbett(2003)}]{alfaro2003value}
Alfaro JA, Corbett CJ (2003) The value of sku rationalization in practice (the
  pooling effect under suboptimal inventory policies and nonnormal demand).
  \emph{Production and Operations Management} 12(1):12--29.

\bibitem[{Angelopoulos et~al.(2023{\natexlab{a}})Angelopoulos, Bates,
  Fannjiang, Jordan, \protect\BIBand{} Zrnic}]{angelopoulos2023prediction}
Angelopoulos AN, Bates S, Fannjiang C, Jordan MI, Zrnic T (2023{\natexlab{a}})
  Prediction-powered inference. \emph{Science} 382(6671):669--674.

\bibitem[{Angelopoulos et~al.(2023{\natexlab{b}})Angelopoulos, Duchi,
  \protect\BIBand{} Zrnic}]{angelopoulos2023ppi}
Angelopoulos AN, Duchi JC, Zrnic T (2023{\natexlab{b}}) {PPI}++: Efficient
  prediction-powered inference. \emph{arXiv preprint arXiv:2311.01453} .

\bibitem[{Argyle et~al.(2023)Argyle, Busby, Fulda, Gubler, Rytting,
  \protect\BIBand{} Wingate}]{argyle2023out}
Argyle LP, Busby EC, Fulda N, Gubler JR, Rytting C, Wingate D (2023) Out of
  one, many: Using language models to simulate human samples. \emph{Political
  Analysis} 31(3):337--351.

\bibitem[{Atkinson et~al.(2007)Atkinson, Donev, \protect\BIBand{}
  Tobias}]{atkinson2007optimum}
Atkinson A, Donev A, Tobias R (2007) \emph{Optimum experimental designs, with
  SAS}, volume~34 (OUP Oxford).

\bibitem[{Bates et~al.(2024)Bates, Hastie, \protect\BIBand{}
  Tibshirani}]{bates2024cross}
Bates S, Hastie T, Tibshirani R (2024) Cross-validation: what does it estimate
  and how well does it do it? \emph{Journal of the American Statistical
  Association} 119(546):1434--1445.

\bibitem[{Berry et~al.(1995)Berry, Levinsohn, \protect\BIBand{}
  Pakes}]{berry1993automobile}
Berry S, Levinsohn J, Pakes A (1995) Automobile prices in market equilibrium.
  \emph{Econometrica} 63(4):841--890.

\bibitem[{Boyd \protect\BIBand{} Vandenberghe(2004)}]{boyd2004convex}
Boyd S, Vandenberghe L (2004) \emph{Convex optimization} (Cambridge university
  press).

\bibitem[{Brand et~al.(2023)Brand, Israeli, \protect\BIBand{}
  Ngwe}]{brand2023using}
Brand J, Israeli A, Ngwe D (2023) Using {LLM}s for market research.
  \emph{Harvard Business School Marketing Unit Working Paper} No.~23-062.

\bibitem[{Brown et~al.(2020)Brown, Mann, Ryder, Subbiah, Kaplan, Dhariwal,
  Neelakantan, Shyam, Sastry, Askell et~al.}]{brown2020gpt3}
Brown T, Mann B, Ryder N, Subbiah M, Kaplan JD, Dhariwal P, Neelakantan A,
  Shyam P, Sastry G, Askell A, et~al. (2020) Language models are few-shot
  learners. \emph{Advances in neural information processing systems}
  33:1877--1901.

\bibitem[{Brucks \protect\BIBand{} Toubia(2023)}]{brucks2023prompt}
Brucks M, Toubia O (2023) Prompt architecture can induce methodological
  artifacts in large language models. \emph{Available at SSRN 4484416} .

\bibitem[{Cheng et~al.(2025)Cheng, Ofek, \protect\BIBand{}
  Yoganarasimhan}]{ofek2025balancing}
Cheng M, Ofek E, Yoganarasimhan H (2025) Balancing engagement and polarization:
  Multi-objective alignment of news content using {LLM}s. \emph{arXiv preprint
  arXiv:2504.13444} .

\bibitem[{Elmachtoub \protect\BIBand{} Grigas(2022)}]{elmachtoub2022smart}
Elmachtoub AN, Grigas P (2022) Smart “predict, then optimize”.
  \emph{Management Science} 68(1):9--26.

\bibitem[{Fedorov(2013)}]{fedorov2013theory}
Fedorov VV (2013) \emph{Theory of optimal experiments} (Elsevier).

\bibitem[{Ferreira et~al.(2016)Ferreira, Lee, \protect\BIBand{}
  Simchi-Levi}]{ferreira2016analytics}
Ferreira KJ, Lee BHA, Simchi-Levi D (2016) Analytics for an online retailer:
  Demand forecasting and price optimization. \emph{Manufacturing \& service
  operations management} 18(1):69--88.

\bibitem[{Gallego et~al.(2019)Gallego, Topaloglu et~al.}]{gallego2019revenue}
Gallego G, Topaloglu H, et~al. (2019) \emph{Revenue management and pricing
  analytics}, volume 209 (Springer).

\bibitem[{Geisser(1975)}]{geisser1975predictive}
Geisser S (1975) The predictive sample reuse method with applications.
  \emph{Journal of the American statistical Association} 70(350):320--328.

\bibitem[{Green \protect\BIBand{} Srinivasan(1990)}]{green1990conjoint}
Green PE, Srinivasan V (1990) Conjoint analysis in marketing: new developments
  with implications for research and practice. \emph{Journal of marketing}
  54(4):3--19.

\bibitem[{Gui \protect\BIBand{} Toubia(2023)}]{gui2023challenge}
Gui G, Toubia O (2023) The challenge of using {LLMs} to simulate human
  behavior: A causal inference perspective. \emph{arXiv preprint
  arXiv:2312.15524} .

\bibitem[{Han et~al.(2026)Han, Hu, \protect\BIBand{} Zhang}]{han2026debiasing}
Han J, Hu M, Zhang Z (2026) Debiasing ai simulations with human data.
  \emph{Available at SSRN 7377338} .

\bibitem[{Han et~al.(2024)Han, Gao, Liu, Zhang, \protect\BIBand{}
  Zhang}]{han2024peft}
Han Z, Gao C, Liu J, Zhang J, Zhang SQ (2024) Parameter-efficient fine-tuning
  for large models: A comprehensive survey. \emph{arXiv preprint
  arXiv:2403.14608} .

\bibitem[{Hernandez et~al.(2021)Hernandez, Kaplan, Henighan, \protect\BIBand{}
  McCandlish}]{hernandez2021transfer}
Hernandez D, Kaplan J, Henighan T, McCandlish S (2021) Scaling laws for
  transfer. \emph{arXiv preprint arXiv:2102.01293} .

\bibitem[{Hoffmann et~al.(2022)Hoffmann, Borgeaud, Mensch, Buchatskaya, Cai,
  Rutherford, Casas, Hendricks, Welbl, Clark et~al.}]{hoffmann2022chinchilla}
Hoffmann J, Borgeaud S, Mensch A, Buchatskaya E, Cai T, Rutherford E, Casas
  DdL, Hendricks LA, Welbl J, Clark A, et~al. (2022) Training compute-optimal
  large language models. \emph{arXiv preprint arXiv:2203.15556} .

\bibitem[{Hu et~al.(2022)Hu, Shen, Wallis, Allen-Zhu, Li, Wang, Wang, Chen
  et~al.}]{hu2021lora}
Hu EJ, Shen Y, Wallis P, Allen-Zhu Z, Li Y, Wang S, Wang L, Chen W, et~al.
  (2022) Lora: Low-rank adaptation of large language models. \emph{ICLR}
  1(2):3.

\bibitem[{Ji et~al.(2025)Ji, Lei, \protect\BIBand{} Zrnic}]{ji2025predictions}
Ji W, Lei L, Zrnic T (2025) Predictions as surrogates: Revisiting surrogate
  outcomes in the age of ai. \emph{arXiv preprint arXiv:2501.09731} .

\bibitem[{Kaplan et~al.(2020)Kaplan, McCandlish, Henighan, Brown, Chess, Child,
  Gray, Radford, Wu, \protect\BIBand{} Amodei}]{kaplan2020scaling}
Kaplan J, McCandlish S, Henighan T, Brown TB, Chess B, Child R, Gray S, Radford
  A, Wu J, Amodei D (2020) Scaling laws for neural language models. \emph{arXiv
  preprint arXiv:2001.08361} .

\bibitem[{Krsteski et~al.(2025)Krsteski, Russo, Chang, West, \protect\BIBand{}
  Gligori{\'c}}]{krsteski2025valid}
Krsteski S, Russo G, Chang S, West R, Gligori{\'c} K (2025) Valid survey
  simulations with limited human data: The roles of prompting, fine-tuning, and
  rectification. \emph{arXiv preprint arXiv:2510.11408} .

\bibitem[{Kumar et~al.(2022)Kumar, Raghunathan, Jones, Ma, \protect\BIBand{}
  Liang}]{kumar2022finetuning}
Kumar A, Raghunathan A, Jones R, Ma T, Liang P (2022) Fine-tuning can distort
  pretrained features and underperform out-of-distribution. \emph{International
  Conference on Learning Representations}.

\bibitem[{Lee et~al.(1993)Lee, Billington, \protect\BIBand{}
  Carter}]{lee1993hewlett}
Lee HL, Billington C, Carter B (1993) Hewlett-packard gains control of
  inventory and service through design for localization. \emph{Interfaces}
  23(4):1--11.

\bibitem[{Lee \protect\BIBand{} Tang(1997)}]{lee1997modelling}
Lee HL, Tang CS (1997) Modelling the costs and benefits of delayed product
  differentiation. \emph{Management science} 43(1):40--53.

\bibitem[{Li et~al.(2025)Li, Chen, Namkoong, \protect\BIBand{}
  Peng}]{li2025llm}
Li A, Chen H, Namkoong H, Peng T (2025) {LLM} generated persona is a promise
  with a catch. \emph{arXiv preprint arXiv:2503.16527} .

\bibitem[{Li et~al.(2024)Li, Castelo, Katona, \protect\BIBand{}
  Sarvary}]{li2024frontiers}
Li P, Castelo N, Katona Z, Sarvary M (2024) Frontiers: Determining the validity
  of large language models for automated perceptual analysis. \emph{Marketing
  Science} 43(2):254--266.

\bibitem[{Motoki et~al.(2024)Motoki, Pinho~Neto, \protect\BIBand{}
  Rodrigues}]{motoki2024more}
Motoki F, Pinho~Neto V, Rodrigues V (2024) More human than human: measuring
  chatgpt political bias. \emph{Public Choice} 198(1):3--23.

\bibitem[{Newey \protect\BIBand{} McFadden(1994)}]{newey1994large}
Newey WK, McFadden D (1994) Large sample estimation and hypothesis testing.
  \emph{Handbook of econometrics} 4:2111--2245.

\bibitem[{{OpenAI}(2025)}]{openai2025gptoss}
{OpenAI} (2025) Introducing gpt-oss.
  \urlprefix\url{https://openai.com/index/introducing-gpt-oss/}, accessed:
  2025-12.

\bibitem[{Ouyang et~al.(2022)Ouyang, Wu, Jiang, Almeida, Wainwright, Mishkin,
  Zhang, Agarwal, Slama, Ray et~al.}]{ouyang2022instruct}
Ouyang L, Wu J, Jiang X, Almeida D, Wainwright C, Mishkin P, Zhang C, Agarwal
  S, Slama K, Ray A, et~al. (2022) Training language models to follow
  instructions with human feedback. \emph{Advances in neural information
  processing systems} 35:27730--27744.

\bibitem[{Peng et~al.(2025)Peng, Gui, Merlau, Fan, Sliman, Brucks, Johnson,
  Morwitz, Althenayyan, Bellezza et~al.}]{peng2025mega}
Peng T, Gui G, Merlau DJ, Fan GJ, Sliman MB, Brucks M, Johnson EJ, Morwitz V,
  Althenayyan A, Bellezza S, et~al. (2025) A mega-study of digital twins
  reveals strengths, weaknesses and opportunities for further improvement.
  \emph{arXiv preprint arXiv:2509.19088} .

\bibitem[{Pezeshkpour \protect\BIBand{} Hruschka(2024)}]{pezeshkpour2023order}
Pezeshkpour P, Hruschka E (2024) Large language models sensitivity to the order
  of options in multiple-choice questions. \emph{Findings of the Association
  for Computational Linguistics: NAACL 2024}, 2006--2017.

\bibitem[{Pukelsheim(2006)}]{pukelsheim2006optimal}
Pukelsheim F (2006) \emph{Optimal design of experiments} (SIAM).

\bibitem[{Rafailov et~al.(2023)Rafailov, Sharma, Mitchell, Manning, Ermon,
  \protect\BIBand{} Finn}]{rafailov2023direct}
Rafailov R, Sharma A, Mitchell E, Manning CD, Ermon S, Finn C (2023) Direct
  preference optimization: Your language model is secretly a reward model.
  \emph{Advances in neural information processing systems} 36:53728--53741.

\bibitem[{Rubin \protect\BIBand{} van~der Laan(2008)}]{rubin2008empirical}
Rubin DB, van~der Laan MJ (2008) Empirical efficiency maximization: improved
  locally efficient covariate adjustment in randomized experiments and survival
  analysis. \emph{The International Journal of Biostatistics} 4(1).

\bibitem[{Silvey(2013)}]{silvey2013optimal}
Silvey S (2013) \emph{Optimal design: an introduction to the theory for
  parameter estimation}, volume~1 (Springer Science \& Business Media).

\bibitem[{Stone(1974)}]{stone1974cross}
Stone M (1974) Cross-validatory choice and assessment of statistical
  predictions. \emph{Journal of the royal statistical society: Series B
  (Methodological)} 36(2):111--133.

\bibitem[{Talluri \protect\BIBand{} Van~Ryzin(2006)}]{talluri2006theory}
Talluri KT, Van~Ryzin GJ (2006) \emph{The theory and practice of revenue
  management}, volume~68 (Springer Science \& Business Media).

\bibitem[{Thoutt(2017)}]{wine_reviews_kaggle}
Thoutt Z (2017) Wine review dataset. Kaggle dataset,
  \urlprefix\url{https://www.kaggle.com/datasets/zynicide/wine-reviews/data},
  accessed: 2025-12.

\bibitem[{Tirole(1988)}]{tirole1988theory}
Tirole J (1988) \emph{The theory of industrial organization} (MIT press).

\bibitem[{Toubia et~al.(2025)Toubia, Gui, Peng, Merlau, Li, \protect\BIBand{}
  Chen}]{toubia2025database}
Toubia O, Gui GZ, Peng T, Merlau DJ, Li A, Chen H (2025) Database report:
  Twin-2k-500: A data set for building digital twins of over 2,000 people based
  on their answers to over 500 questions. \emph{Marketing Science}
  44(6):1446--1455.

\bibitem[{Vafa et~al.(2025)Vafa, Athey, \protect\BIBand{}
  Blei}]{vafa2025estimating}
Vafa K, Athey S, Blei DM (2025) Estimating wage disparities using foundation
  models. \emph{Proceedings of the National Academy of Sciences}
  122(22):e2427298122.

\bibitem[{van~der Laan \protect\BIBand{} Van
  Der~Laan(2026)}]{van2026calibeating}
van~der Laan L, Van Der~Laan M (2026) Calibeating prediction-powered inference.
  \emph{arXiv preprint arXiv:2604.21260} .

\bibitem[{Wang et~al.(2024)Wang, Zhang, \protect\BIBand{}
  Zhang}]{wang2024market}
Wang M, Zhang DJ, Zhang H (2024) Large language models for market research: A
  data-augmentation approach. \emph{arXiv preprint arXiv:2412.19363} .

\bibitem[{Wei et~al.(2022)Wei, Wang, Schuurmans, Bosma, Xia, Chi, Le, Zhou
  et~al.}]{wei2022cot}
Wei J, Wang X, Schuurmans D, Bosma M, Xia F, Chi E, Le QV, Zhou D, et~al.
  (2022) Chain-of-thought prompting elicits reasoning in large language models.
  \emph{Advances in neural information processing systems} 35:24824--24837.

\bibitem[{Wertenbroch \protect\BIBand{}
  Skiera(2002)}]{wertenbroch2002measuring}
Wertenbroch K, Skiera B (2002) Measuring consumers' willingness to pay at the
  point of purchase. \emph{Journal of marketing research} 39(2):228--241.

\bibitem[{Wind \protect\BIBand{} Mahajan(1997)}]{wind1997issues}
Wind J, Mahajan V (1997) Issues and opportunities in new product development:
  An introduction to the special issue. \emph{Journal of marketing research}
  34(1):1--12.

\bibitem[{Ye et~al.(2025)Ye, Yoganarasimhan, \protect\BIBand{}
  Zheng}]{ye2025lola}
Ye Z, Yoganarasimhan H, Zheng Y (2025) {LOLA}: {LLM}-assisted online learning
  algorithm for content experiments. \emph{Marketing Science} 44(5):995--1016.

\bibitem[{Zhang et~al.(2024)Zhang, Liu, Cherry, \protect\BIBand{}
  Firat}]{zhang2024scalingft}
Zhang B, Liu Z, Cherry C, Firat O (2024) When scaling meets llm finetuning: The
  effect of data, model and finetuning method. \emph{arXiv preprint
  arXiv:2402.17193} .

\bibitem[{Ziems et~al.(2024)Ziems, Held, Shaikh, Chen, Zhang, \protect\BIBand{}
  Yang}]{ziems2024can}
Ziems C, Held W, Shaikh O, Chen J, Zhang Z, Yang D (2024) Can large language
  models transform computational social science? \emph{Computational
  Linguistics} 50(1):237--291.

\bibitem[{Zrnic \protect\BIBand{} Cand{\`e}s(2024)}]{zrnic2024cross}
Zrnic T, Cand{\`e}s EJ (2024) Cross-prediction-powered inference.
  \emph{Proceedings of the National Academy of Sciences} 121(15):e2322083121.

\end{thebibliography}
\end{document}